\documentclass[10pt,journal,compsoc]{IEEEtran}
\ifCLASSOPTIONcompsoc
\usepackage[nocompress]{cite}
\else
\usepackage{cite}
\fi
\ifCLASSINFOpdf
\else
\fi
\usepackage{times}
\usepackage{epsfig}
\usepackage{graphicx}
\usepackage{amsmath}
\usepackage{amssymb}

\usepackage{subfig}
\usepackage{float}
\usepackage[pagebackref=true,breaklinks=true,letterpaper=true,colorlinks,bookmarks=false]{hyperref}
\hypersetup{draft}
\begin{document}
%
\title{MCMLSD: A Probabilistic Algorithm and Evaluation Framework for Line Segment Detection}
%
%
%
%

\author{James~H.~Elder,~\IEEEmembership{Member,~IEEE,}
        Emilio~J.~Almaz\`{a}n,
Yiming~Qian,~\IEEEmembership{Student Member,~IEEE,}
and~       
Ron~Tal
        }

\IEEEtitleabstractindextext{%
\begin{abstract}
Traditional approaches to line segment detection typically involve perceptual grouping in the image domain and/or global accumulation in the Hough domain.  Here we propose a probabilistic algorithm that merges the advantages of both approaches.  In a first stage lines are detected using a global probabilistic Hough approach.  In the second stage each detected line is analyzed in the image domain to localize the line segments that generated the peak in the Hough map.  By limiting search to a line, the distribution of segments over the sequence of points on the line can be modeled as a Markov chain, and a probabilistically optimal labelling can be computed exactly using a standard dynamic programming algorithm, in linear time.  The Markov assumption also leads to an intuitive ranking method that uses the local marginal posterior probabilities to estimate the expected number of correctly labelled points on a segment.  To assess the resulting Markov Chain Marginal Line Segment Detector (MCMLSD) we develop and apply a novel quantitative evaluation methodology that controls for under- and over-segmentation.  Evaluation on the YorkUrbanDB and Wireframe datasets shows that the proposed MCMLSD method outperforms prior traditional approaches, as well as more recent deep learning methods.    

\end{abstract}

\begin{IEEEkeywords}
Line Segment Detection, Hough Transform, Probabilistic Models, Perceptual Organization, Evaluation Methodologies
\end{IEEEkeywords}}
\maketitle

\IEEEdisplaynontitleabstractindextext

%
\IEEEpeerreviewmaketitle

\IEEEraisesectionheading{\section{Introduction}\label{sec:introduction}}
Much of our visual world can be approximated as piecewise planar, particularly in built environments.  The boundaries and creases of these piecewise planar surfaces project to the image as line segments, and as a consequence the accurate detection of line segments continues to be one of the most important low-level problems in the field of computer vision.  Line segments are important features for many tasks, including feature matching across views~\cite{schmid1997automatic}, vanishing point detection~\cite{kovsecka2002video} and 3D reconstruction~\cite{parodi19963d,zhang2014structure,hofer2017efficient}.

Two frameworks have been popular for line segment detection: perceptual grouping and global Hough analysis.

\subsection{The perceptual grouping approach}  
In the perceptual grouping framework, a set of heuristics typically based upon geometric grouping cues (e.g., proximity, good continuation) is used to group roughly collinear local features (e.g., edges or vectors tangent to isophotes) into extended line segments, which are evaluated according to some quality of fit measure.  An early  example is the hierarchical heuristic framework developed by Boldt and colleagues~\cite{boldt1989token}.  More recent multi-stage grouping efforts include the SSWMS approach of Nieto et al.~\cite{nieto2011line}, which involves an iterative selection of image points with strongly oriented gradient structure, followed by an iterative growing process, the approach of Lu et al.~\cite{lu2015cannylines}, which involves both linking and splitting, and the biologically inspired approach of Liu et al.~\cite{liu2015intelligent}, which employs `simple cell' filters to detect local oriented structure, `complex cell' mechanisms that locally integrate these responses and `hyper-complex' mechanisms to detect endpoints.  

An alternative to this multi-stage grouping approach  is to analyze the covariance matrix of image locations in a set of connected edges and label a set as a line segment if the smallest eigenvalue falls below a threshold~\cite{guru2004simple,liu2014robust}.  While beautifully simple, these methods are not robust to gaps or intersections in the edge map.

Another  issue in this perceptual grouping framework is that some threshold on the quality of fit measure must be applied in order to discriminate `true' line segments from false conjunctions that might arise by chance.  This issue was addressed in the LSD framework introduced by von Gioi et al. \cite{von2008lsd} and based on earlier work by Desolneux et al. \cite{desolneux2000meaningful}. In this framework the so-called {\em a-contrario} approach is used to explicitly compute the probability that inferred line segments might have occurred by chance, given a maximum entropy model of the edge map.   (This is related to the minimum reliable scale null hypothesis testing framework for edge detection developed by Elder \& Zucker~\cite{elder1998local}.)  While this approach does not eliminate the need for a threshold, it transfers the threshold to a quantity (e.g., expected number of false positives per image) that is much easier to set rationally.  A much faster version of this method dubbed EDLines was later introduced by Akinlar \& Topal~\cite{akinlar2011edlines}. 

Recent work in this area has focused on trying to discriminate salient or important line segments from less important `background' segments.  
Kim et al.~\cite{kim2015extracting} used a combination of luminance and geometric features to select the most significant edges, reporting superior performance to LSD on two test images.
Brown et al.~\cite{brown2015generalisable} used a measure of divergence between colour statistics on either side of a hypothesized line segment to favour salient segments.  The method outperformed LSD and Hough methods using quantitative measures of repeatability and registration accuracy on image pairs (see Section \ref{sec:prior_evaluation} below).  

\subsection{The Hough approach}
A drawback of the perceptual grouping approach is that local decisions are made before potentially relevant global information can be brought to bear.  The Hough approach avoids this problem by accumulating edges over the entire image into a histogram of potential line positions and orientations.  Accuracy can be improved by modeling uncertainty in local edges and propagating that uncertainty to the Hough map~\cite{tal2013accurate}.

While the Hough approach to line detection has the advantage of integrating information globally,  identifying the endpoints that define the extent of the line segment in the image is not necessarily straightforward.   A number of methods scan the detected lines in the image space looking for a maximal chain of connected or nearly-connected edges~\cite{guil1995fast,matas2000robust}.  Others have attempted to identify the endpoints of each line segment by analyzing the exact shape of a characteristic `butterfly' pattern around the associated peak in the Hough map~\cite{kamat1998complete,furukawa2003accurate,xu2015accurate,xu2015closed,xu2015statistical}.  One major limitation of this approach is that only one segment can be found per line, whereas in built environments it is quite common to find multiple co-linear segments.  

\section{Our approach}
The advantage of the Hough approach is that it can integrate all evidence for line hypotheses prior to inference.  The perceptual grouping approach, on the other hand, allows endpoints to be detected more directly, and permits the identification of multiple segments per line.  

Our two-stage method, an early version of which was published at CVPR 2017~\cite{almazen2017dynamic}, combines the advantages of these two approaches.  In the first stage we employ the probabilistic Hough method of Tal \& Elder~\cite{tal2013accurate} to identify globally optimal lines. In the second stage we search each of these lines in the image for the segment(s) that gave rise to it.

The key observation that recommends this approach is that narrowing the search for segments from the  2D image to 1D lines allows the problem to be modeled as the labelling of hidden states in a linear Markov chain model.  The problem of determining the maximum probability (MAP) assignment of segments can then be shown to have an optimal substructure property that leads to an exact dynamic programming solution in linear time.    

The benefits of this approach are several:
\begin{enumerate}
	\item Each of the lines identified by a peak in the Hough map results from careful accumulation of the global evidence for the line, and thus will more accurately identify the position and orientation $(\rho,\theta)$ parameters of the line segments than will a few local edges.
	\item The lines identified by the probabilistic Hough method have a natural order according to their significance in the Hough map, allowing the line segment search to be limited to the most significant lines.
	\item In urban scenes, co-linear line segments are common, arising from architectural repetition seen in cladding,  windows, etc. Unlike many Hough methods, our approach allows multiple segments to be recovered for each line.
	\item Limiting search to a line allows the problem of determining maximum probability segments to be solved exactly, using dynamic programming, in linear time. 
\end{enumerate}

\section{The deep learning approach}
Most recently, two deep neural network (DNN) algorithms for line segment detection have been reported.  The Wireframe algorithm~\cite{wireframe_cvpr18}  is based upon a stacked hourglass network~\cite{newell2016stacked} that takes as input a $320\times320$ pixel RGB image and produces as output a $320\times320$ pixel map encoding the estimated locations and lengths of the line segments in the image.  In particular, if a pixel is judged to lie on a segment, the pixel value indicates the estimated length of that segment, while a pixel not lying on any segment should have value 0.  The network is trained to minimize an $L^2$ loss and the scalar output is thresholded to filter out shorter or lower-confidence segments.

Note that the Wireframe network delivers a raster map - essentially an edge map where edges are constrained to lie on straight lines - rather than a vectorized description of the locations, lengths and orientations of the line segments in the image.  To obtain the latter,  a parallel network is trained to detect junctions in the image, and then a somewhat complex process is followed to segment the edge map into line segments between junctions.

The Attraction Field algorithm~\cite{xue2018learning} also employs a deep network, but in a rather different way.  The key insight is that it is easier for a deep network to map the input image to a dense pixel grid of values than to a sparse boundary map.  Thus to adapt the problem of line segment detection to deep networks, each sparse ground truth line segment map is used to generate a dense ground truth `attraction field' map (AFM) that represents the vector displacement to the nearest line segment point at every pixel in the image.  A network is then trained to estimate this dense AFM.  At inference, the estimated dense AFM is reduced to a sparse line segment map using a `squeeze module' that  accumulates votes for line segment pixels by summing discretized displacement vectors over all pixels in the AFM.  

The authors experiment with two network architectures:  A U-Net~\cite{ronneberger2015unet} and a modified U-Net, referred to by the authors as {\em a-trous}, that uses the ASSP module of DeepLab v3+~\cite{chen2018encoder} and the skip connections of ResNet~\cite{he2016deep}.  An $L^1$ loss on the AFM is employed.

As for Wireframe, the AFM approach produces a raster map of edge pixels that must then somehow be grouped into line segments.  In the case of AFM, a heuristic, iterative, greedy region-growing approach similar to that used in LSD\cite{von2008lsd} is employed.

Both Wireframe and Attraction Field algorithms are trained on the Wireframe training dataset.

Both Wireframe and AFM are claimed to outperform all prior non-DNN approaches to line segment detection, including our MCMLSD approach~\cite{almazen2017dynamic}.  However, we argue in this paper that the evaluation performed in these two DNN papers is limited, and that
a more careful analysis reveals that for the problem of line segment detection (not edge detection), MCMLSD and indeed some older
non-DNN algorithms substantially outperform both of these DNN approaches on a number of metrics.  
Given that these networks also
involve tens of millions of free parameters, we argue that more explainable methods such as MCMLSD may be preferable for many applications.

\section{Prior Evaluation Methodology}
\label{sec:prior_evaluation}
Due in part to a lack of high quality labelled ground truth, 
most traditional line segment detection methods were evaluated only qualitatively on real imagery~\cite{boldt1989token,furukawa2003accurate,von2008lsd,akinlar2011edlines,liu2014robust}.  More recently, quantitative evaluations have been conducted based on datasets consisting of pairs of images related by a known homography~\cite{brown2015generalisable}.  This is a promising method, but it does suffer from two potential drawbacks.  First, it is restricted to an analysis of co-planar line segments.  Second, the evaluation presupposes that the goal of line segment detection is for the association of these segments across images for the purposes of homography or disparity estimation.  However there are many other possible applications - single view reconstruction, for example.

While task-specific evaluation methodologies may be appropriate in some cases,  it would be nice to have an evaluation method that is more general.  
In this work we present a new methodology for quantitative evaluation of line segment detectors on real images that does not assume a specific task, using images from the YorkUrbanDB~\cite{denis2008efficient}  (\url{www.elderlab.yorku.ca/YorkUrbanDB/}) and Wireframe (\url{https://github.com/huangkuns/wireframe}) datasets.

\section{Algorithm}
\label{sec:alg}

\subsection{Line Detection}
One problem with traditional Houghing methods is that noise in the observations tends to cause each line to generate multiple peaks in the Hough map.  To address this issue 
we employ a probabilistic Hough transform method~\cite{tal2013accurate} (code available from \url{elderlab.yorku.ca/resources}).   The method uses edges detected by the multi-scale Elder \& Zucker edge detector~\cite{elder1998local}, models  uncertainty in the location and orientation of the detected edges and propagates this uncertainty to the Hough map. This propagation of uncertainty produces a smooth Hough map that is roughly resolution invariant and greatly reduces the multiple response problem.  The problem is mitigated further by a sequential line extraction step in which each peak in the Hough map is visited in descending order of significance, and edges contributing to the peak are subtracted from the Hough map when it is visited.  

\subsection{Line Segment Detection}
\label{sec:segDetection}
Each selected peak in the Hough map identifies a line that extends from one of the image borders to another.  In general, this line is only partially occupied by line segments in the image.  The goal is now to find these segments, based on the location and orientation of nearby edges.

\begin{figure}[htbp]
	\centering
	\includegraphics[width=0.65\columnwidth]{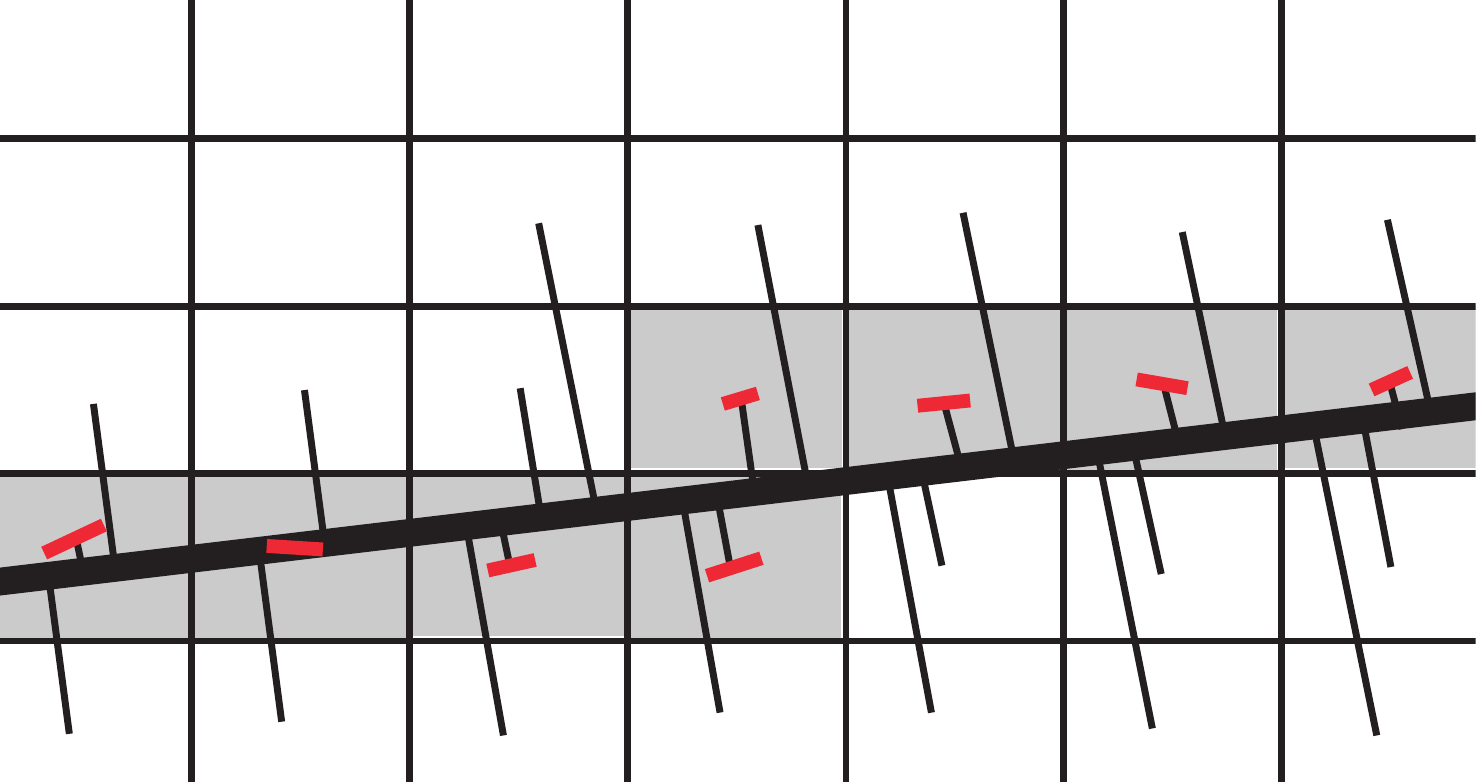}
	\caption{Orthogonal projections (thin black lines) of all pixels within two pixels of a detected line (thick black line) define an ordinal sampling of the line $i\in\left[1\ldots N\right]$.   Pixels within this band occupied by edges (shown red on grey) with orientations similar to the line support the assignment of the ON state for the associated segment variable $x_i$ at sampled line locations.}
	\label{fig:obs}
\end{figure}

Prior work~\cite{denis2008efficient} suggests that most edges generated by a line and detected by the Elder \& Zucker edge detector  lie within one pixel of the line.  To ensure we capture all edges related to a line we extend our search to all pixels within two pixels of the line  (Fig.~\ref{fig:obs}).  The orthogonal projections of these pixel locations onto the line then define an ordinal sampling $i\in[1,\ldots,N]$ of the line.  
We let $x_i$ represent the binary hidden segment state (ON or OFF) indicating whether a visible segment is present at position $i$ on the line, $d_i$ the distance from the line to the associated pixel and $y_i$ the associated image observation at that pixel. Each observation $y_i$ consists of 1-2 features:
\begin{enumerate}
	\item A binary variable $e_i$ indicating whether an edge exists at this pixel.
	\item The angular deviation $\theta_i$ of the edge from the line, if the edge exists.\end{enumerate}

These features provide information about the probable state of the line at the associated position: \\ 
$p(y_i|x_i) \propto p(e_i = 1|x_i,d_i) p(\theta_i|x_i, e_i = 1)$ for edge pixels.\\
$p(y_i|x_i) \propto p(e_i = 0|x_i, d_i)$ for non-edge pixels. \\
(Note that we have assumed that the angular deviation $\theta_i$ is independent of the distance $d_i$ of the pixel from the line.)

\begin{figure}[htbp]
	\centering
	\begin{tabular}{rr}
		\subfloat[]{
			\includegraphics[width=0.42\columnwidth]{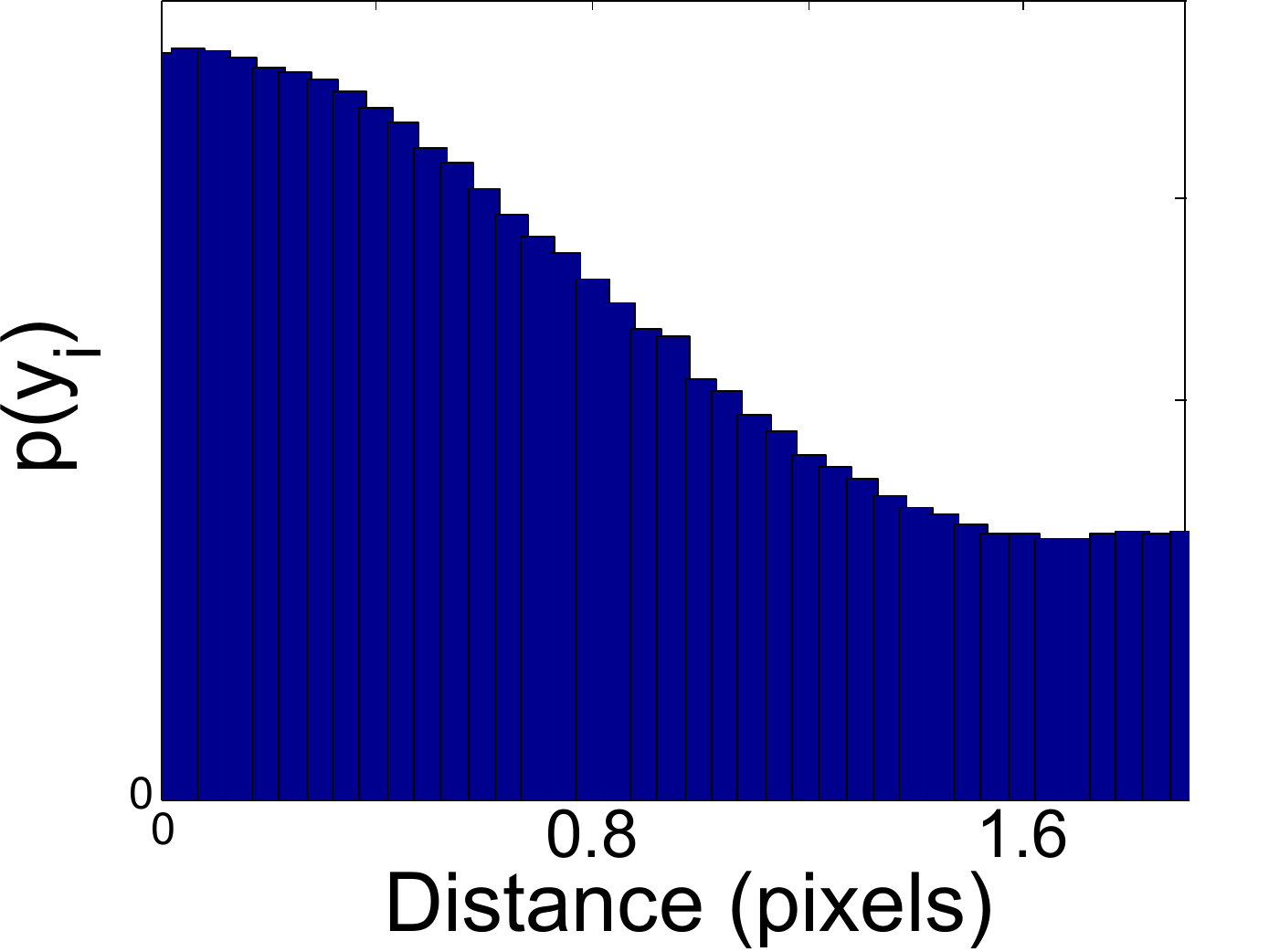}
		} &
		\subfloat[]{
			\includegraphics[width=0.42\columnwidth]{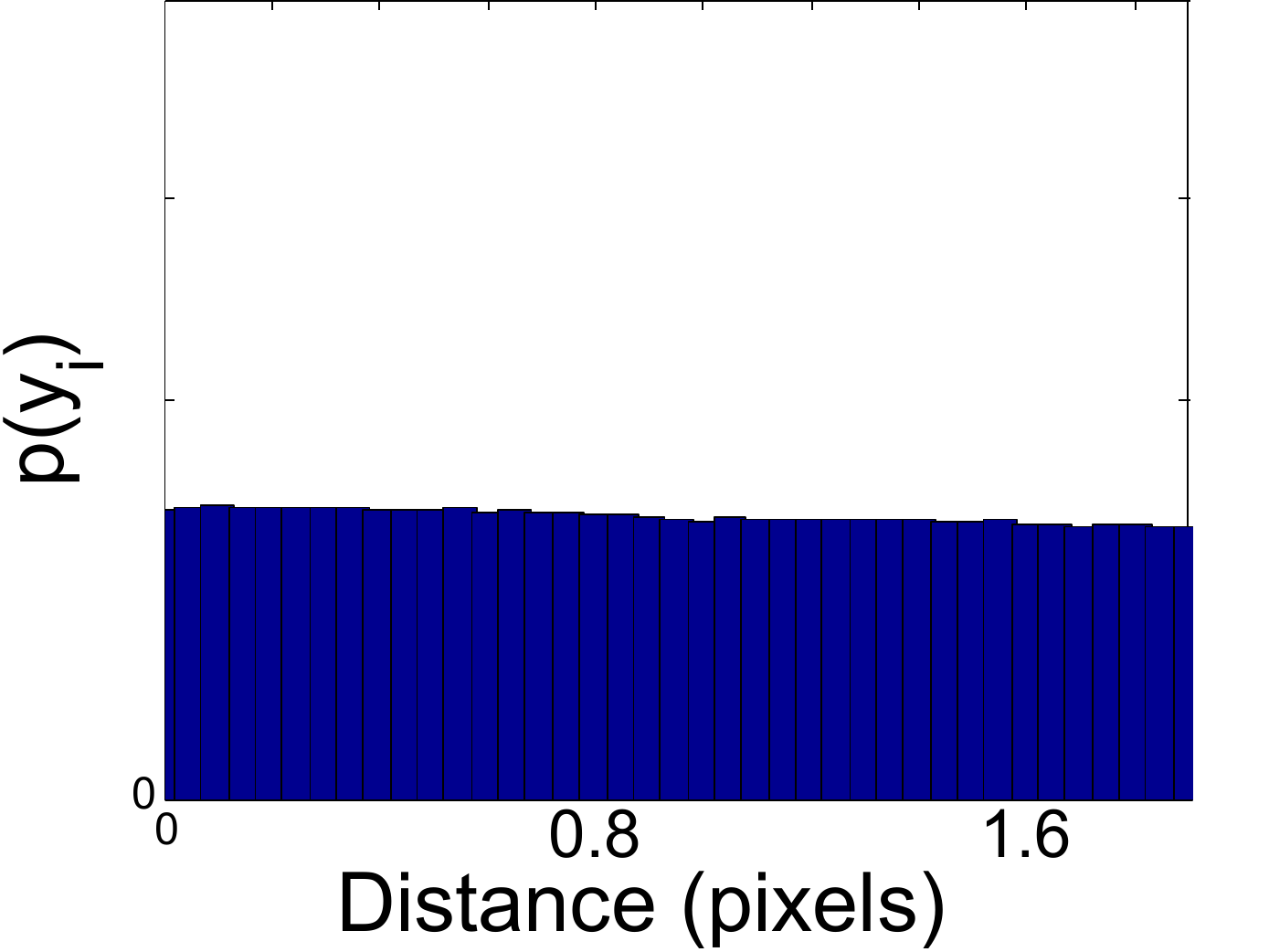}
		}\\
		\subfloat[]{
			\includegraphics[width=0.38\columnwidth]{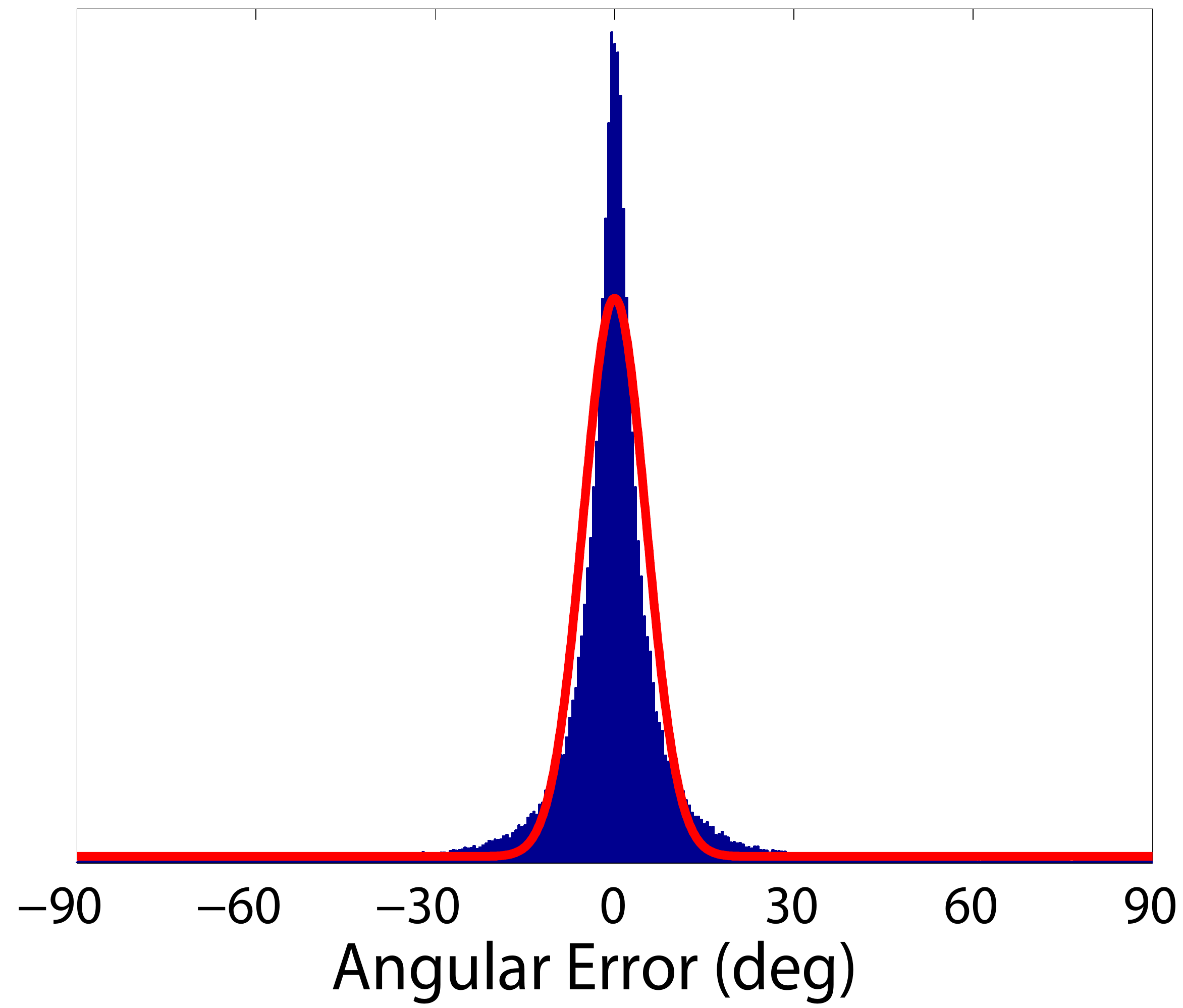}
		} &
		\subfloat[]{
			\includegraphics[width=0.38\columnwidth]{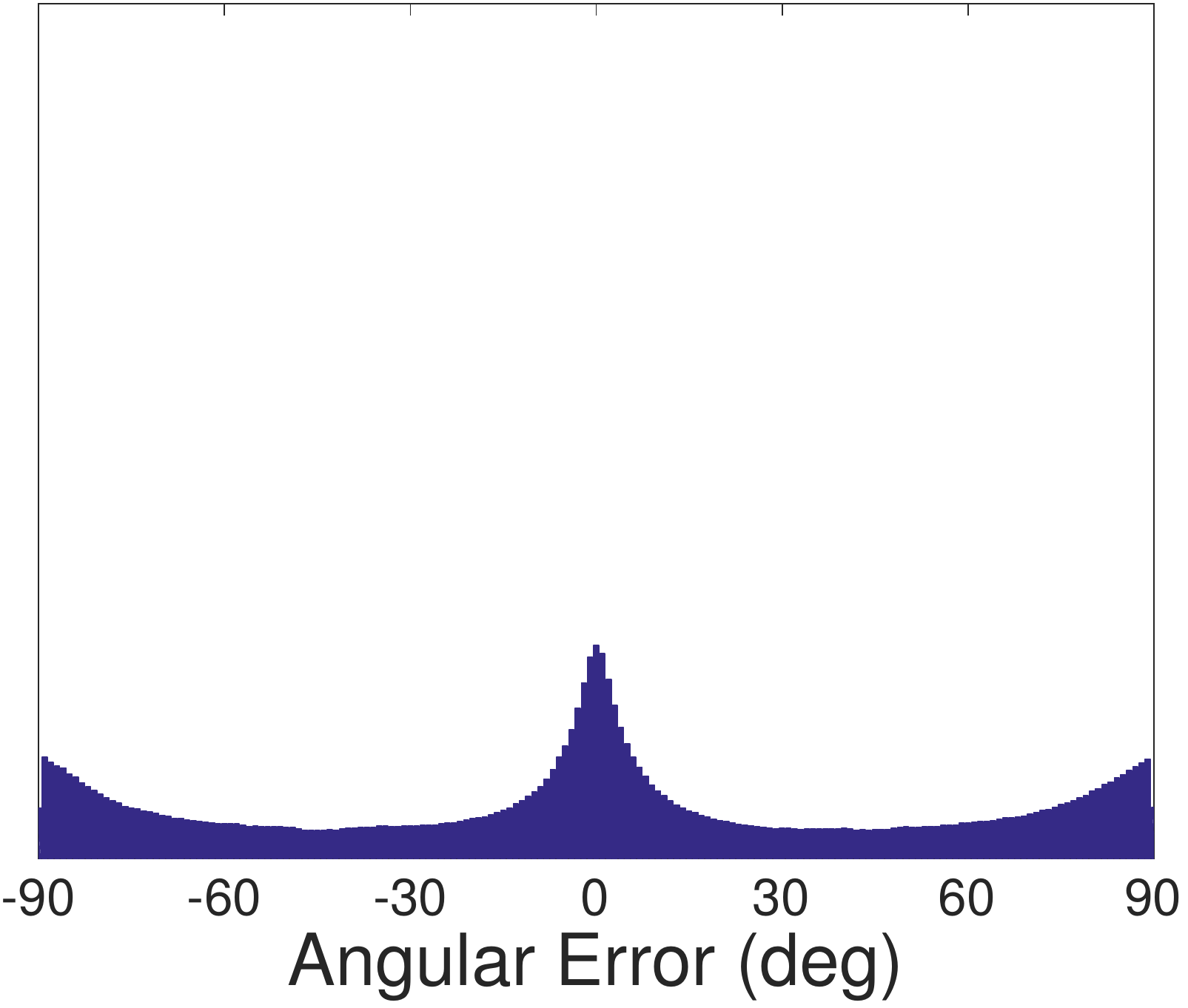}
		}
	\end{tabular}
	\caption{Likelihoods for line segment extraction, learned from the YorkUrbanDB training dataset~\cite{denis2008efficient}.  (a-b) Likelihood $p\left(e_i|x_i,d_i\right)$ for distance $d_i$ of observations from line for (a) ON ($x_i=1$) and (b) OFF ($x_i=0$) states. (c-d) Probability  $p\left(\theta_i|x_i,e_i\right)$ for the angular deviation $\theta_i$ of observed edges from the line for (c) ON ($x_i=1$) and (d) OFF ($x_i=0$) states.}
	\label{fig:like}
\end{figure}

We learned these distributions from the $640{\times} 480$ pixel images and hand-labelled ground truth lines of the YorkUrbanDB training dataset~\cite{denis2008efficient}.  
Figs.~\ref{fig:like}(a-b) show the likelihoods \hbox{$p(e_i = 1|x_i={\rm ON}, d_i)$} and $p(e_i = 1|x_i={\rm OFF}, d_i)$ as functions of $d_i$  for ON and OFF states respectively.  We represent these distributions as histograms.   (The likelihoods for non-edge observations $p(e_i = 0|x_i={\rm ON}, d_i)$ and $p(e_i = 0|x_i={\rm OFF}, d_i)$ are the complements of the edge likelihoods.)

Figs.~\ref{fig:like}(c-d) show the probability $p(\theta_i|x_i, e_i = 1)$ as a function of the angular deviation $\theta_i$ for ON ($x_i=1$) and OFF ($x_i=0$) states, respectively.  For the ON state we approximate the heavy-tailed distribution as a mixture of a uniform and a Gaussian distribution (shown in red).  For the OFF state we employ a histogram representation.

Given these observations, we wish to determine the sequence of hidden states $x_1, ..., x_N$ that maximizes
\begin{multline}
p(x_1, \ldots, x_N | y_1, \ldots, y_N )\\ 
\propto p(y_1, \ldots, y_N | x_1, \ldots, x_N)p(x_1, \ldots, x_N) \label{eq:dp_first}
\end{multline}

We assume that, when conditioned on the hidden states $x_i$, the observations $y_i$ are mutually independent and independent of all $x_j$, $j\neq i$. We further assume that the hidden states are first order Markov so that Eqn. \ref{eq:dp_first} becomes
\begin{multline}
p(x_1, \ldots, x_N | y_1, \ldots, y_N )\\ \propto p(y_1|x_1)p(x_1)\prod_{i=2}^N p(y_i|x_i)p(x_i|x_{i-1})
\label{eq:dp_third}
\end{multline}
The Markov assumption implies an exponential distribution of segment lengths;  for the YorkUrbanDB training dataset we have verified that this distribution is indeed very close to exponential for segments down to $\sim15$ pixels in length.  (For smaller segments the density falls off, possibly due to difficulties in hand-labelling shorter segments.)

Table~\ref{tab:priors} shows values for the priors $p(x_1)$ and $p(x_i|x_{i-1})$, estimated from the 51 $640{\times} 480$ pixel images of the YorkUrbanDB labeled training dataset \cite{denis2008efficient}.  (Note that since the probabilities for ON and OFF states sum to 1 there are only 3 free parameters.)  We make the approximation that $p(x_i|x_{i-1})$ is independent of the variation in spacing between points on the line; since the average segment in the YorkUrbanDB generates more than 500 point samples, errors due to this approximation tend to average out.

The standard errors for these parameter estimates are relatively small, and we have verified that variation within
this range has negligible effect on results.  While these parameters are specific to the YorkUrbanDB dataset and may therefore be sub-optimal for other kinds of imagery,
they {\em can} be generalized to other image resolutions.  Assuming that the number of segments per line and their relative length are functions of the scene and not the sensor,  $p(OFF)$ and $p(ON)$ will be resolution-invariant and the probability of state changes will vary inversely with resolution.  For example, doubling the resolution to $1280{\times}960$ pixels will halve the probability of transition from OFF to ON or ON to OFF.

\begin{table}[htbp]
	\caption{Prior marginal probabilities $p\left(x_i\right)$ and conditional transition probabilities $p\left(x_i|x_{i-1}\right)$ for the hidden segment state $x_i$, derived from the  YorkUrbanDB training dataset.\vspace{1em}}
	\centering
	\begin{tabular}{|l|l|l|}
		\hline
		{\bf Parameter} & {\bf Mean} & {\bf Std. Err.} \\
		\hline
		$p(OFF)$ & 0.75 & 0.0079\\
		\hline
		$p(ON)$ & 0.25 & 0.0079\\
		\hline
		$p(OFF| OFF)$ & 0.9986 & 0.0001\\
		\hline
		$p(ON|OFF)$ & 0.0014 & 0.0001\\
		\hline
		$p(ON|ON)$ & 0.9949 & 0.0004\\
		\hline
		$p(OFF|ON)$ & 0.0051 & 0.0004\\ 
		\hline
	\end{tabular}
	\vspace{0.25cm}
	\label{tab:priors}
\end{table}

The factoring of the global probability of the line segment configuration along the line confers an optimal substructure property that allows a dynamic programming solution to the problem of finding the maximum a posteriori configuration. In particular, let the cost function $C_i(j)$ represent the minimum negative log probability of all sequences $\{x_1,\ldots,x_i\}$ ending in state $x_i=j$. Then the maximum probability sequence of states over the whole line is the sequence that minimizes \mbox{$\min_jC_N(j)$}.

Defining the cost of transitioning from state $j$ at location $i-1$ to state $k$ at location $i$ as 
\begin{multline}
c_i(j,k) = -\log\left(p(y_i|x_i=k)p(x_i=k|x_{i-1}=j)\right),\\ i=2,\ldots,N
\end{multline}
we then have that
\begin{eqnarray}
C_1(k) &=& -\log\left(p(y_1|x_1=k)p(x_1=k)\right)\\
C_i(k)&=&\min_j\left(C_{i-1}(j)+c_i(j,k)\right), i=2,\ldots,N
\end{eqnarray}

\begin{figure}[htbp]
	\centering
	\includegraphics[width=0.8\columnwidth]{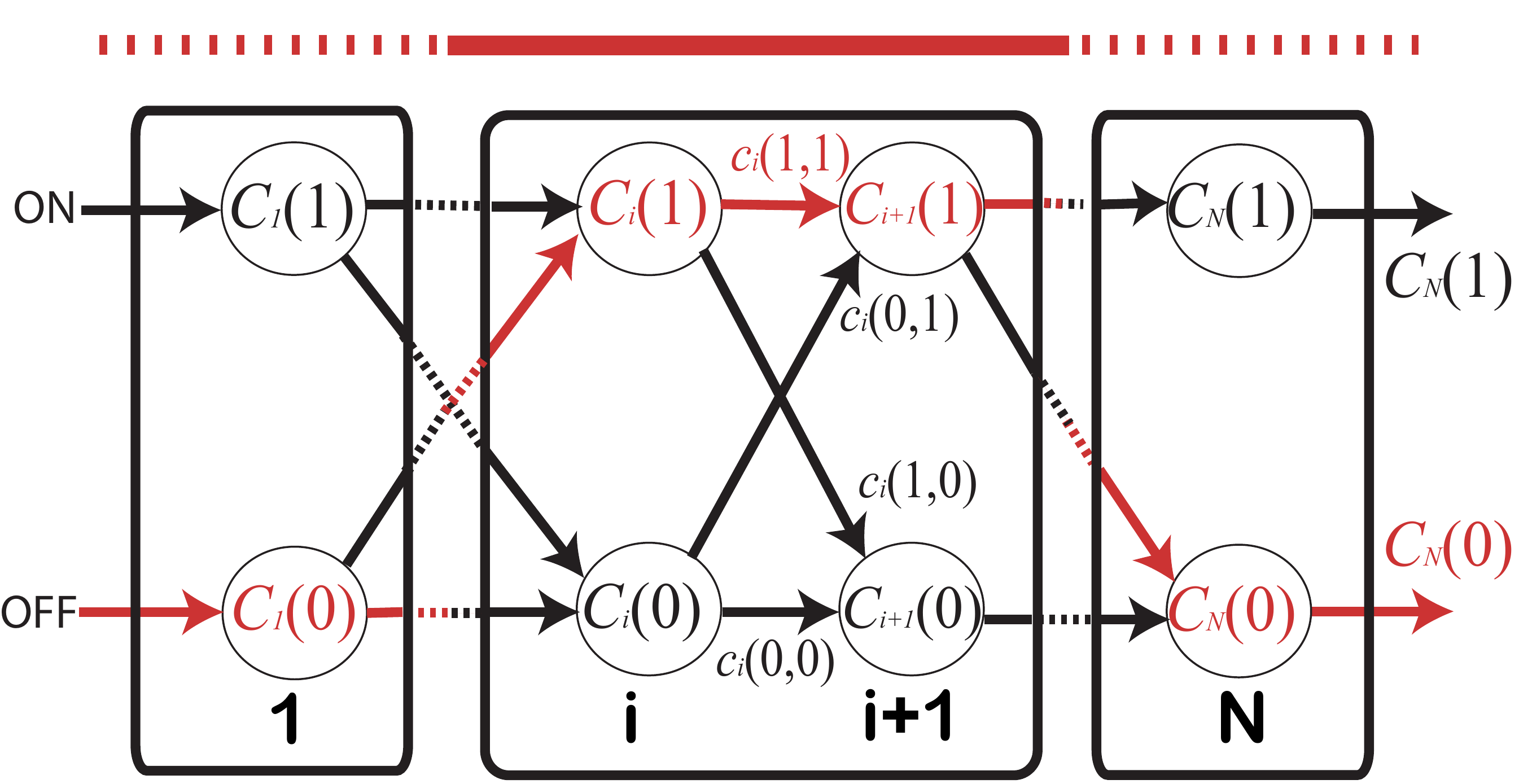}
	\caption{\label{fig:trellis} The sequence of segment state variables $x_i$ are assumed to form a Markov chain. To compute the MAP solution we build a trellis table from the first line position $i=1$ to the last line position $i=N$ that identifies the minimum cost (negative log probability) to reach either possible state (ON or OFF) at each position $i$.   The selected MAP path is shown in red, and the resulting ON/OFF states are indicated by the solid/dashed line above the trellis.}
\end{figure}

Thus the cost function $C_i(k)$ can be computed sequentially from $i = 1$ to $i = N$ in $O(N)$ time (Fig.~\ref{fig:trellis}). In order to recover the maximum probability configuration, an auxiliary data structure containing 
\begin{equation}
\hat{s}_i(k) = {\rm arg} \min_j (C_{i-1}(j) + c_i(j,k))
\end{equation} 
is maintained, allowing the maximum probability configuration to be unwound from $x_N$ back to $x_1$.

Once a line segment is detected all associated edges (i.e., edges within two pixels of the segment) are removed from the image.  This serves to reduce the incidence of multiple detections for the same segment. 

\section{Ranking}
\label{sec:ranking}
Having extracted MAP segments for each line in the image, we would like to rank their significance.  This will allow downstream applications to select only the number of segments needed to support their application, and can serve to eliminate low-ranked noise segments.
Our Markov chain model allows us to approach the ranking problem from a probabilistic perspective. In particular, we evaluate the following four probabilistic  methods for ranking a segment of length $M$ extending from position $i$ to position $i+M$:

\noindent{\bf Ranking Method 1.} Posterior probability of line segment. 
\begin{equation*}
p(x_{i \ldots i+M} = ON | y_{i \ldots i+M})
\end{equation*}
This ranking criterion will maximize the expected number of segments with no false alarms.

\noindent{\bf Ranking Method 2.} Posterior probability of line segment multiplied by length. 
\begin{equation*}
p(x_{i \ldots i+M} = ON | y_{i \ldots i+M})*M
\end{equation*}
This criterion will maximize the expected total length of segments with no false alarms.

\noindent{\bf Ranking Method 3.} Posterior odds for fully ON vs fully OFF configurations.  \begin{equation*}
\frac{p(x_{i \ldots i+M} = ON | y_{i \ldots i+M})}{p(x_{i \ldots i+M} = OFF | y_{i \ldots i+M})}
\end{equation*}

\noindent{\bf Ranking Method 4.} Sum of  marginal posterior probabilities for ON states.  The forward-backward algorithm is used to compute the posterior probability at each location.  
\begin{equation*}
\sum\limits_{j=i}^{i+M} p(x_j= ON | y_{i:i+M})
\end{equation*}
This measure reflects the expected number of ON samples on the segment, and thus will maximize the expected number of correctly labelled locations within the segment.  

\section{Evaluation Methodology}
\label{sec:eval}
It is important to evaluate line segment detection algorithms on real, complex images.  Prior evaluations have generally been qualitative (i.e., visual).  Recent efforts to quantify the evaluation require pairs of images related by a known homography, and are perhaps thus best suited for matching tasks~\cite{brown2015generalisable}.   Here we propose an alternative quantitative evaluation methodology that does not assume the existence of image pairs or known homographies and thus could be applicable for a broader range of tasks.

Our proposed evaluation method does require an image dataset in which important segments have been labelled.  Here we employ two.  The YorkUrbanDB dataset~\cite{denis2008efficient}, which consists of 102 images of urban scenes, randomly divided into training and test subsets of 51 images each.  In each image, major line segments that conform to one of the three so-called Manhattan directions~\cite{coughlan2003manhattan} (i.e., vertical or horizontal and conforming to the main directions of orthogonal walls, streets, etc.) have been identified and labelled by hand.  This database has been used widely to train and evaluate algorithms for vanishing point detection~\cite{tardif2009non}, line detection~\cite{barinova2012detection} and Manhattan frame estimation~\cite{denis2008efficient,tal2013accurate}.  We also evaluate on the more recent and much larger Wireframe dataset~\cite{wireframe_cvpr18}, which consists of 5,462 images (5,000 for training, 462 for test) of man-made scenes.  

We assume that the line segment detector under evaluation returns a list of line segments in ranked order.    We sample each ground truth and detector segment uniformly with a sample spacing of one pixel and use these point samples to evaluate the detector as a function of the number $k$ of top-ranked segments selected, varying $k$ from 10 to 500.  

For each value of $k$ we first identify potential point matches as those (ground truth, detector) point pairs lying within a threshold distance of $2\sqrt{2}$ pixels of each other.  
This threshold was selected to associate any pair of lines that could potentially appear in the image with less than a one-pixel intervening gap.
We then sort these candidate matches by Euclidean distance and accept matches in greedy fashion starting with the smallest distance, matching each point at most once, and thus arriving at a near-optimal bipartite match.   Having associated ground truth and detector points, we employ the Hungarian algorithm~\cite{kuhn1955hungarian} to identify the optimal bipartite match between ground truth and detector segments that maximizes the total number of points matched.  

Now that we have a 1:1 association between ground truth and detector segments, it remains to evaluate the quality of this association.  We propose three 
evaluative measures.

\subsection{Recall as a Function of the Number of Segments}    
We can compute a measure of recall as the number of ground truth point samples matched to detector samples, divided by the total number of ground truth point samples.  
This measure of recall is problematic if we allow matches without regard to the segments on which the points lie, as it does not penalize under-segmentation (joining multiple short segments into a single long segment) or over-segmentation (breaking up a long segment into multiple short segments).

However, constraining matches to lie on 1:1 associated segments solves both of these problems.  In the case of under-segmentation, only one of the shorter ground truth segments
is matched, leading to a high penalty.  In the case of over-segmentation, only one of the detector segments is matched, again generating a high penalty.

Without additional constraints, using recall by itself is still problematic, as it is biased toward detectors that report a larger number of segments, thereby maximizing the probability of detecting ground truth points.  We address this by comparing recall as a function of the same number $k$ of segments  reported.

\subsection{Recall as a Function of Total Segment Length}  There is still a potential bias in this recall-vs-$k$ measure.  Neglecting co-linear ground truth segments, the method can be biased toward detectors that report segments of maximal length (i.e., global lines) as this minimizes the risk of missing ground truth points.  To address this potential bias, our second performance measure reports recall as a function of the sum $L$ of the lengths of detected segments.  This severely penalizes detectors that report over-long segments.

\subsection{Precision-Recall}  Our third and final performance measure is conventional precision-recall.   We can take as a measure of precision the number of ground truth point samples matched to detector samples, divided by the total number of detector point samples.   Again, by enforcing a 1:1 matching at the segment level, both under-segmentation and over-segmentation are penalized.

To facilitate future comparisons, the code that performs these evaluations as well as the code for the MCMLSD algorithm is available online at \url{elderlab.yorku.ca/resources}.  

\subsection{Limitations of Precision-Based Measures of Performance}

Since the YorkUrbanDB dataset does not provide a complete labelling of all segments in an image, detection of a segment that is not in the dataset does not necessarily represent an error.  For this reason, the absolute precision values reported here should be interpreted with caution.   Nevertheless, since the segments labelled in the YorkUrbanDB dataset are highly-visible Manhattan features projecting from prominent structures in the scene,  it is reasonable to expect a superior detector to rank these highly, and therefore attain higher {\em relative} precision values compared to inferior detectors.

The creators of the Wireframe dataset attempted to label `all the line
segments associated with the scene structures'.  Unlike the YorkUrbanDB dataset, these are not restricted to Manhattan lines, and so one expects the dataset to contain a more complete labelling, potentially allowing for higher precision.  However, the authors also avoided labelling line segments in what they considered `texture'.  This includes straight line segments projecting from regular tiling and cladding patterns on horizontal and vertical surfaces, which are very common in the built environment.  Since these can be quite useful in establishing surface orientation for both human and computer vision systems, for many applications one would want a line segment detector to detect these, yet such detectors will tend to generate lower precision on the Wireframe dataset.

Given the limitations of precision measures for these two datasets, we feel it is important to consider multiple different measures of performance when evaluating and comparing algorithms, and so we report performance using all three measures in what follows.  

\section{Results}
\subsection{Ranking}
Our first goal is to evaluate the four candidate ranking methods discussed in Section \ref{sec:ranking}.
Using the default Hough resolution recommended by Tal \& Elder~\cite{tal2013accurate} ($\Delta\rho= 0.2$ pixels,  $\Delta\theta=0.1$ deg), we find that 
our MCMLSD algorithm generates an average of 414 lines and 488 line segments for each $640{\times}480$-pixel image of the YorkUrbanDB training dataset.   Note that not  all lines generate a segment and some generate several segments.  

\begin{figure}[htbp] 
	\centering
	\subfloat[Method 1]{\includegraphics[width=0.23\columnwidth]{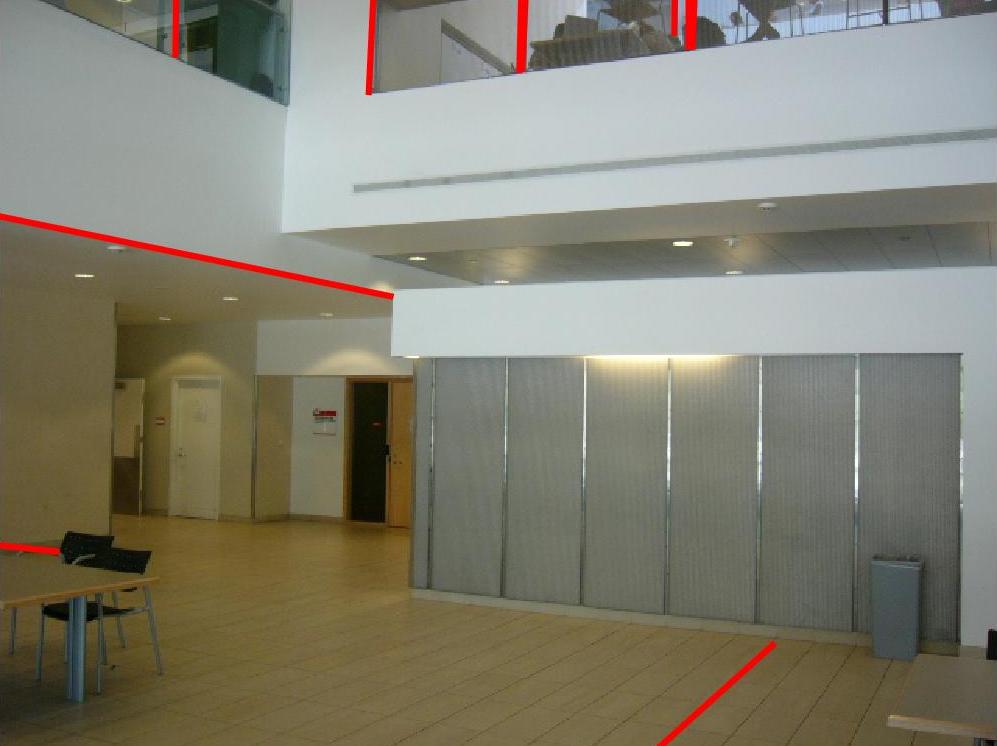}} \hspace{0.005\columnwidth}
	\subfloat[Method 2]{\includegraphics[width=0.23\columnwidth]{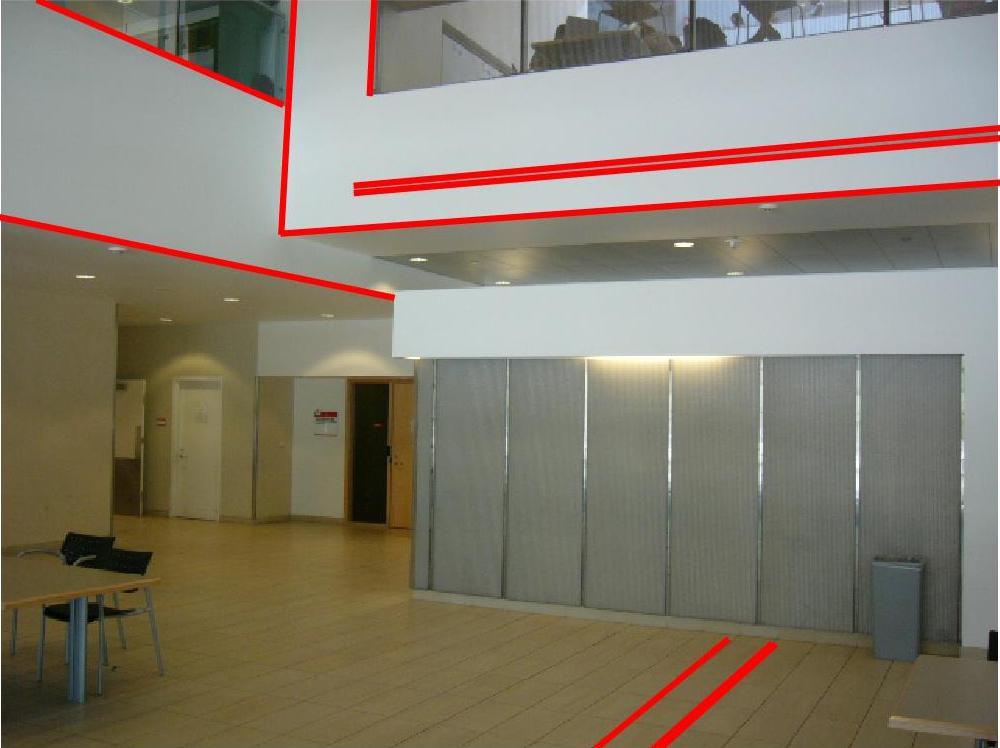}} \hspace{0.005\columnwidth}
	\subfloat[Method 3]{\includegraphics[width=0.23\columnwidth]{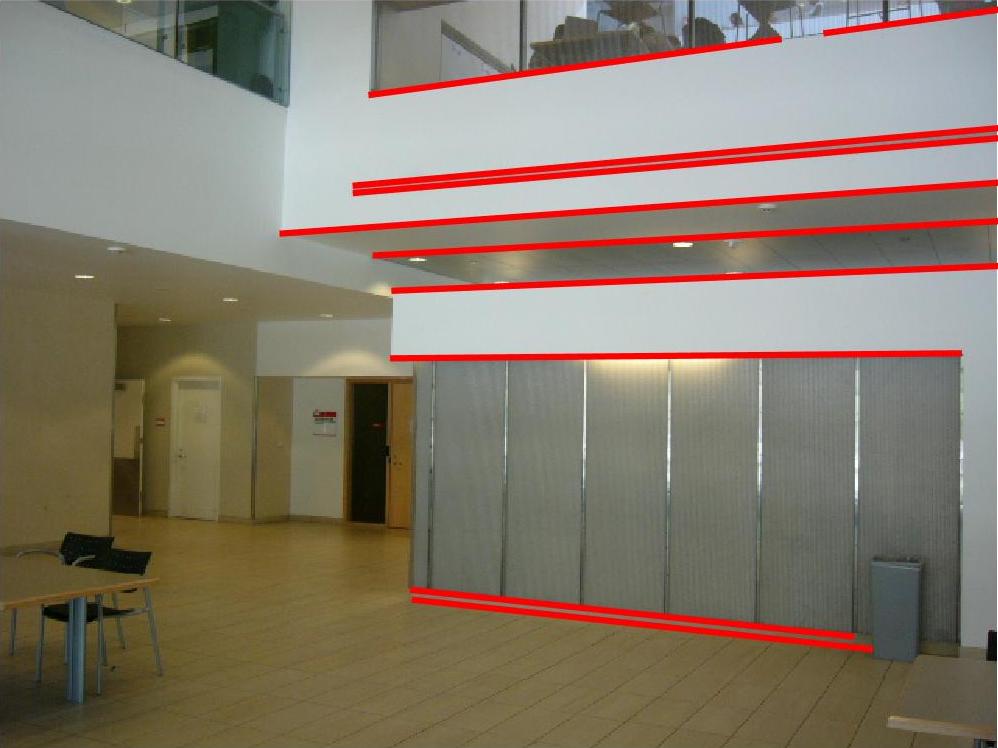}} \hspace{0.005\columnwidth}
	\subfloat[Method 4]{\includegraphics[width=0.23\columnwidth]{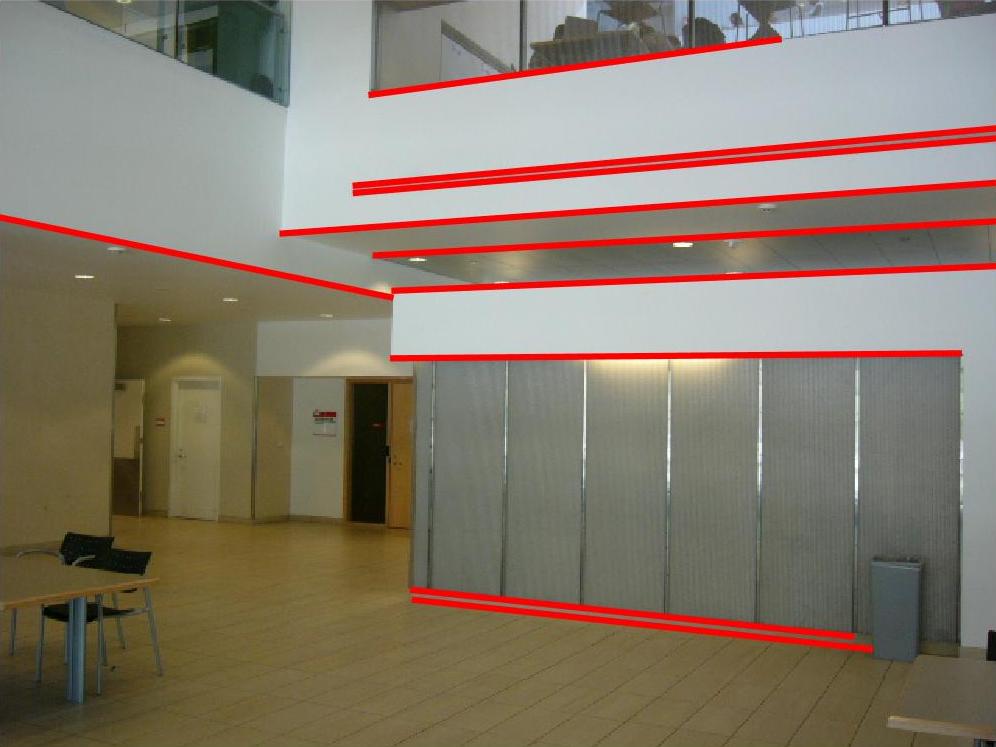}} 
	\caption{10 top-ranked segments for four ranking methods on example image.}
	\label{fig:rankings}
\end{figure}

Fig.~\ref{fig:rankings} shows the 10 top-ranked segments produced by each of our four ranking methods on an example image.  We find that the multiplicative nature of the first criterion favours short high-confidence segments.   This problem can be addressed by multiplying by segment length (Method 2), forming a contrast between purely ON and purely OFF configurations (Method 3), or summing the ON point marginals (Method 4) to estimate the number of correctly labeled points.  

\begin{figure}[htbp]
	\centering
	\includegraphics[width=0.75\columnwidth]{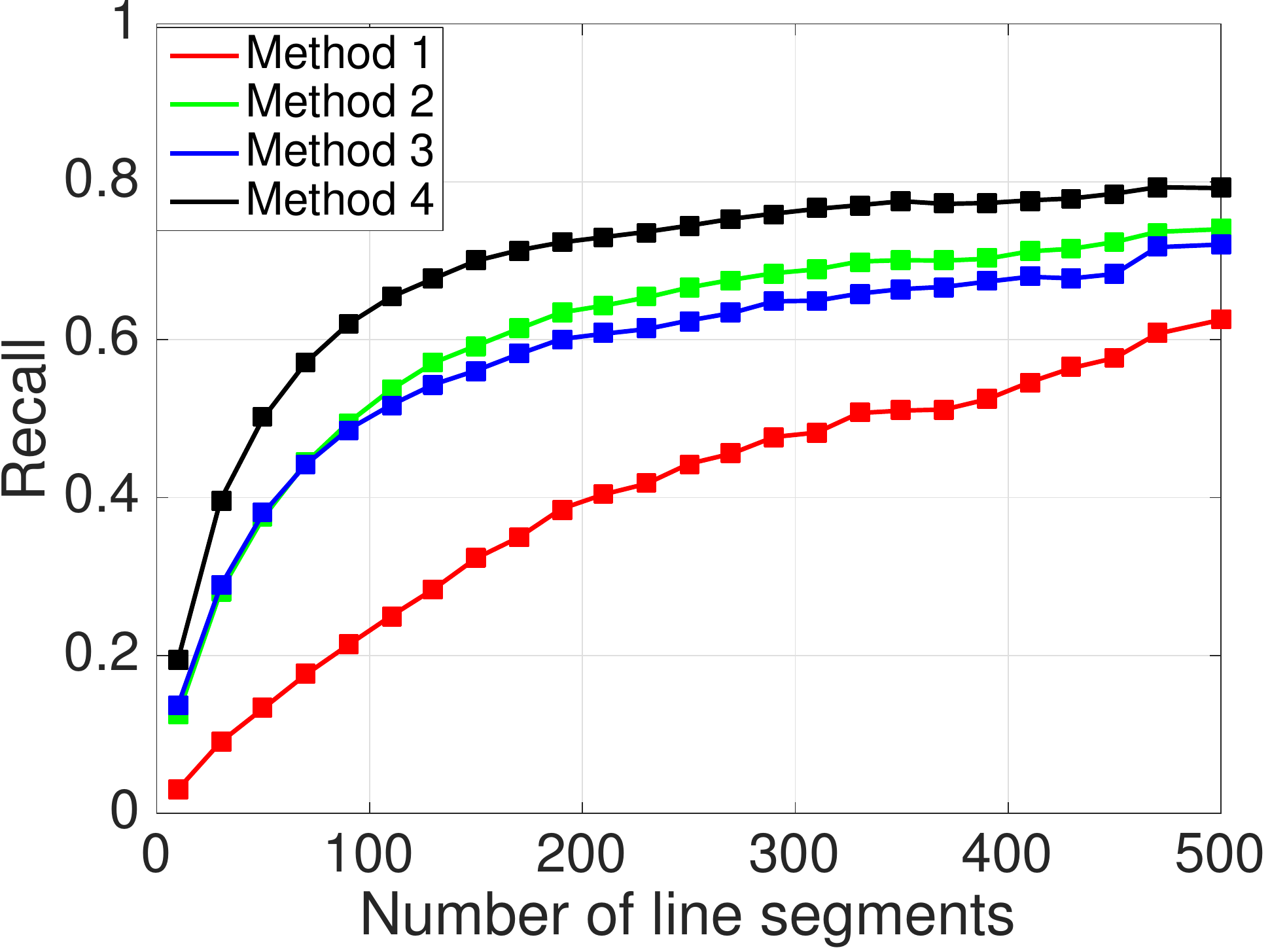}
	\caption{Performance of the four ranking methods described in section \ref{sec:ranking}, as measured by recall vs number of segments returned, on the YorkUrbanDB training dataset.}
	\label{fig:eval_ranks} 
\end{figure}

Figure \ref{fig:eval_ranks} shows the recall for each of these ranking methods as a function of the number of segments returned, on the YorkUrbanDB training dataset.  The bias toward shorter segments leads to poor recall for Method 1.  Methods 2-4 yield much better results and in the sequel we adopt Method 4 as our ranking method of choice, given its superior performance and intuitive probabilistic interpretation.  We call the resulting algorithm the Markov Chain Marginal Line Segment Detector (MCMLSD) to capture the importance of the Markov chain model of the line as well as the probabilistic ranking that maximizes the expected number of correctly labelled points on the segment. 

\subsection{Hough Resolution}
\begin{figure*}[htbp]
	\centering
	\subfloat[]{\includegraphics[width=0.3\textwidth]{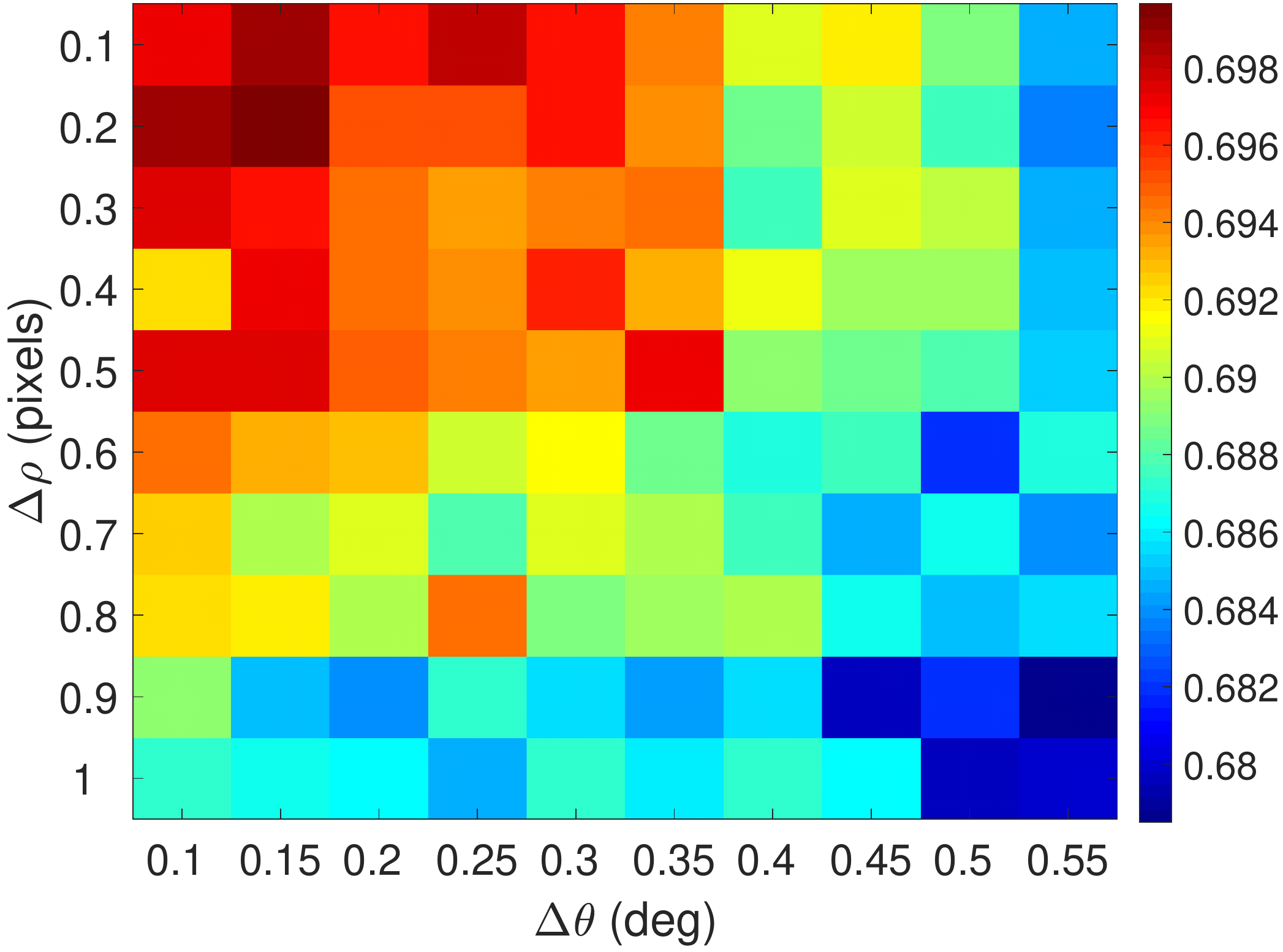}}
	\hspace{5mm}
	\subfloat[]{\includegraphics[width=0.3\textwidth]{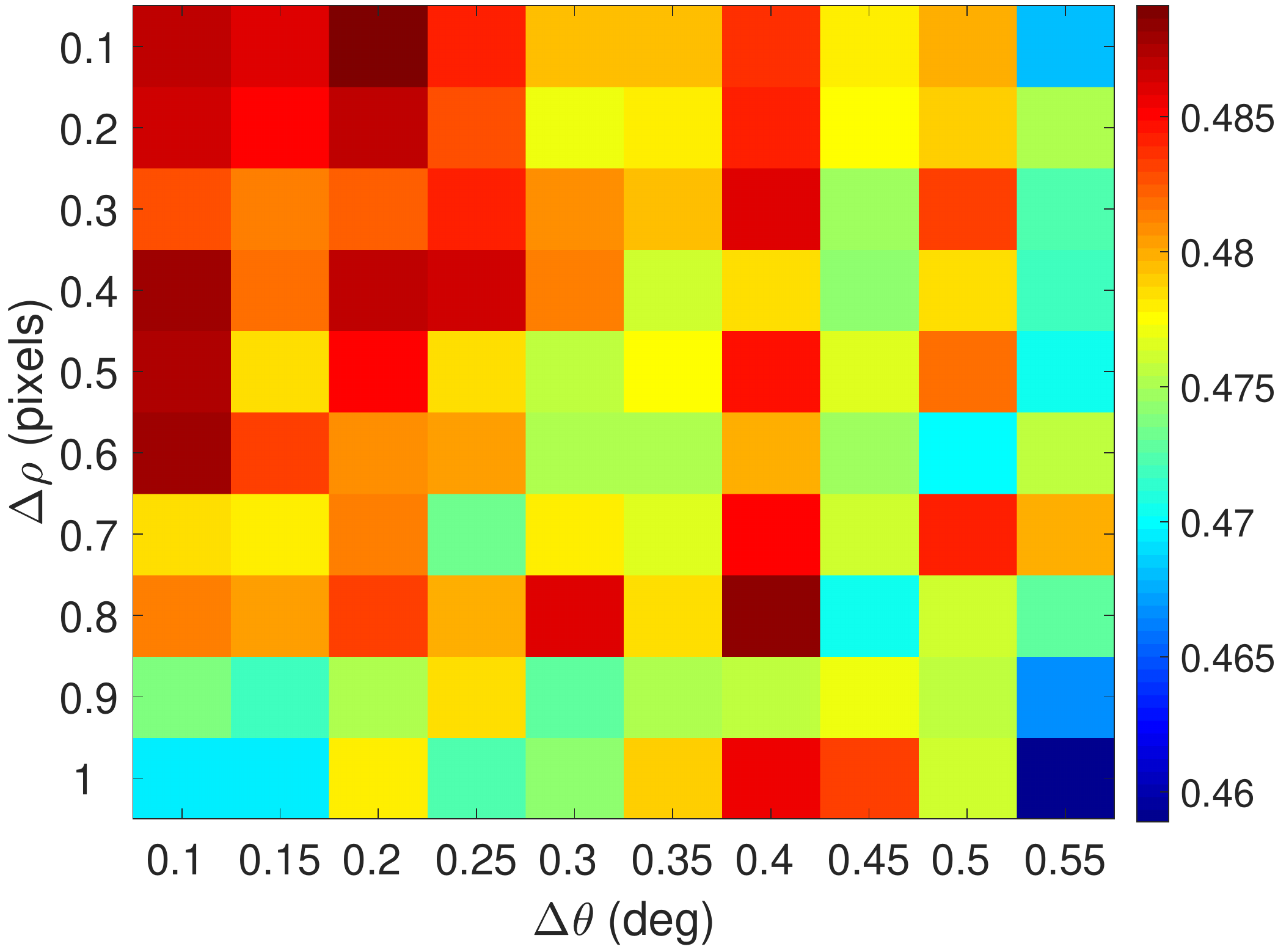}}\
	\hspace{5mm}
	\subfloat[]{\includegraphics[width=0.3\textwidth]{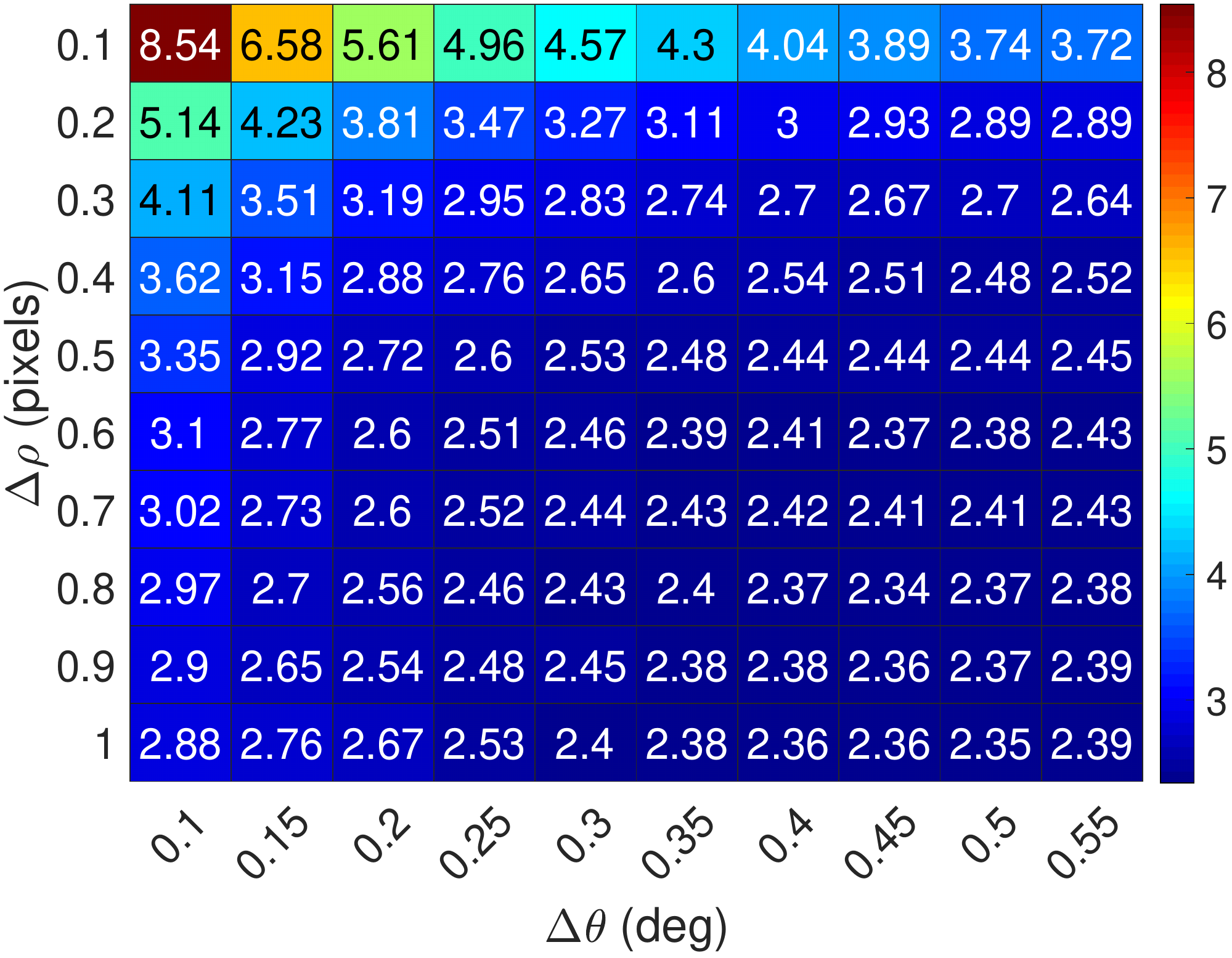}}
	\caption{Performance and run-time analysis of Hough map resolution.  (a) Mean recall over number of segments $k=1,\ldots,500$ returned.  (b) Mean precision over number of segments $k=1,\ldots,30$ returned. (c) Corresponding run time of MCMLSD algorithm per image (sec).
	}
	\label{fig:houghGridSearch} 
\end{figure*}

Having selected the ranking method, we fine-tune the Hough resolution parameters 
($\Delta\rho, \Delta\theta$) on the YorkUrbanDB training data,  computing recall for the top 100 lines over a range of parameter
values and then using kernel regression with bandwidths selected by leave-one-out cross-validation
to generate a smooth objective surface (Figure \ref{fig:houghGridSearch}). The optimal parameter values using this approach were found to be $\Delta\rho$= 0.4 pixels and $\Delta\theta$= 0.46 deg.  We adopt these parameter values for all subsequent experiments on both the YorkUrbanDB and Wireframe datasets.

\subsection{Algorithms Evaluated}
We compare the proposed MCMLSD method against five leading methods:\footnote{
In the MCMLSD conference paper~\cite{almazen2017dynamic}, we also compared against the 
Progressive Probabilistic Hough Transform (PPHT) method of Matas et al.~\cite{matas2000robust}.  
We have removed this comparison as the PPHT algorithm has not proven to be competitive with more recent methods and there seems to be quite a diversity in the literature on how exactly is it implemented and parameterized.  We add here comparison to the Wireframe Parser~\cite{wireframe_cvpr18} and Attraction Field~\cite{xue2018learning} methods, which were  published after the MCMLSD conference paper was published.}
\begin{enumerate}
\item The slice sampling weighted mean shift (SSWMS) method of Nieto et al.~\cite{nieto2011line}
\item The widely-used line segment detector (LSD) method of Grompone von Gioi et al.~\cite{von2008lsd}
\item The linelet-based method (linelet) of Nam-Gyu et al.~\cite{7926451}
\item The deep-learning Wireframe Parser method of Huang et al.~\cite{wireframe_cvpr18} 
\item The deep-learning Attraction Field method of Xue et al.~\cite{xue2018learning}, with {\em a-trous} architecture  
\end{enumerate}

\noindent{\bf SSWMS.} We obtained the code for the SSWMS method from \url{sourceforge.net/projects/lswms}. (The authors renamed the method LSWMS there.)  There are two parameters - we used the author-recommended default values for both:
\begin{itemize}
\item orientation threshold $\Delta\theta=22.5$ deg 
\item mean shift bandwidth $=$ 3 pixels
\end{itemize}
The SSWMS algorithm  is designed to output segments in descending order of salience - we therefore use this order to rank the segments.  

\noindent{\bf LSD.} We obtained the code for LSD from \url{www.ipol.im/pub/art/2012/gjmr-lsd/}.  We rank segments using the criterion recommended by the authors and employed in later work~\cite{brown2015generalisable}, namely in increasing order of the number of expected false alarms, which is one of the outputs of the LSD detector.   

\noindent{\bf Linelet.} We obtained the code for the Linelet\cite{7926451} algorithm from \url{https://github.com/NamgyuCho/Linelet-code-and-YorkUrban-LineSegment-DB}. We rank segments using the criterion recommended by the authors.

\noindent{\bf Wireframe Parser.} 
Xue et al.~\cite{xue2018learning} provide the segments generated by the Wireframe Parser for both YorkUrbanDB and Wireframe test sets; we use these to compute performance.  Since the authors do not specify a ranking method, we rank the segments in descending order of segment length.

\noindent{\bf Attraction Field.} 
As for the Wireframe Parser, Xue et al.~\cite{xue2018learning} provide the segments generated by the Attraction Field method for both YorkUrbanDB and Wireframe test sets, so we use these to compute performance.  We rank the segments using the criterion recommended by the authors.

\subsection{Qualitative Results}
\begin{figure*}[htbp]
	\label{fig:lsdvsour}
	\centering
	\begin{tabular}{c c c c c}
		
		
	\raisebox{3\normalbaselineskip}[0pt][0pt]{\rotatebox[origin=c]{90}{SSWMS}}&
		{\includegraphics[width=0.225\textwidth]{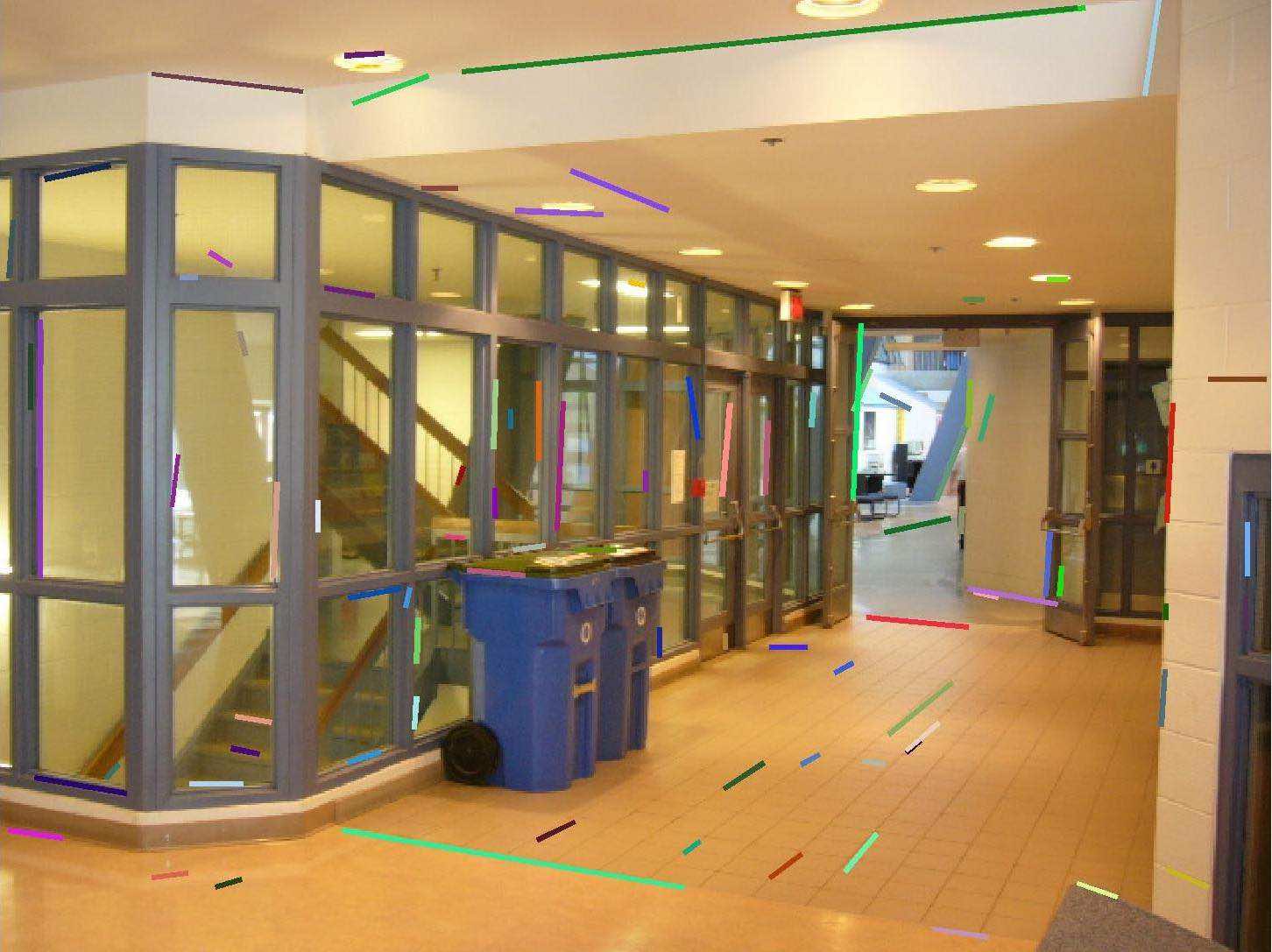}}&
		{\includegraphics[width=0.225\textwidth]{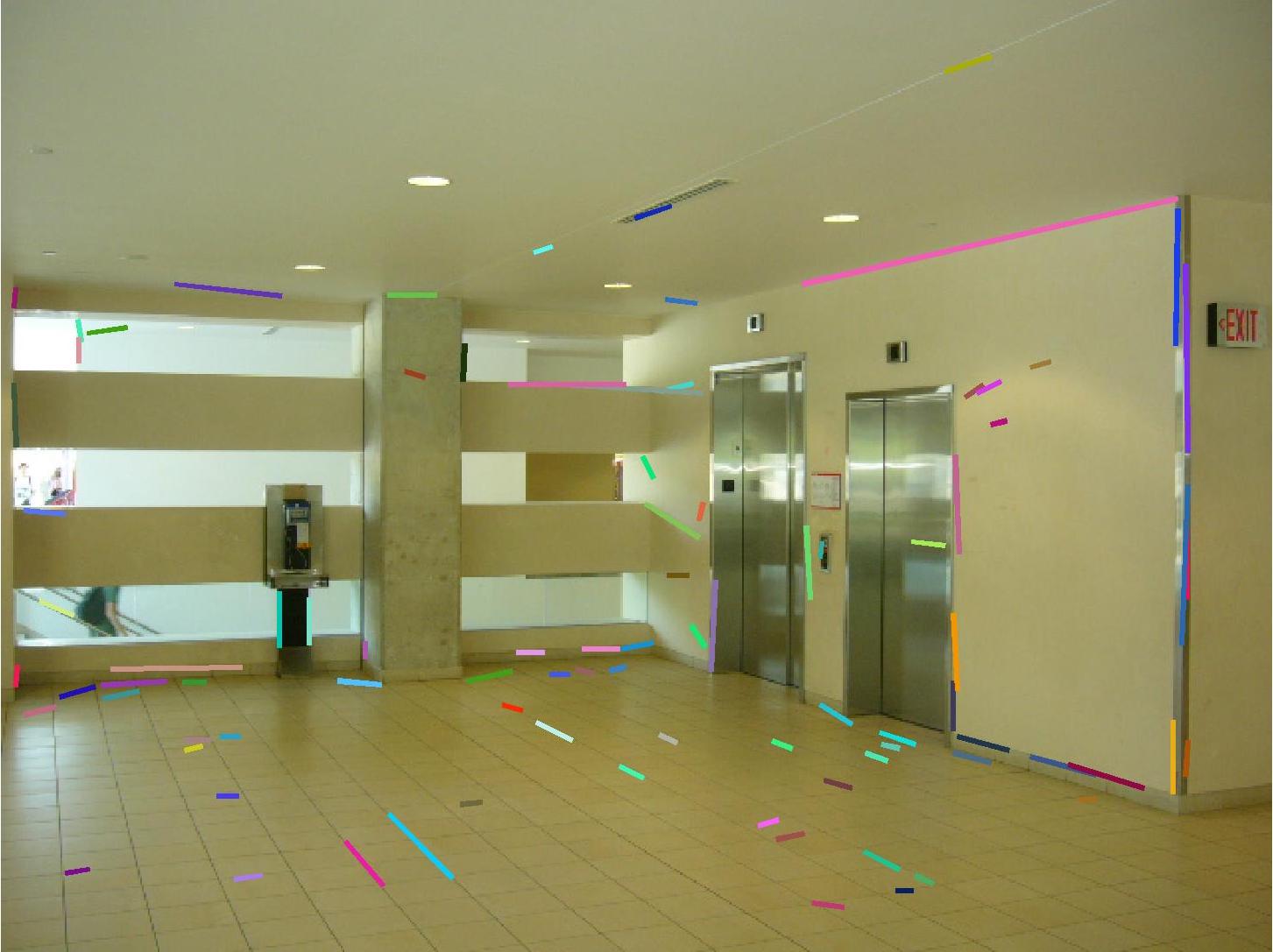}}&
		{\includegraphics[width=0.225\textwidth]{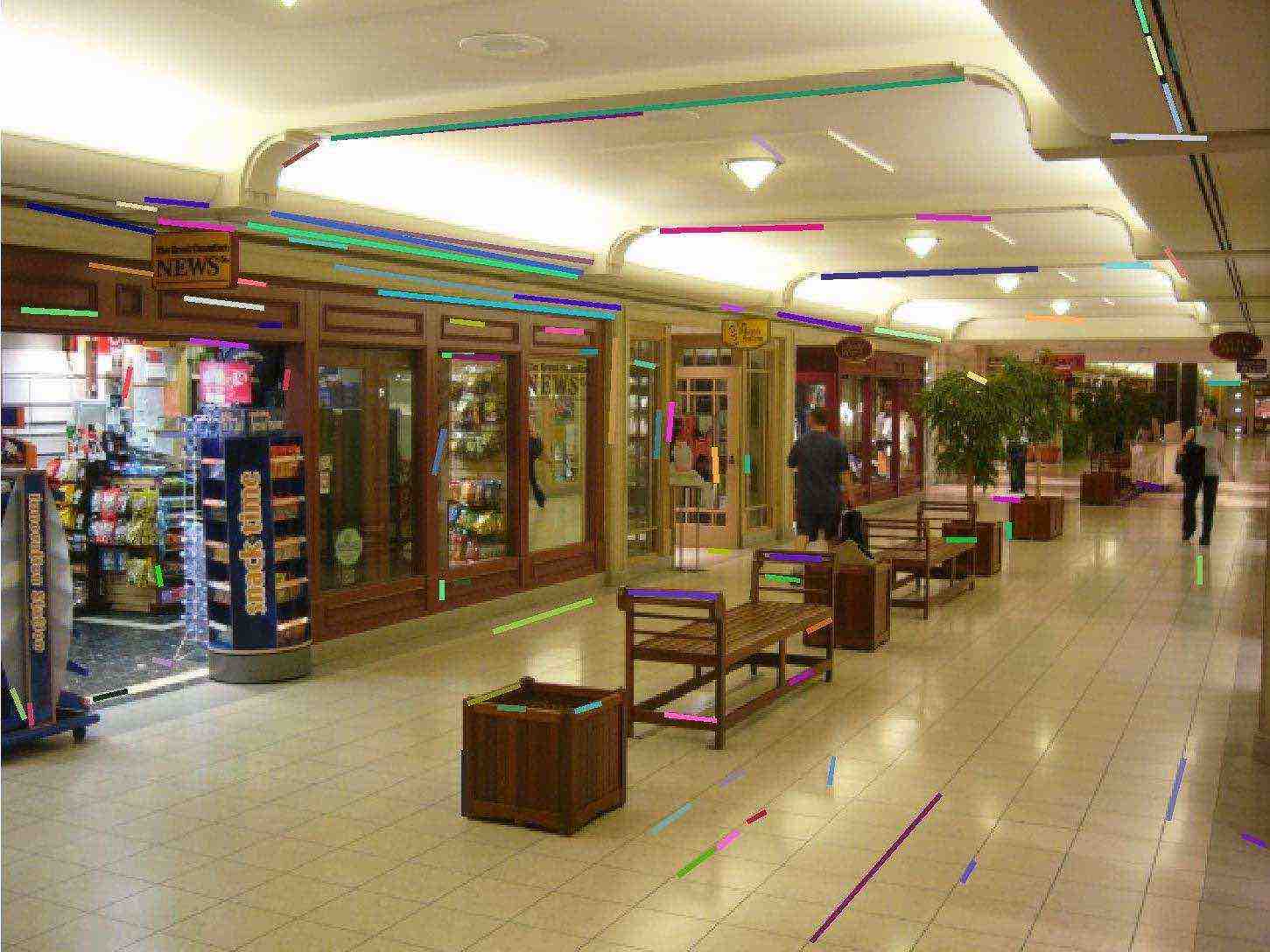}}&
		{\includegraphics[width=0.225\textwidth]{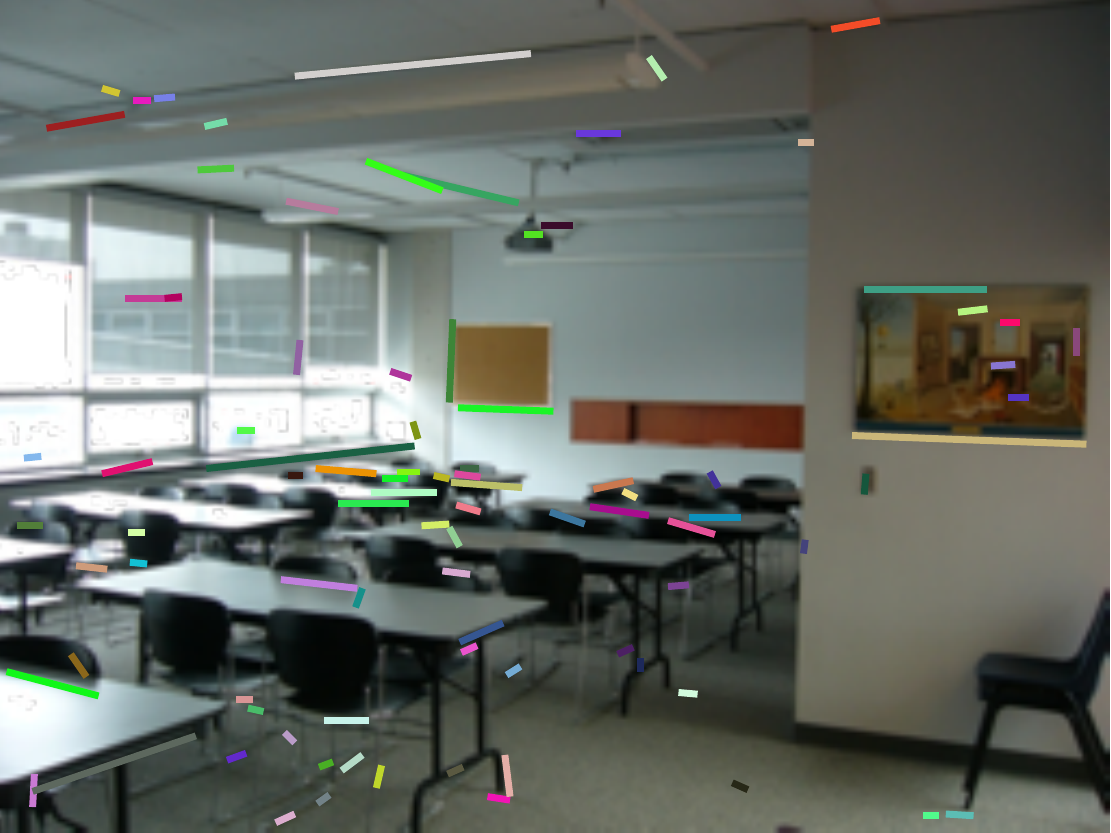}}\\
		
		\raisebox{3\normalbaselineskip}[0pt][0pt]{\rotatebox[origin=c]{90}{LSD}}&
		{\includegraphics[width=0.225\textwidth]{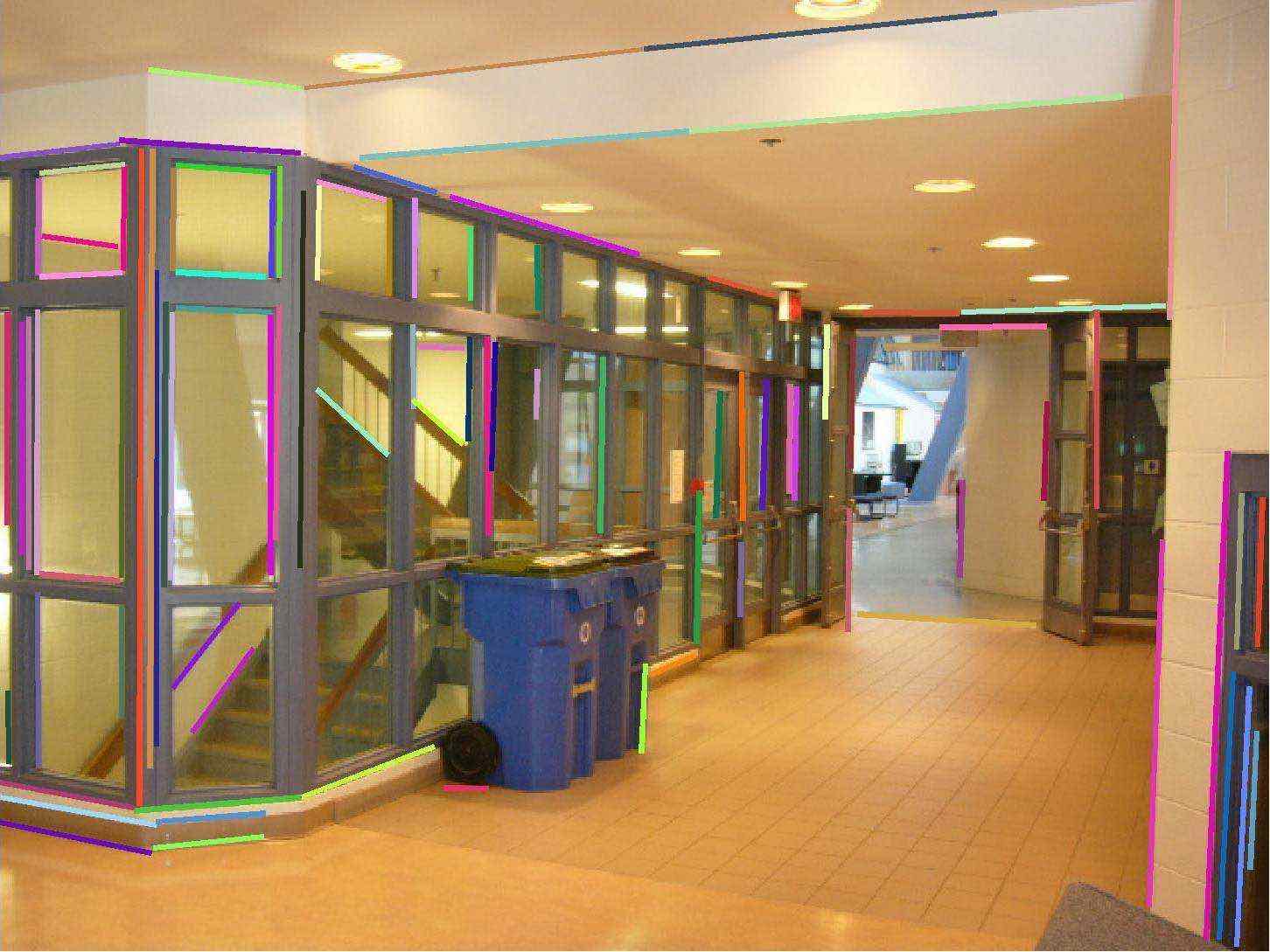}}&
		{\includegraphics[width=0.225\textwidth]{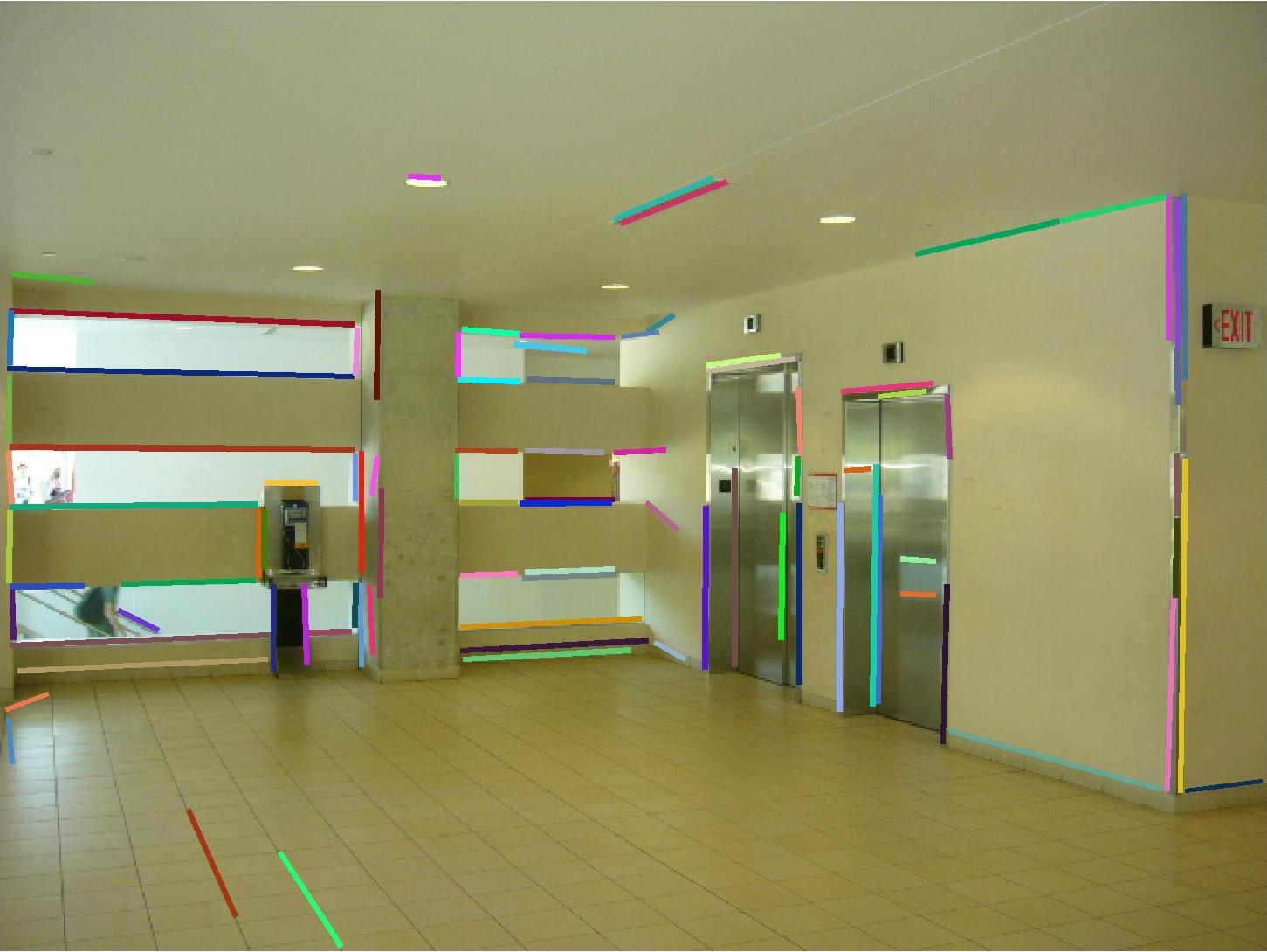}}&
		{\includegraphics[width=0.225\textwidth]{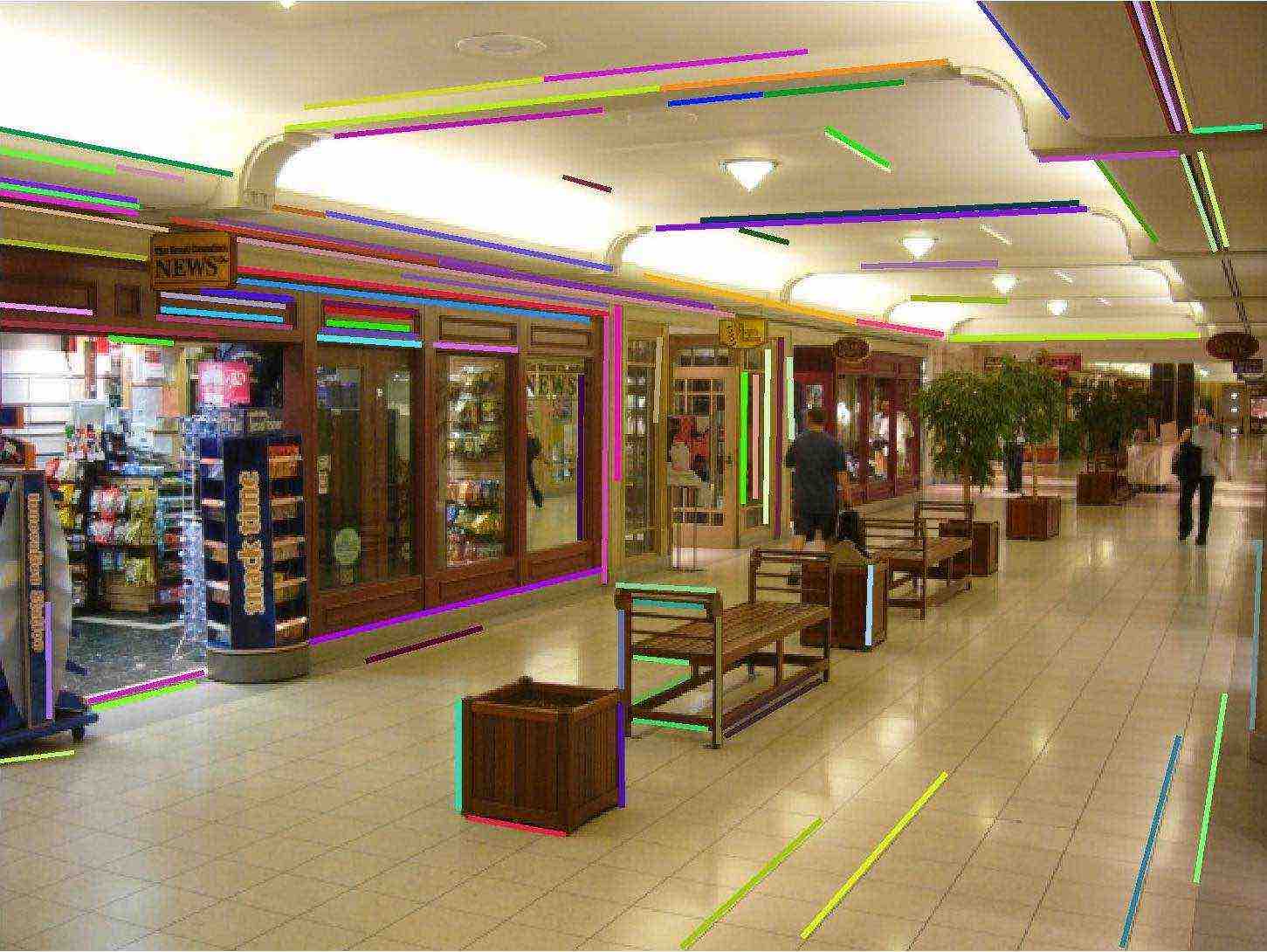}}&
		{\includegraphics[width=0.225\textwidth]{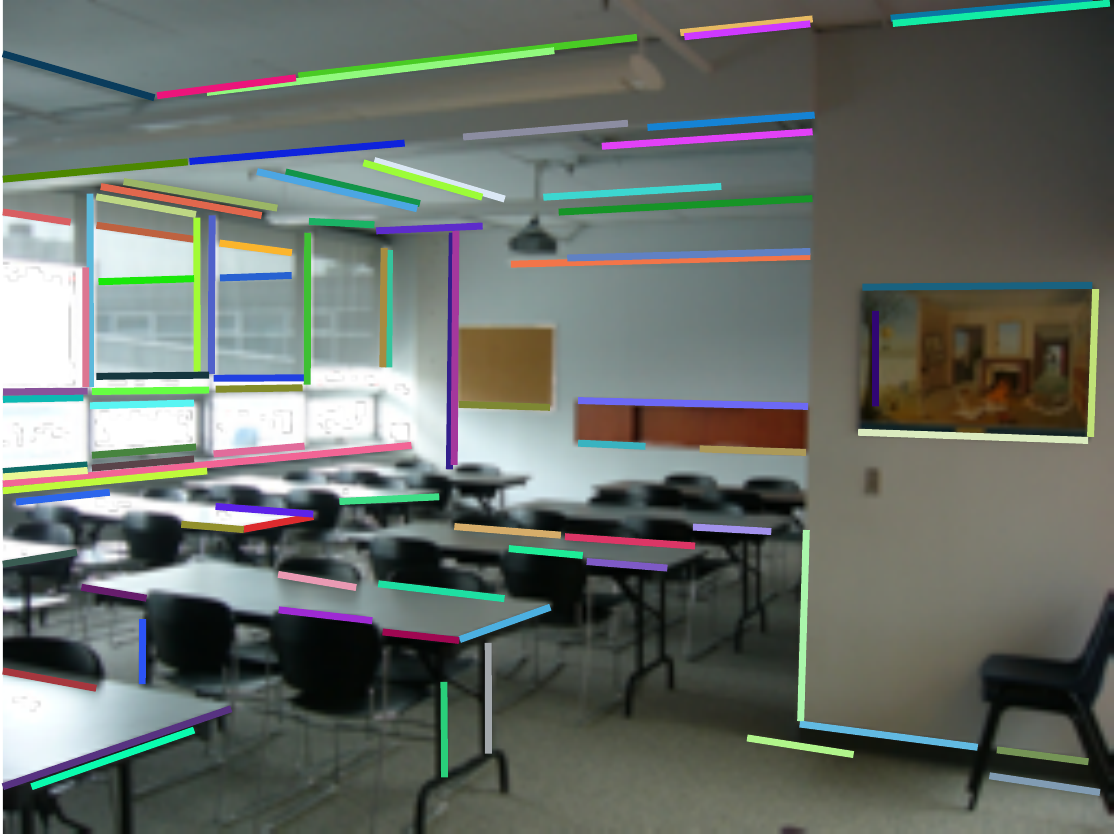}}\\
		
		\raisebox{3\normalbaselineskip}[0pt][0pt]{\rotatebox[origin=c]{90}{Linelet}}&
		{\includegraphics[width=0.225\textwidth]{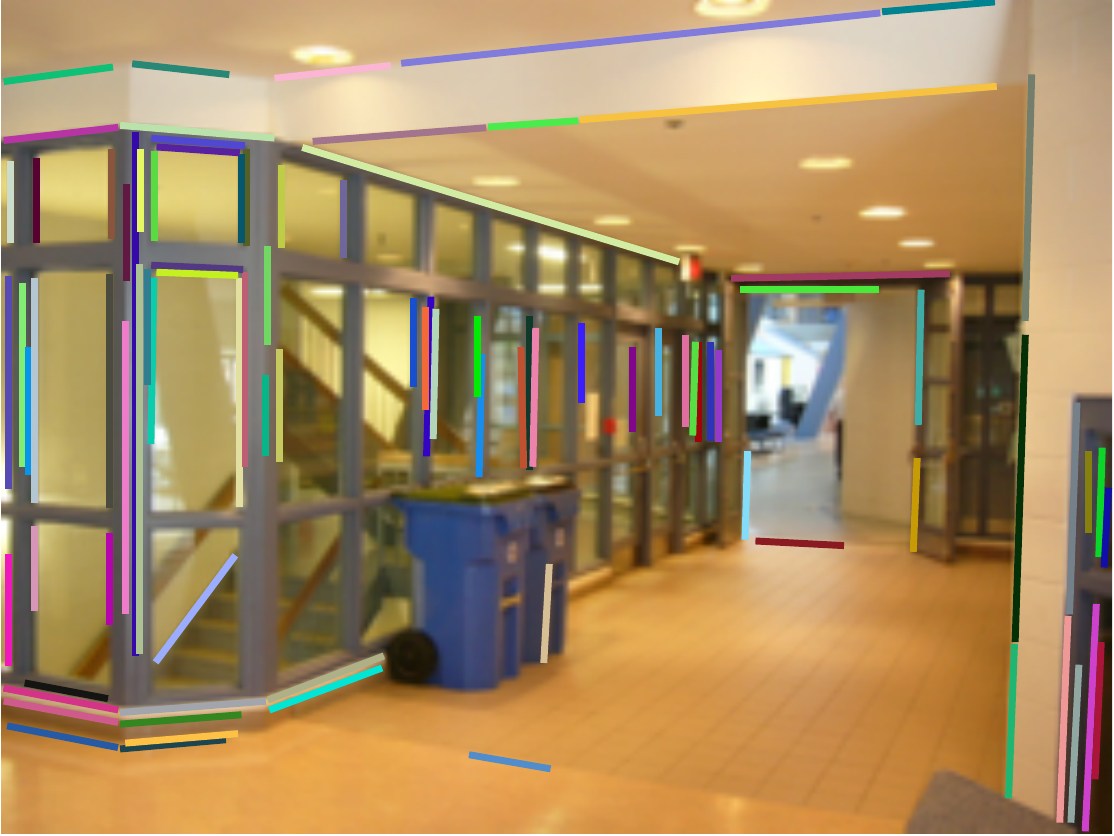}}&
		{\includegraphics[width=0.225\textwidth]{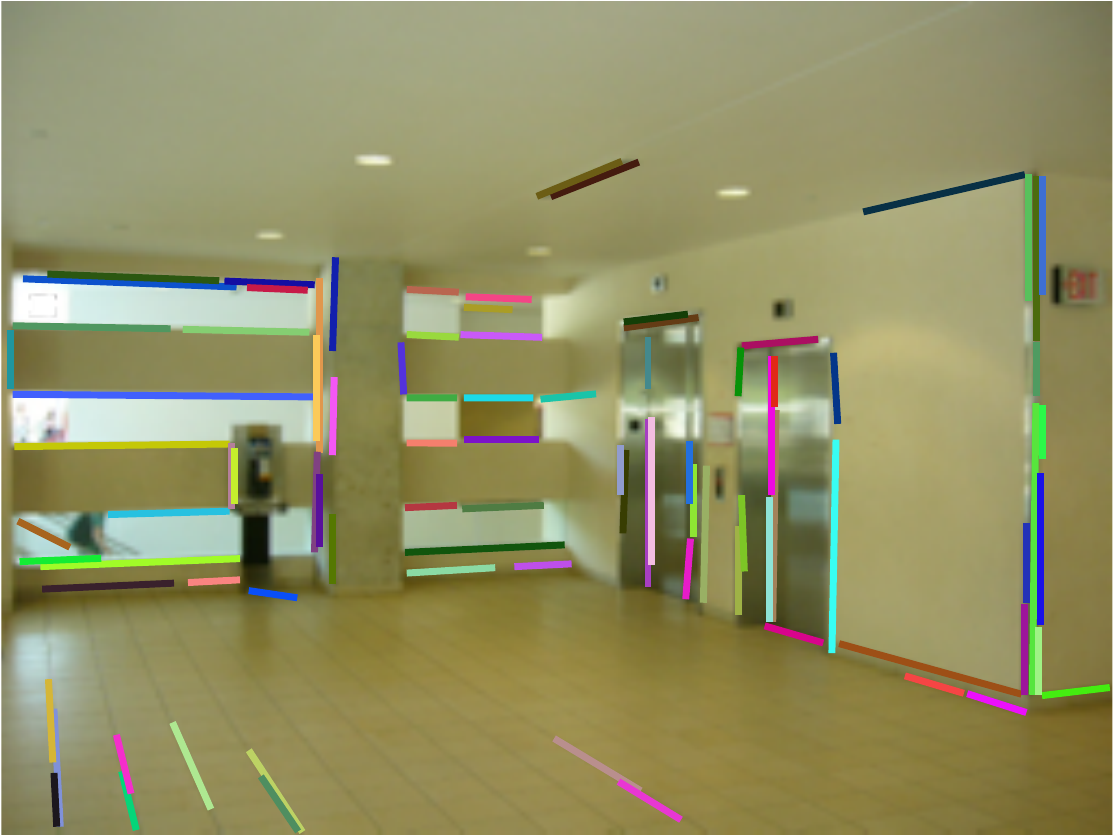}}&
		{\includegraphics[width=0.225\textwidth]{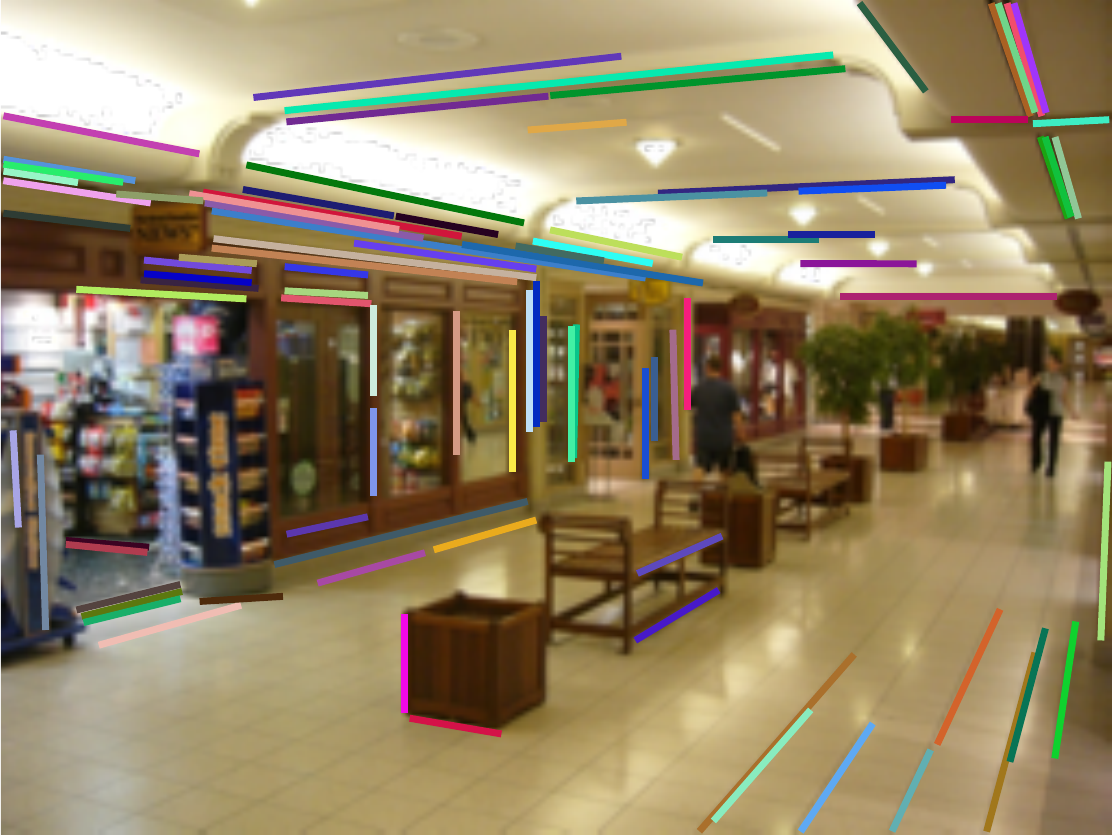}}& 
		{\includegraphics[width=0.225\textwidth]{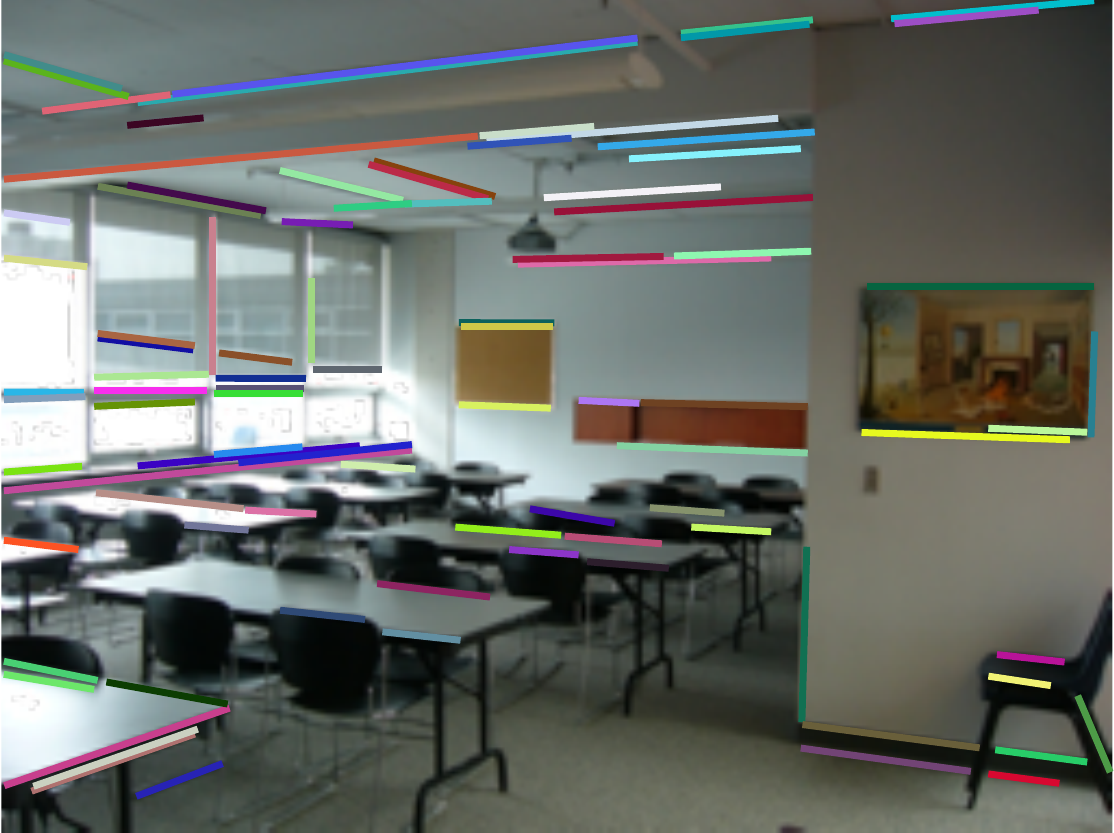}}\\
		
		\raisebox{3\normalbaselineskip}[0pt][0pt]{\rotatebox[origin=c]{90}{Wireframe}}&
		{\includegraphics[width=0.225\textwidth]{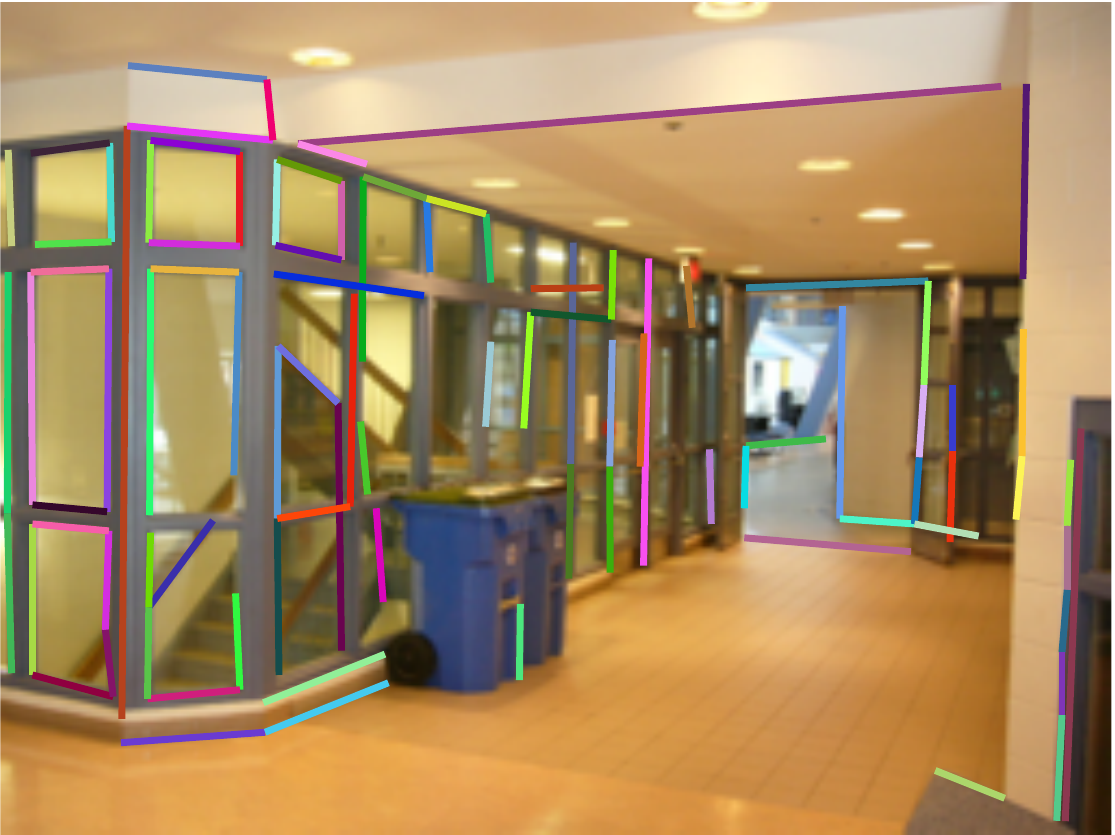}}&
		{\includegraphics[width=0.225\textwidth]{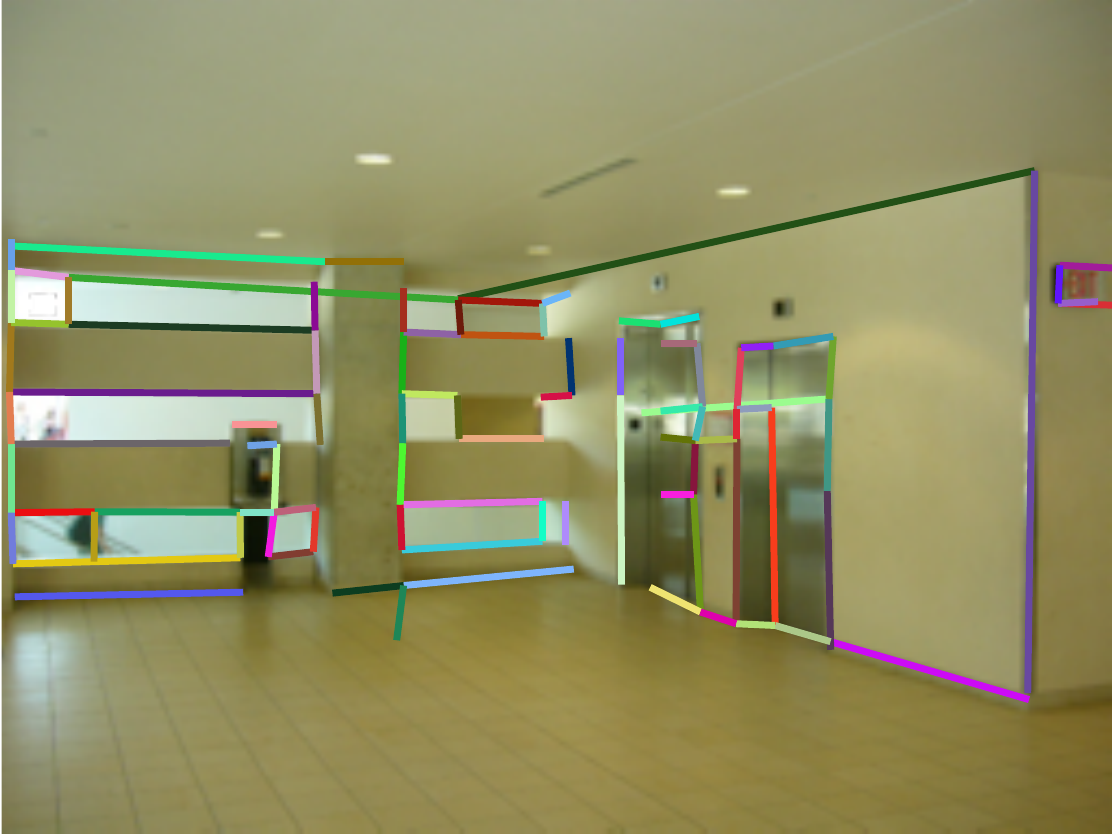}}&
		{\includegraphics[width=0.225\textwidth]{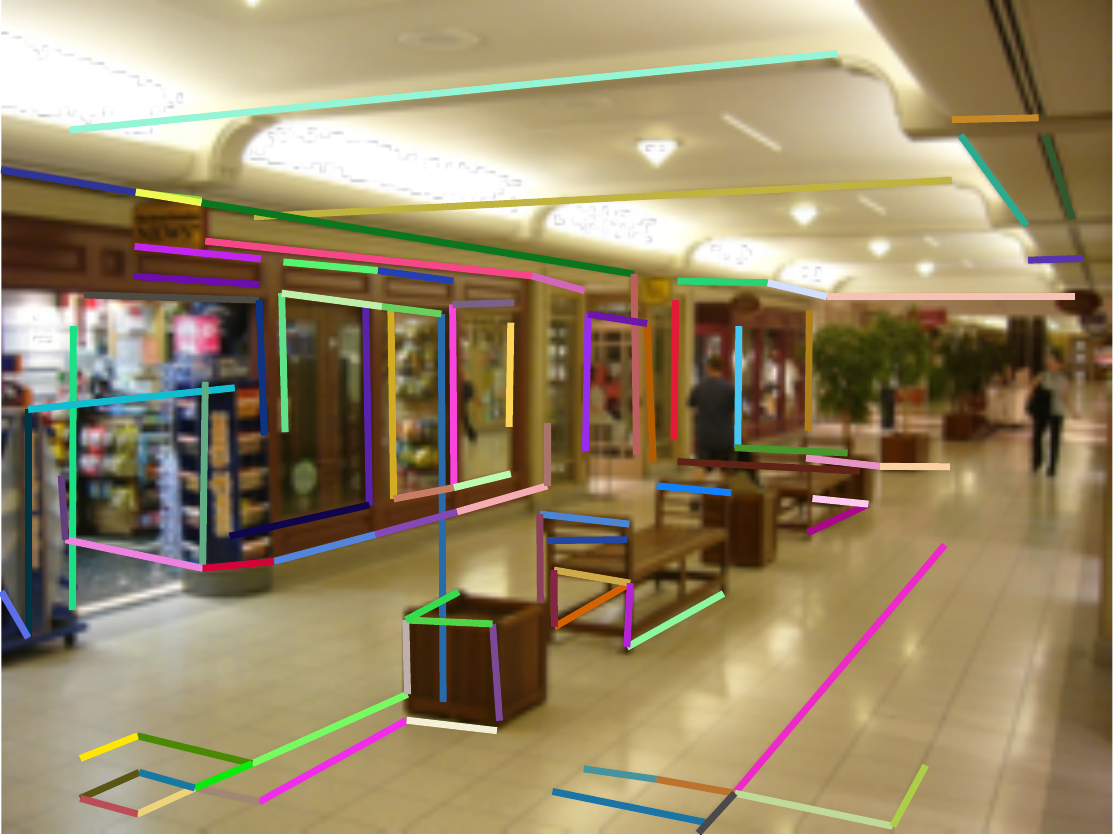}}&
		{\includegraphics[width=0.225\textwidth]{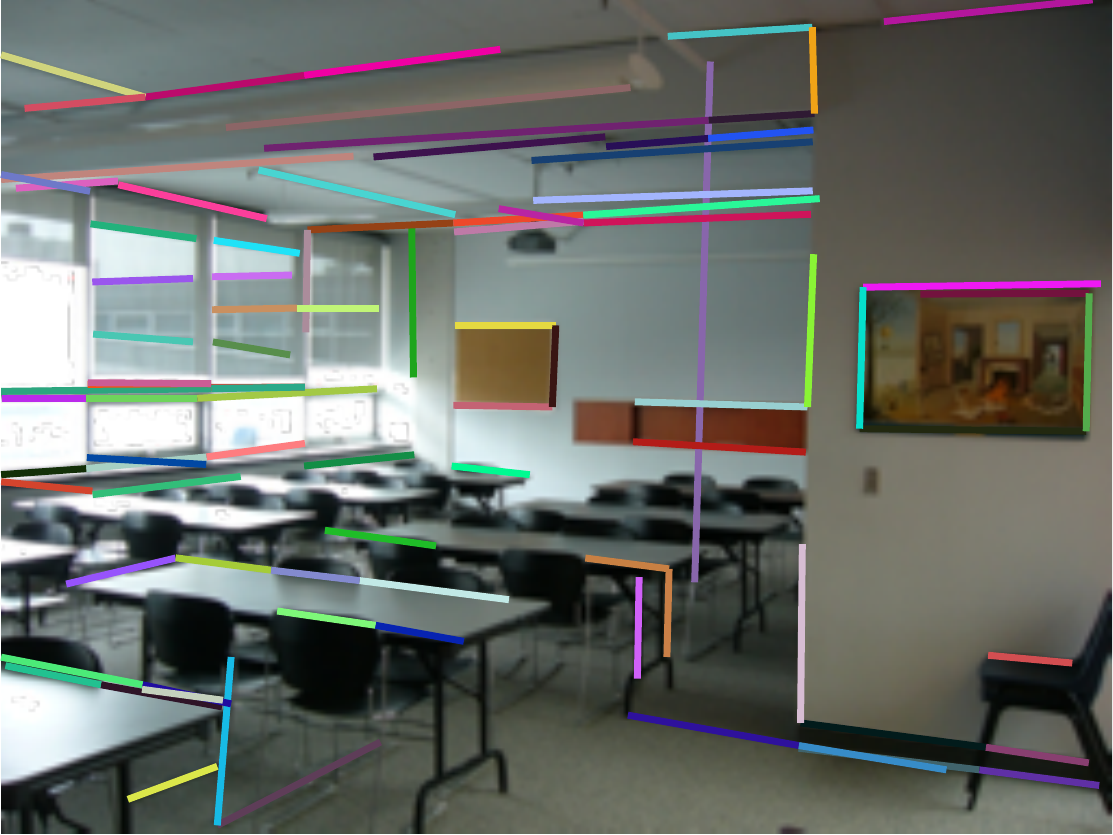}}\\
		
		\raisebox{3\normalbaselineskip}[0pt][0pt]{\rotatebox[origin=c]{90}{Attraction Field}}&
		{\includegraphics[width=0.225\textwidth]{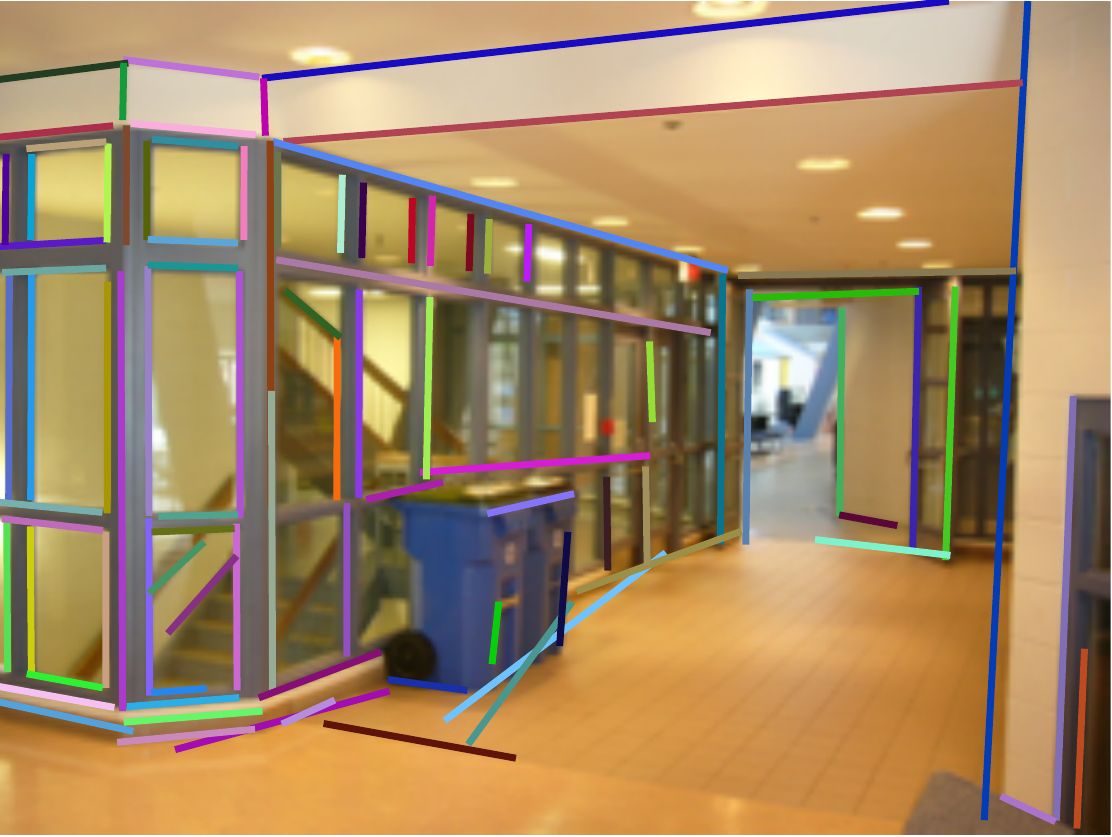}}&
		{\includegraphics[width=0.225\textwidth]{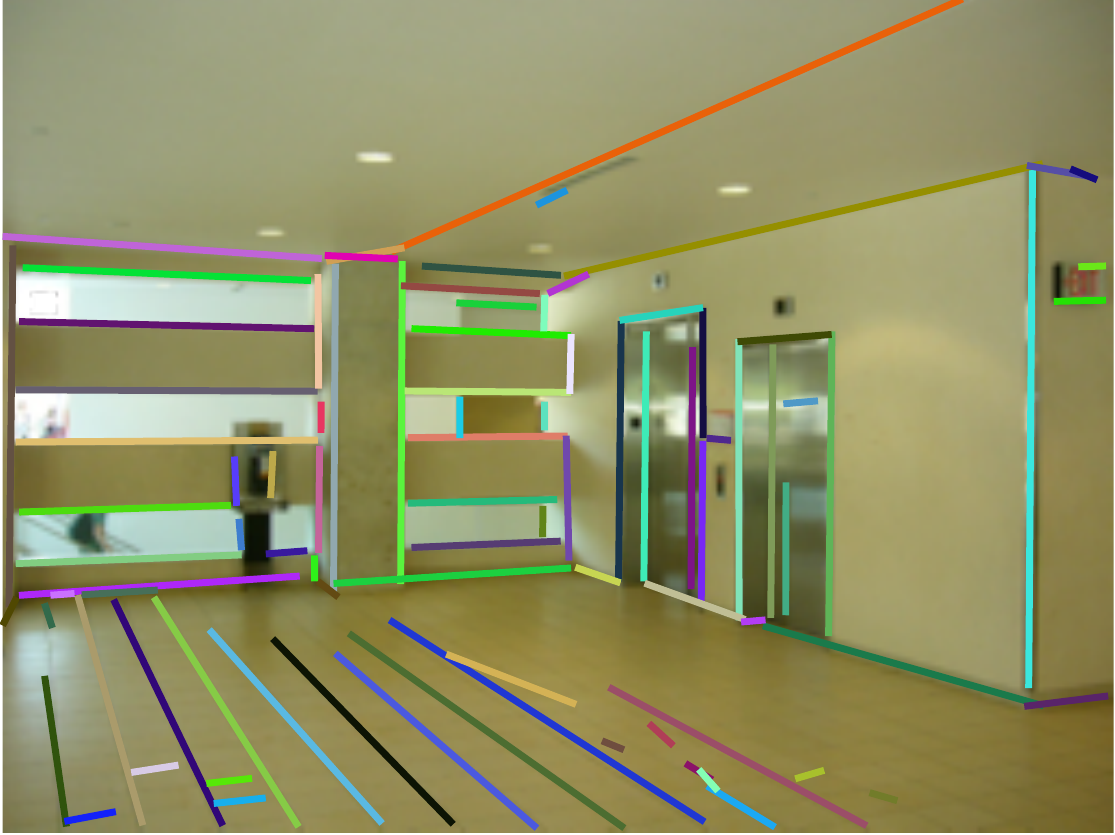}}&
		{\includegraphics[width=0.225\textwidth]{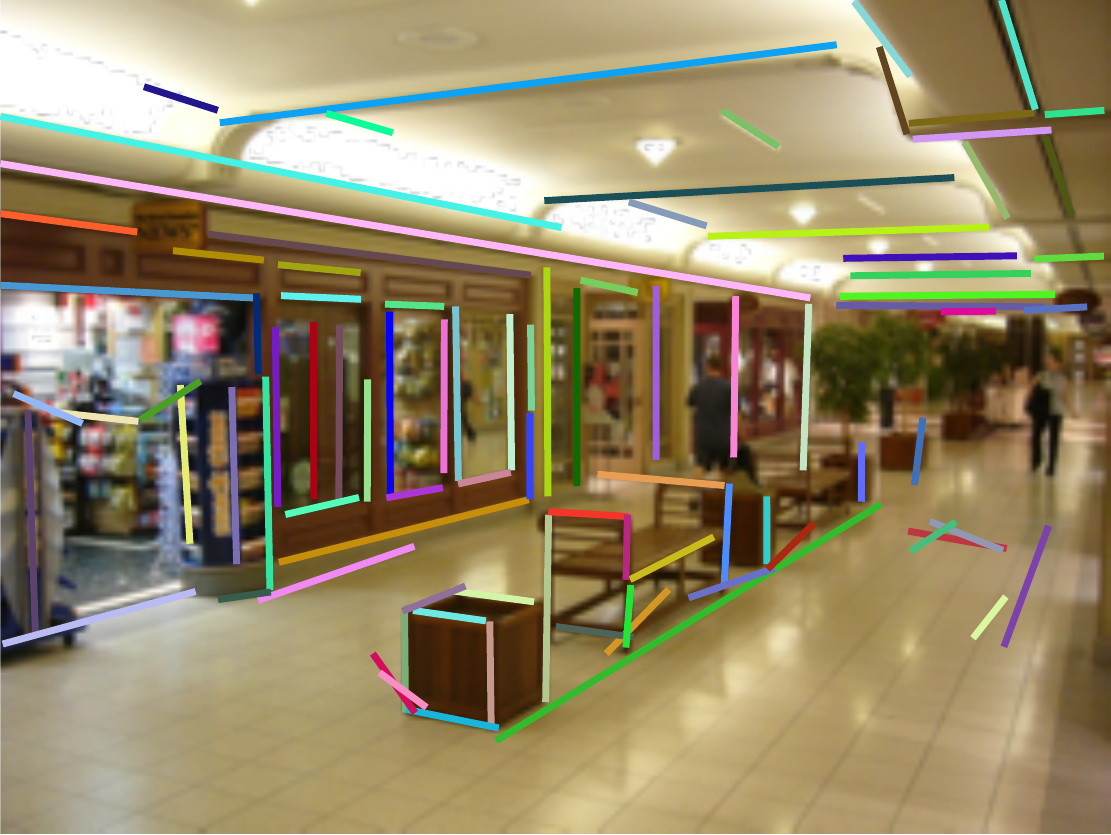}}& 
		{\includegraphics[width=0.225\textwidth]{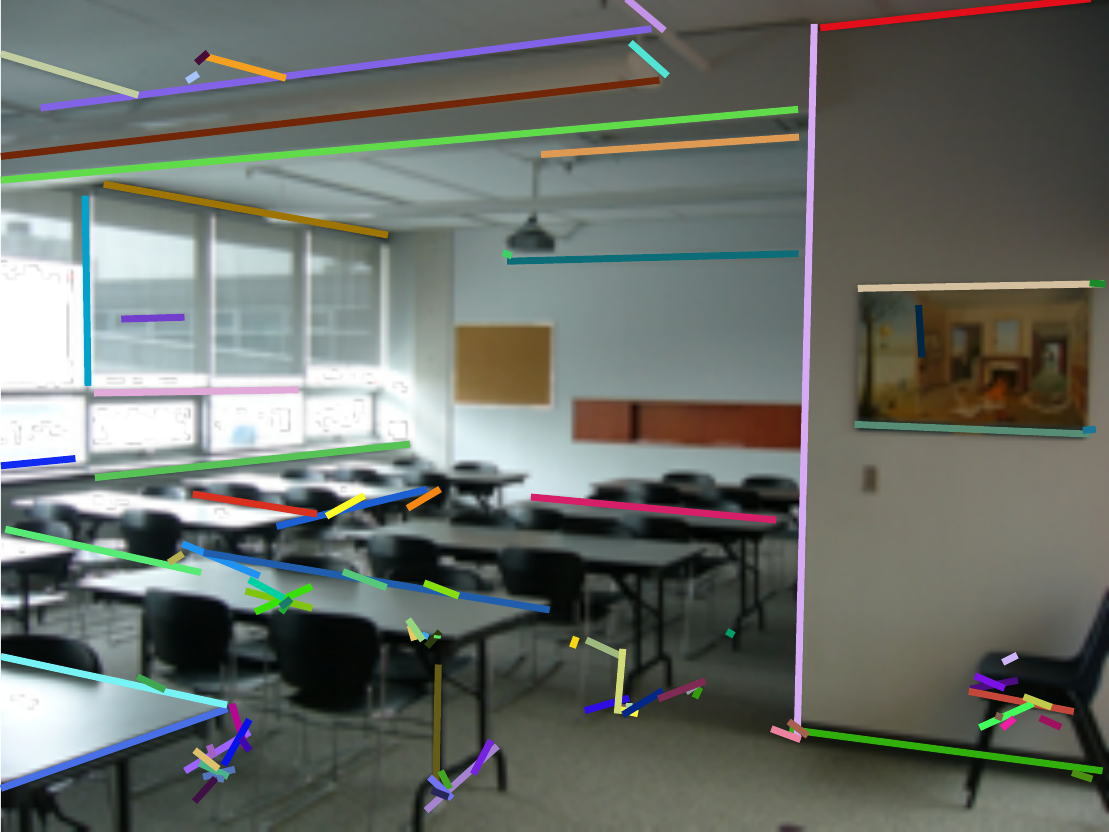}}\\
		
		\raisebox{3\normalbaselineskip}[0pt][0pt]{\rotatebox[origin=c]{90}{MCMLSD}}&
		{\includegraphics[width=0.225\textwidth]{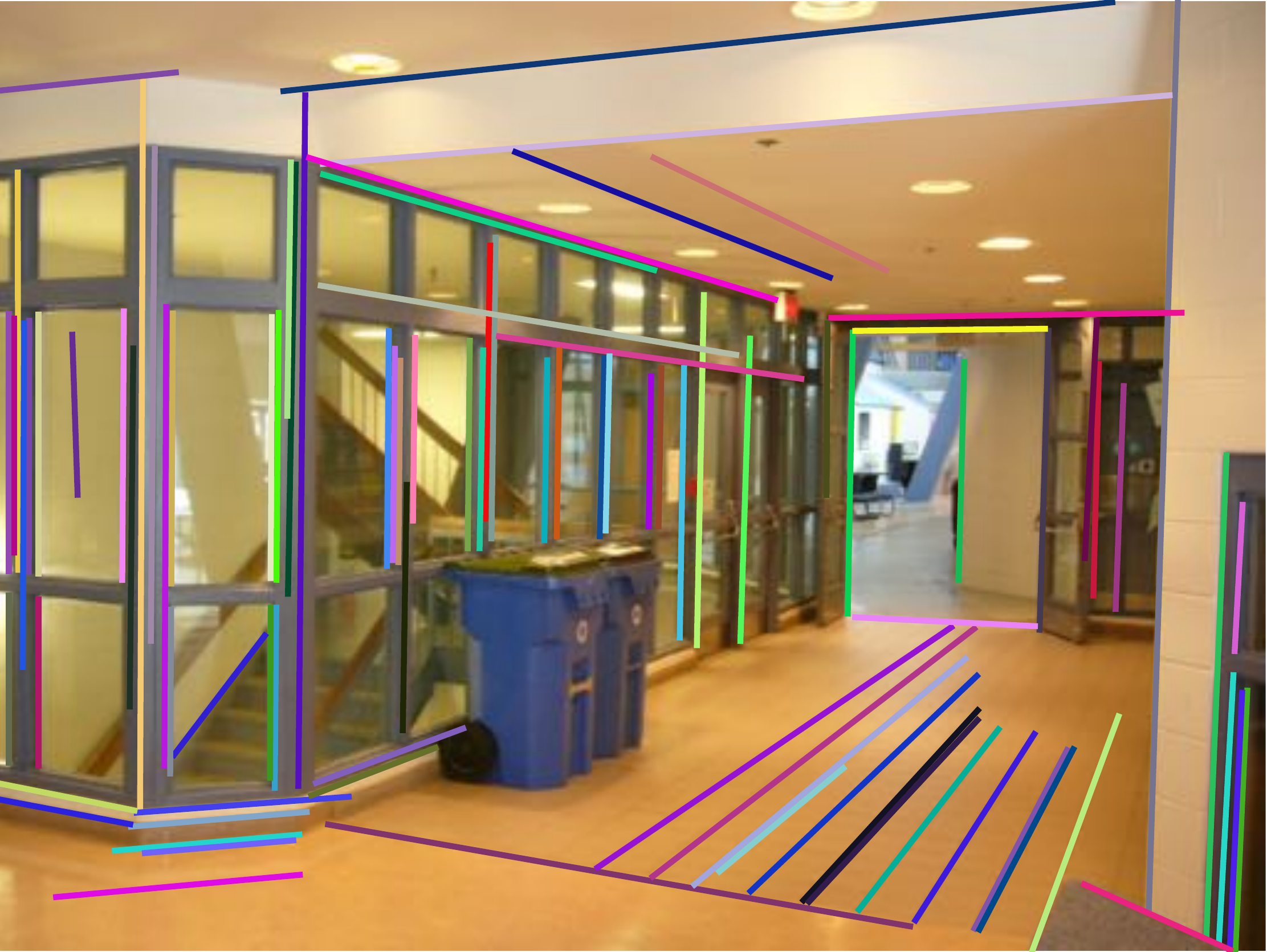}}&
		{\includegraphics[width=0.225\textwidth]{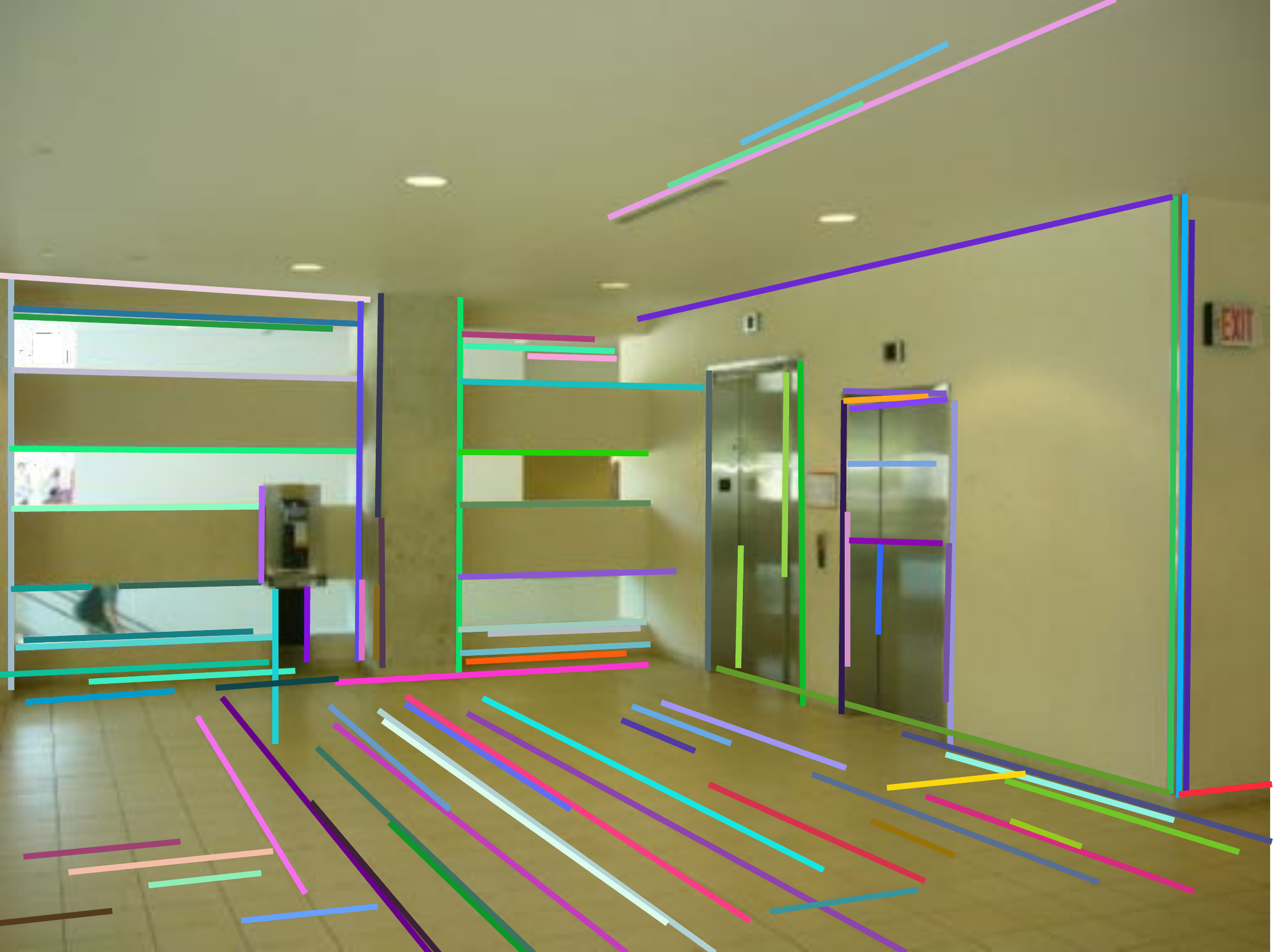}}&
		{\includegraphics[width=0.225\textwidth]{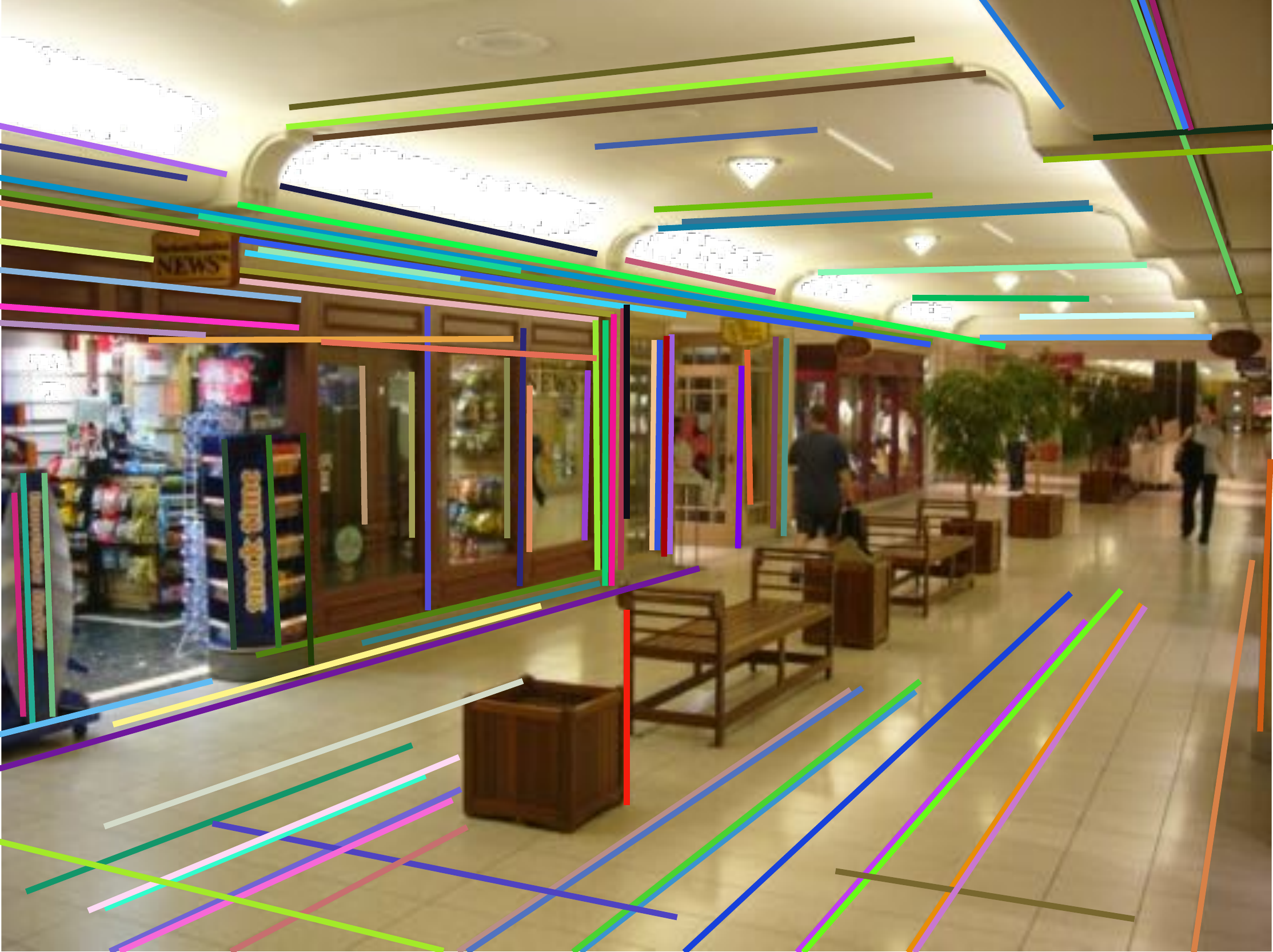}}& 
		{\includegraphics[width=0.225\textwidth]{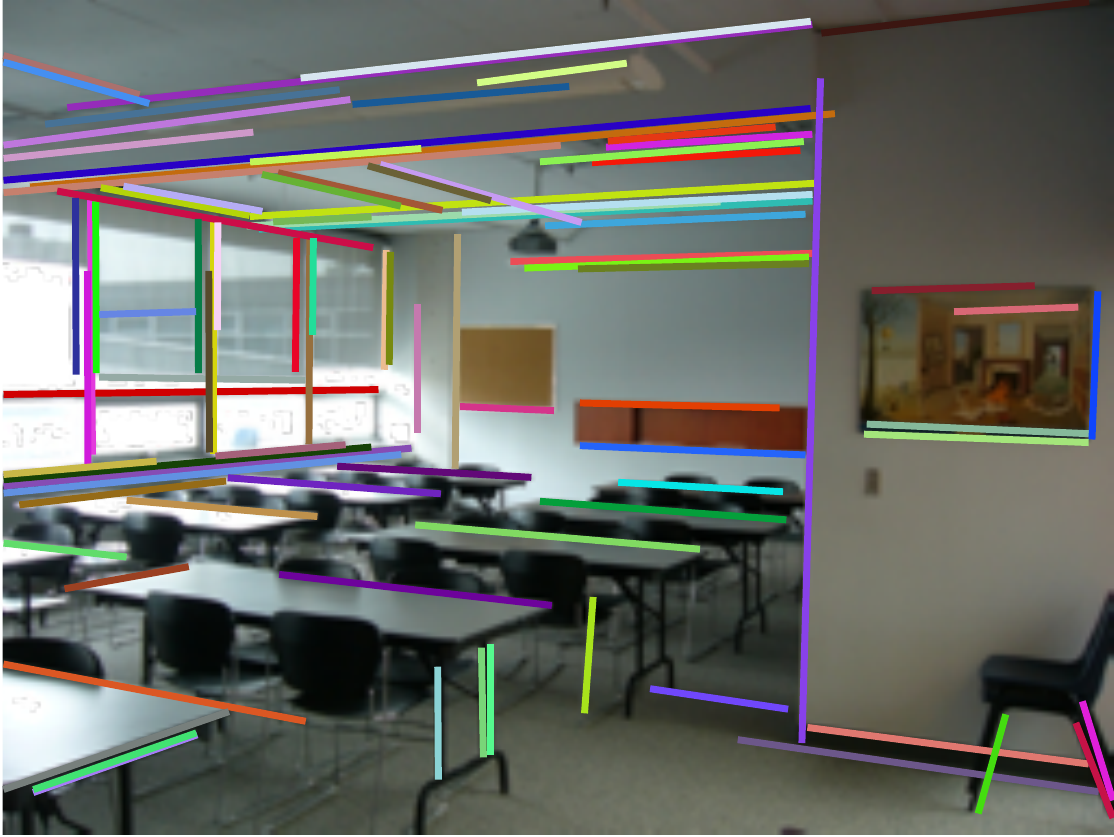}}\\
		
		\raisebox{3\normalbaselineskip}[0pt][0pt]{\rotatebox[origin=c]{90}{Ground Truth}}&
		{\includegraphics[width=0.225\textwidth]{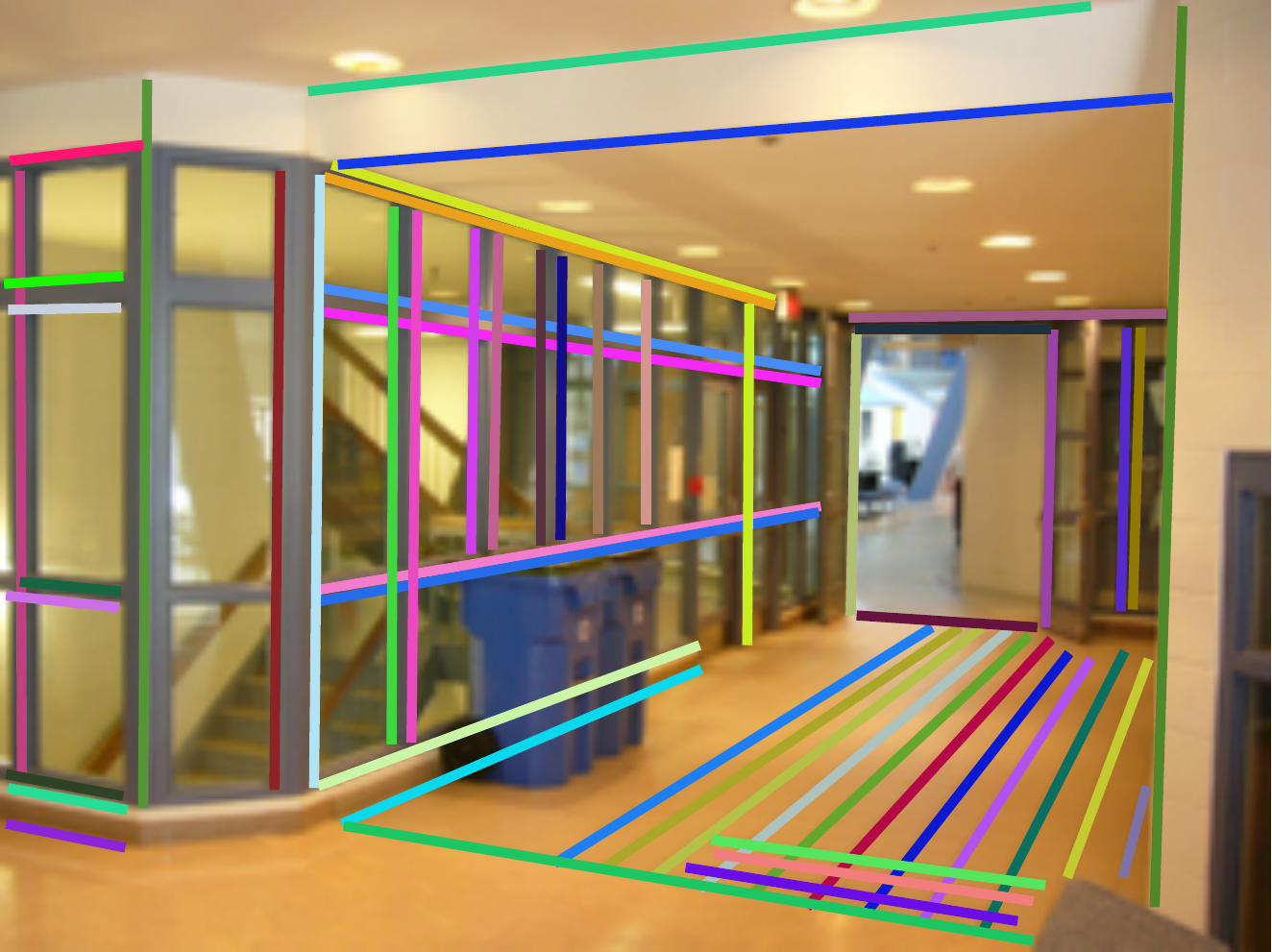}}&
		{\includegraphics[width=0.225\textwidth]{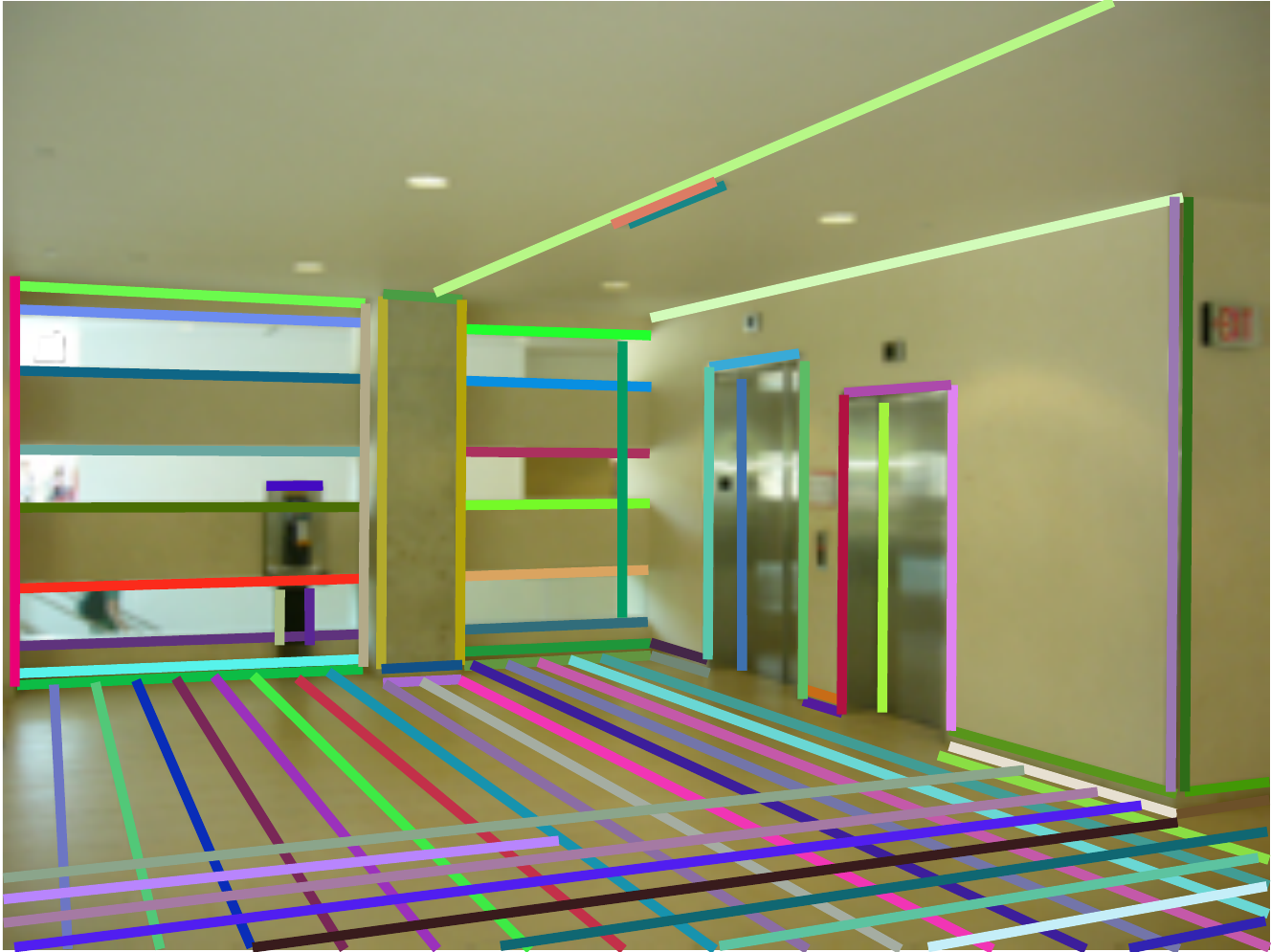}}&
		{\includegraphics[width=0.225\textwidth]{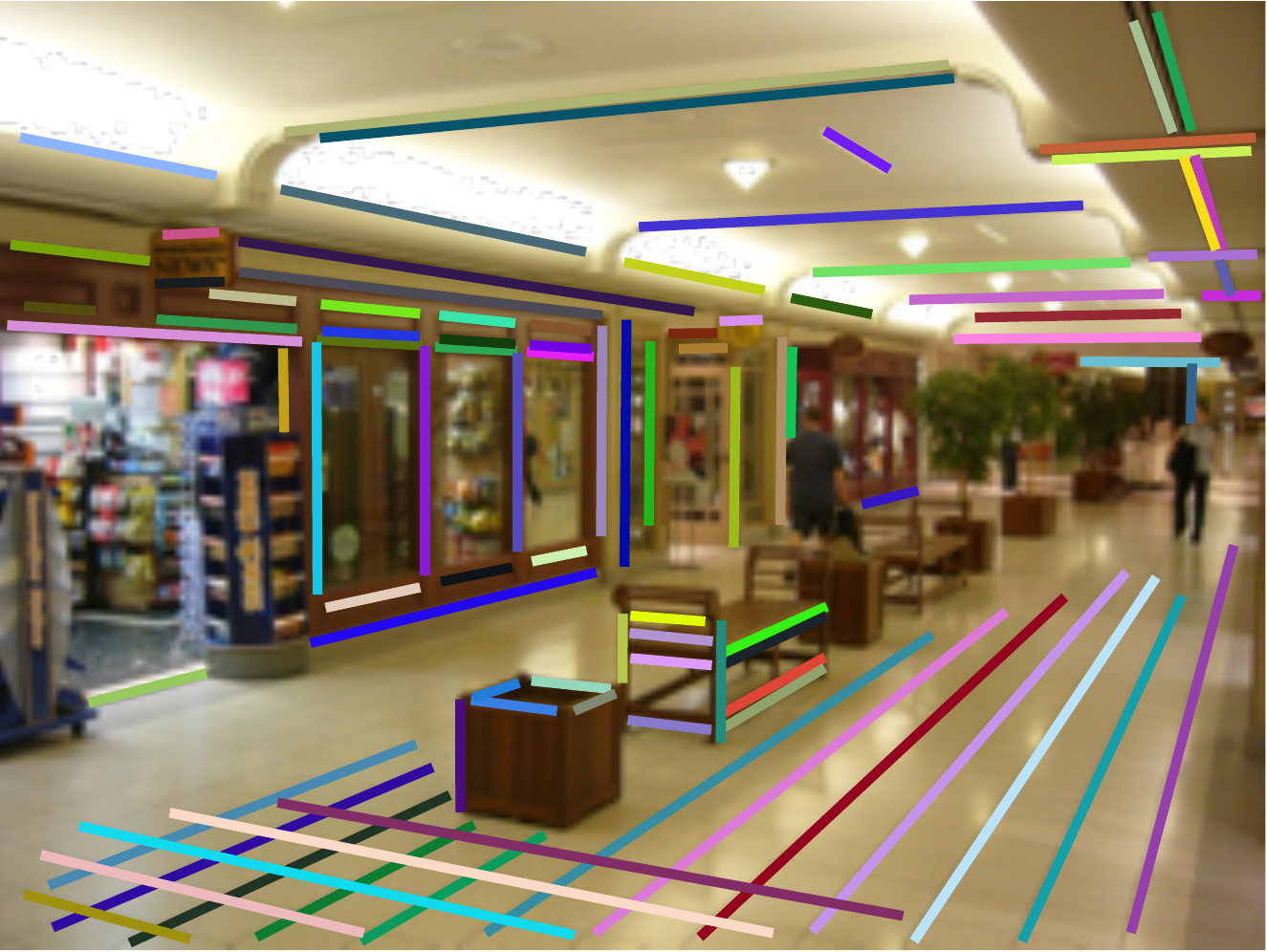}}& 
		{\includegraphics[width=0.225\textwidth]{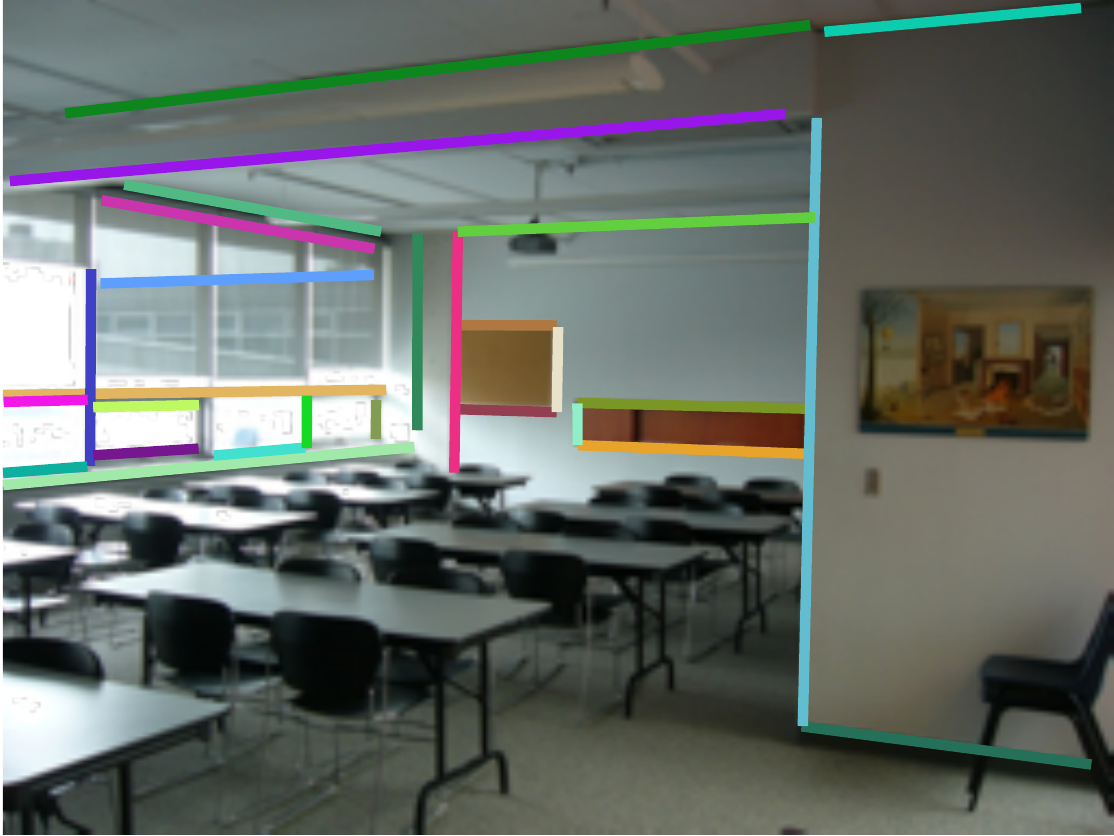}}\\
		 
	\end{tabular}
	\caption{Top 90 segments returned by the six algorithms under evaluation, together with hand-labelled ground truth, for four example test images drawn from the YorkUrbanDB dataset.} 
	\label{fig:datasamples}
\end{figure*}

Fig.~\ref{fig:datasamples} shows the top-ranked 90 segments returned by each algorithm on four example images from the YorkUrbanDB test dataset.  To our eyes, the Attraction Field and MCMLSD results look strongest, but in complementary ways.  While the Attraction Field method appears more adept at picking out short segments (e.g., the windows in the first example), MCMLSD is more successful at recovering the longer segments (e.g., the lines on the ground plane in the first three examples).  

Interestingly, in the second example the Attraction Field method succeeds in detecting some of the  line segments projecting from the tiling pattern on the floor, despite being trained (on the Wireframe dataset) to ignore these.

\subsection{Quantitative Results}

\subsubsection{YorkUrbanDB Test Set}

\begin{figure*}[htbp]
	\centering
	\subfloat[]{\includegraphics[width=0.32\textwidth]{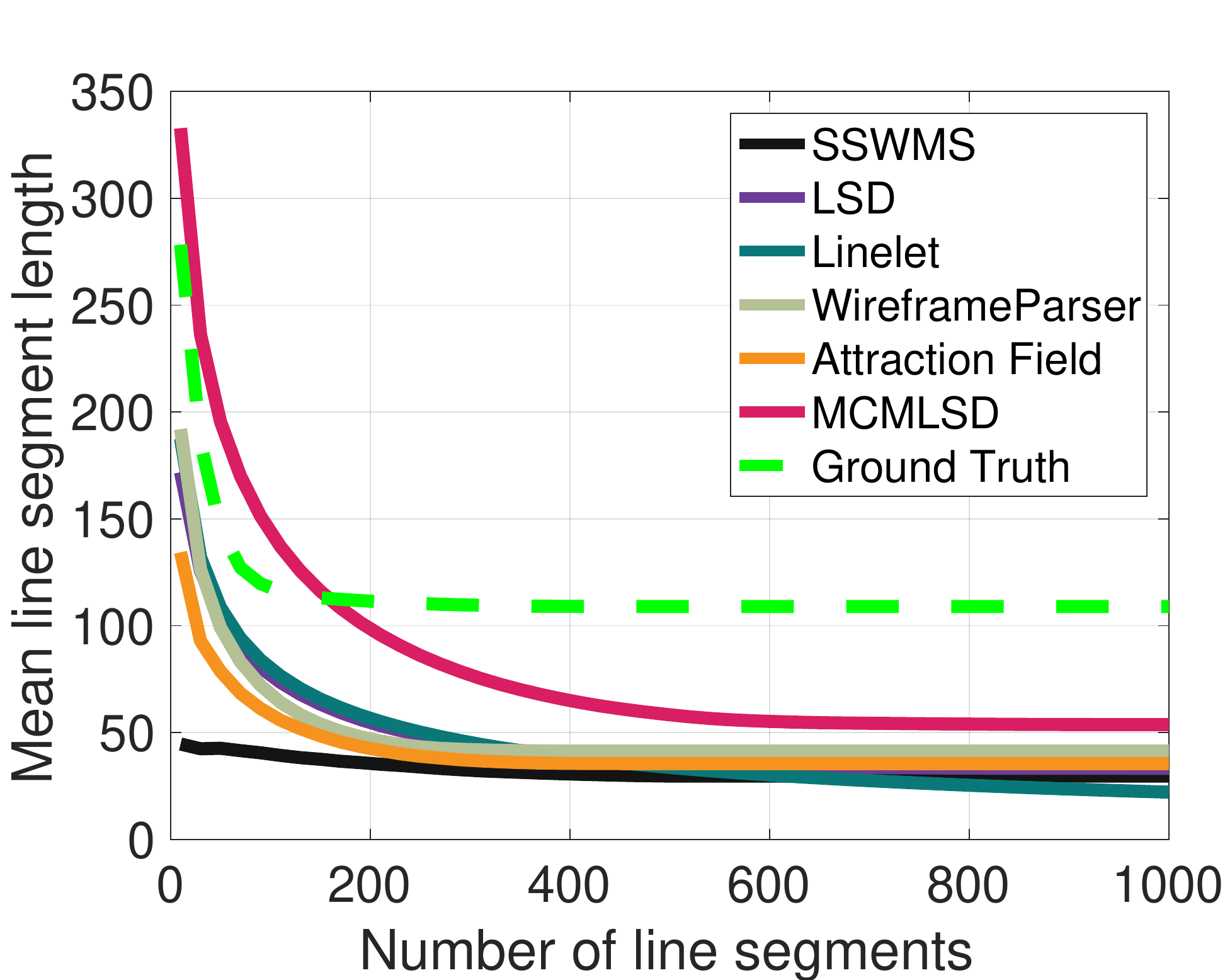}}\hspace{0.05\textwidth}
	\subfloat[]{\includegraphics[width=0.32\textwidth]{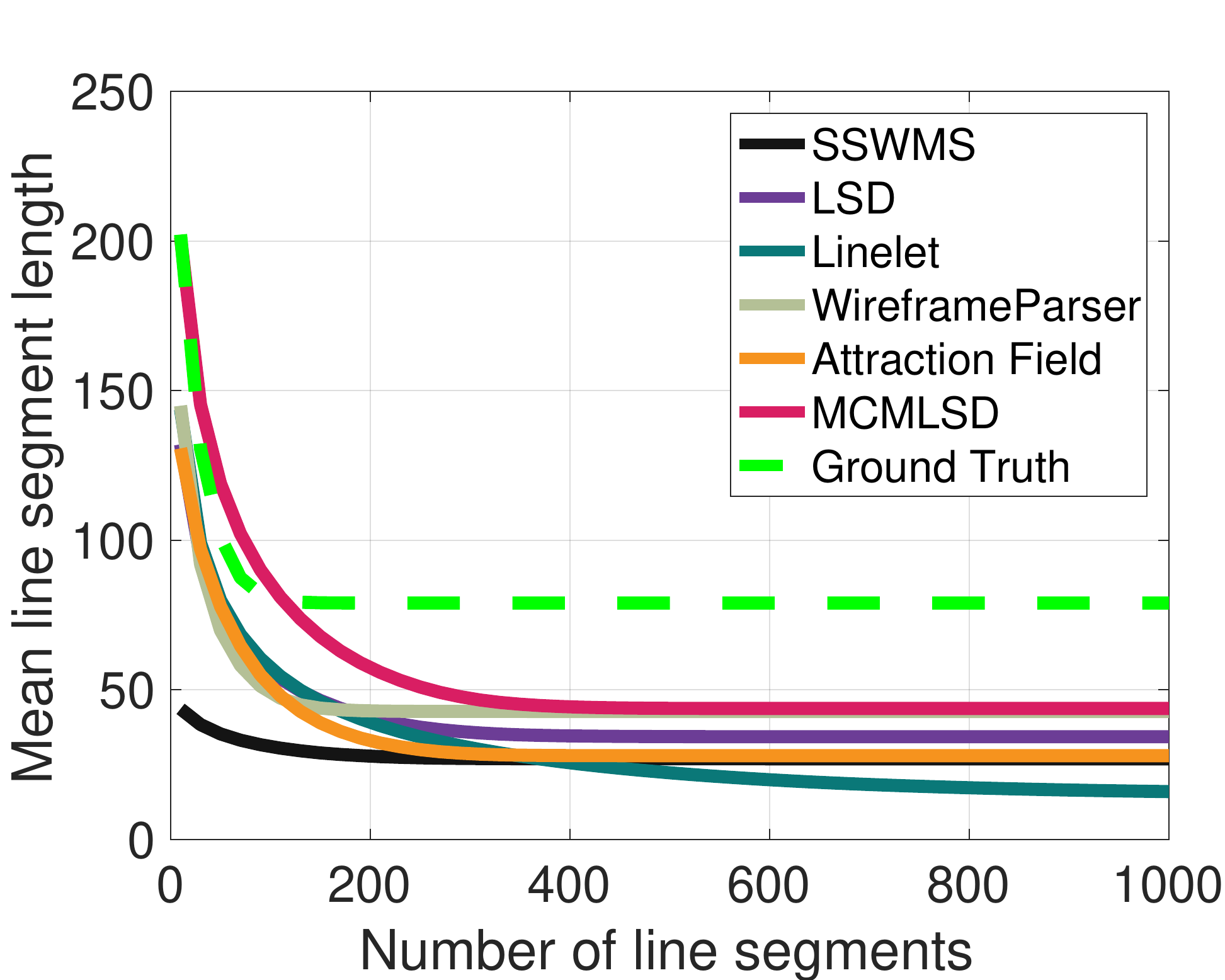}}
	\caption{Mean length of ranked line segments returned by each algorithm for (a) YorkUrbanDB and (b) Wireframe test sets, as a function of the number of segments returned.  Ground truth line segments are ranked from longest to shortest.}
	\label{fig:length}
\end{figure*}

Fig. ~\ref{fig:length}(a) shows the mean length of ranked line segments returned by each of the six algorithms, compared to the ground truth line segment lengths.
Although the algorithms generally rank longer segments higher, all ultimately return segments that are on average shorter than the ground truth segments.
Consistent with the qualitative observations above, MCMLSD tends to return longer segments than the other approaches.

\begin{figure*}[htbp]
	\centering
	\subfloat[]{\includegraphics[width=0.32\textwidth]{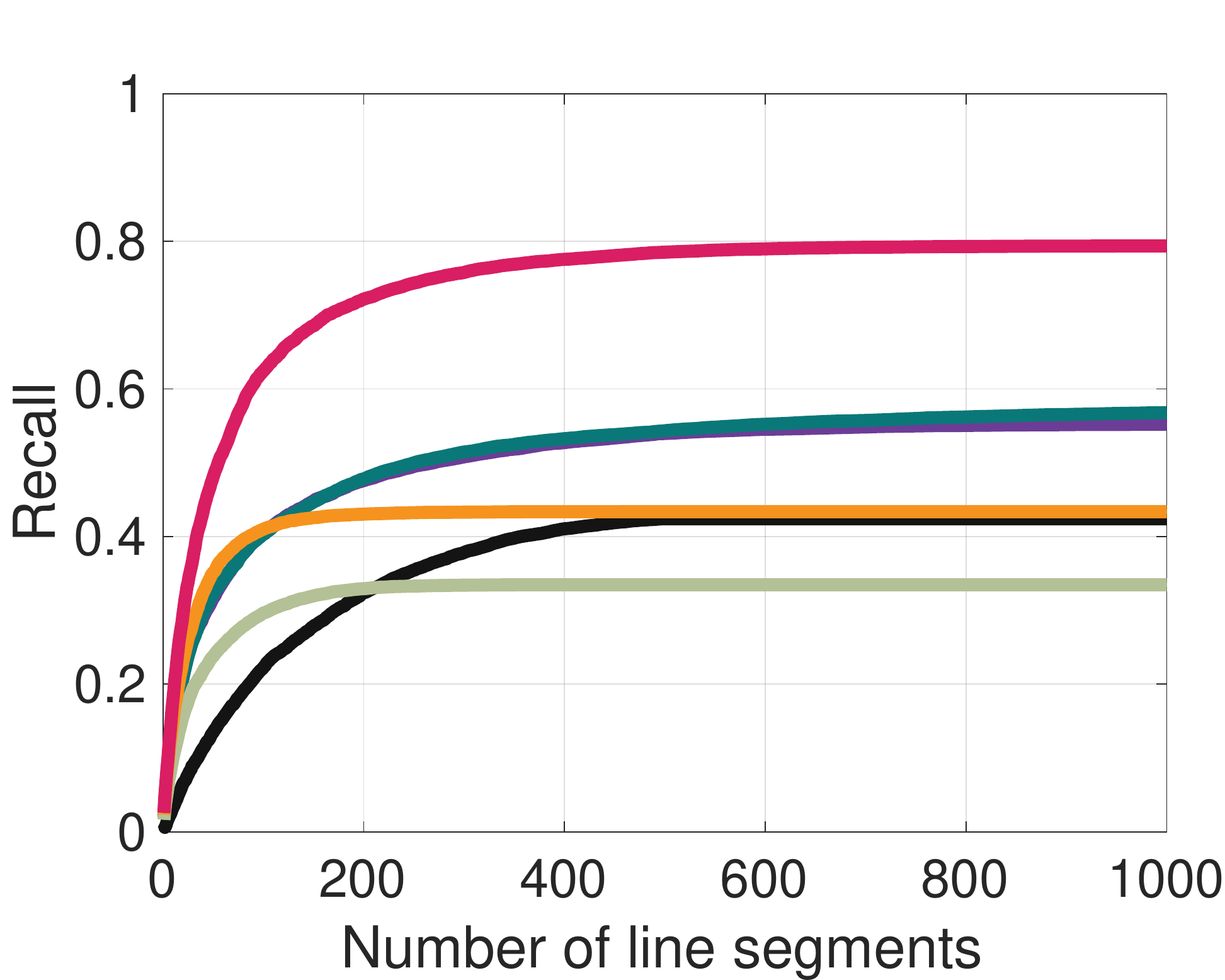}}
	\subfloat[]{\includegraphics[width=0.32\textwidth]{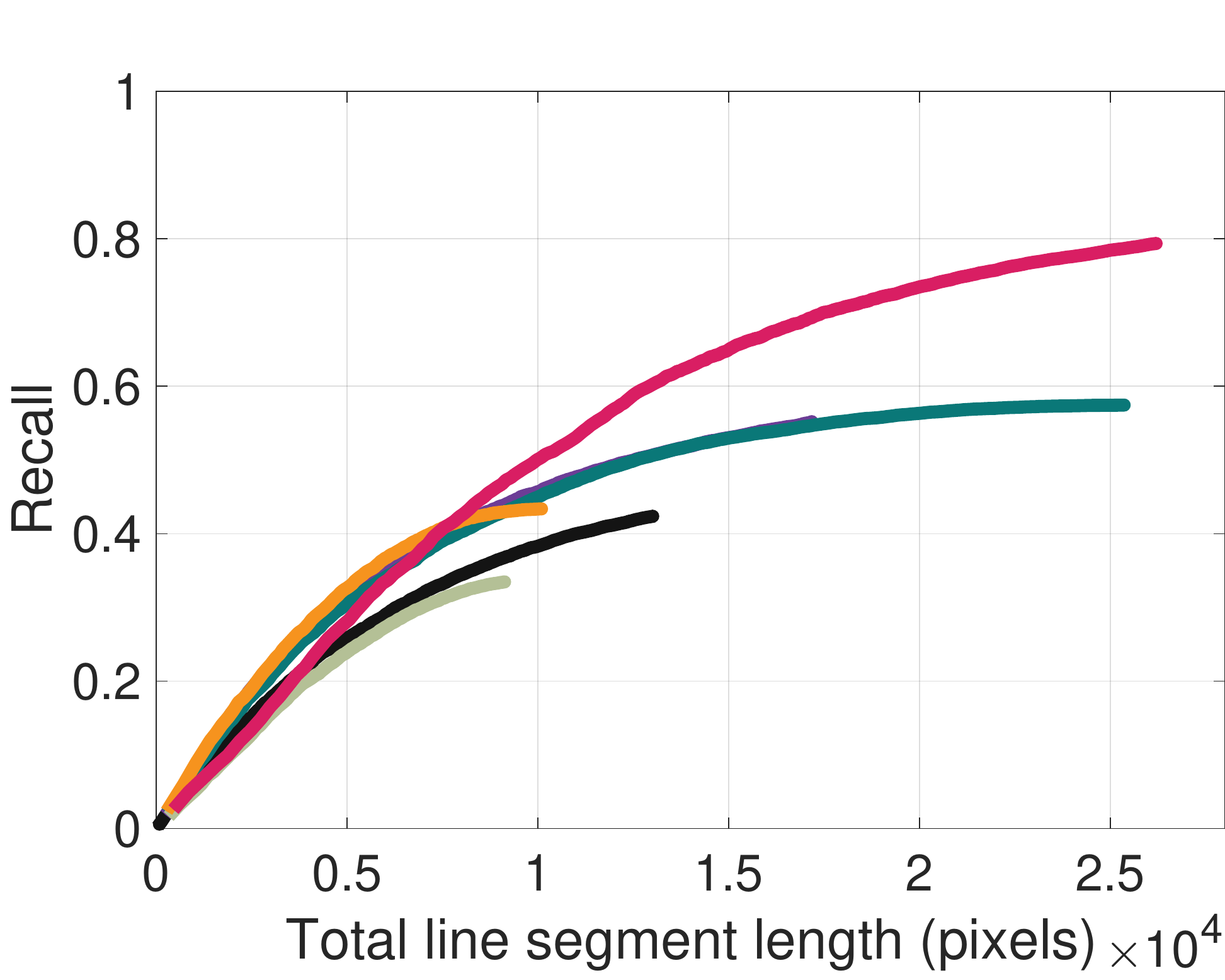}}
	\subfloat[]{\includegraphics[width=0.32\textwidth]{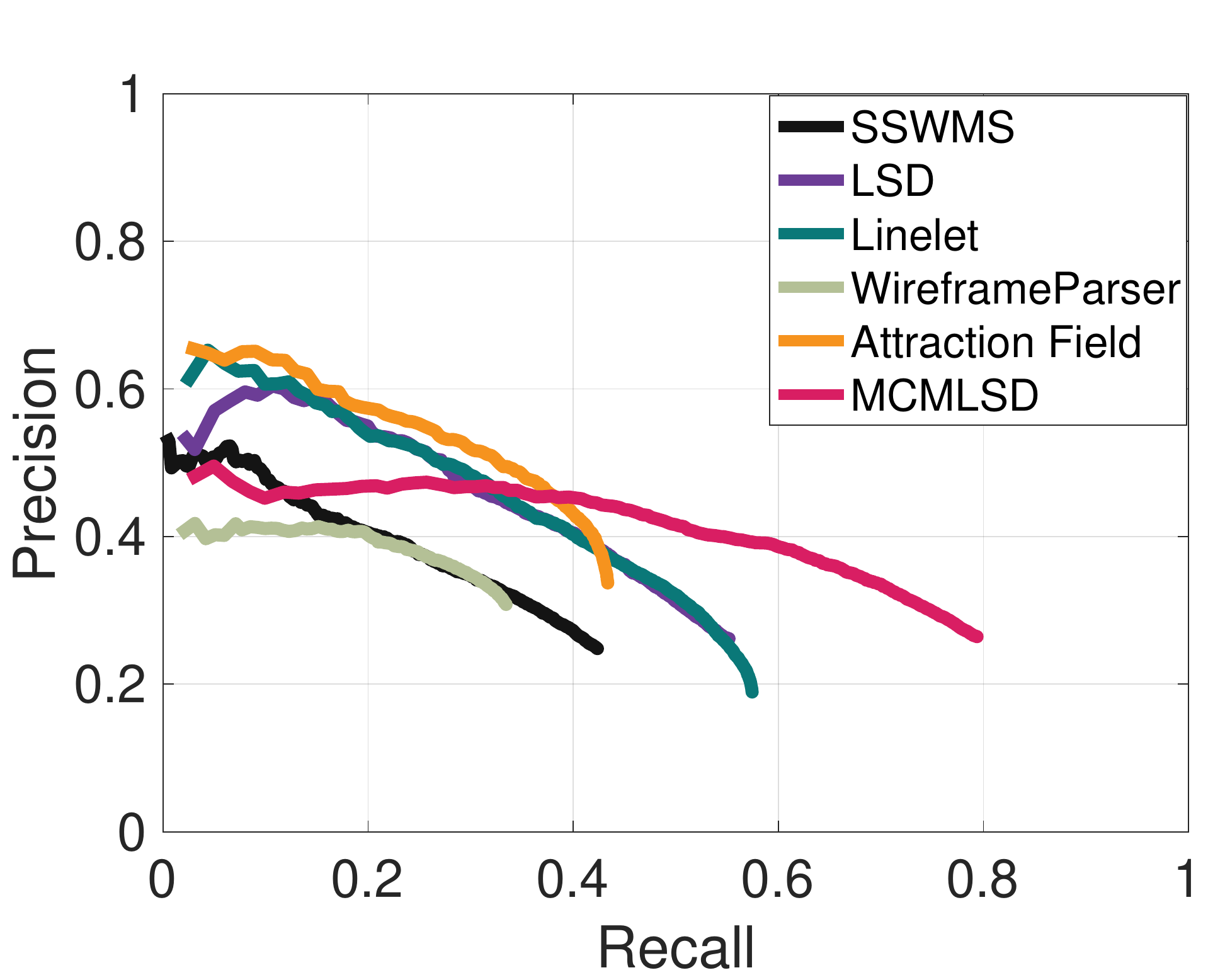}}
	\caption{Performance of the six algorithms under evaluation on the YorkUrban Dataset.  (a) Recall as a function of number of segments returned.  (b) Recall as a function of the total length of segments returned.  (c) Precision-Recall.}
	\label{fig:eval}
\end{figure*}

Fig.~\ref{fig:eval} provides a quantitative comparison of all six algorithms on the YorkUrbanDB test set.  MCMLSD achieves a maximum recall of 0.8, roughly 45\% better than the LSD and Linelet methods.  Interestingly, MCMLSD outperforms the more recent deep learning algorithms by an even larger margin - maximum recall for MCMLSD is  roughly 140\% higher than for the Wireframe Parser algorithm and 90\%  higher than for the Attraction Field algorithm.  

Analysis of each of the three performance measures yields additional insights.  Fig.~\ref{fig:eval}(a) shows that if a constraint is placed on the number of segments returned, e.g., to limit complexity for downstream analysis, MCMLSD consistently achieves higher recall.  While the deep Attraction Field algorithm is competitive with the traditional LSD and Linelet algorithms for very tight constraints (fewer than 100 segments), it falls behind  as this constraint is relaxed.

The story is a little different if the constraint is placed on the total segment length rather than the total number of segments (Fig. \ref{fig:eval}(b)).  Here we see that while MCMLSD vastly outperforms the other methods for more relaxed constraints (more than $10^4$ pixels), for tighter constraints, the Attraction Field, LSD and Linelet algorithms become slightly superior.  This can be accounted for by the tendency of MCMLSD to extract longer segments than the Attraction Field, LSD and Linelet algorithms.

Finally, Fig. \ref{fig:eval}(c) shows that the algorithm of choice very much depends upon the relative value of precision and recall for a particular application.  If recall greater than 0.4 is required, MCMLSD is clearly preferred.  However, if precision greater than 0.45 is required, then recall must be sacrificed and the Attraction Field or Linelet algorithms are preferred.  It should be remembered, however, that since the YorkUrbanDB ground truth is incomplete, lower precision may be due to detection of useful segments that just do not happen to be labelled in the ground truth.

\subsubsection{Wireframe Test Set}
\begin{figure*}[htbp]
	\centering
	\subfloat[]{\includegraphics[width=0.32\textwidth]{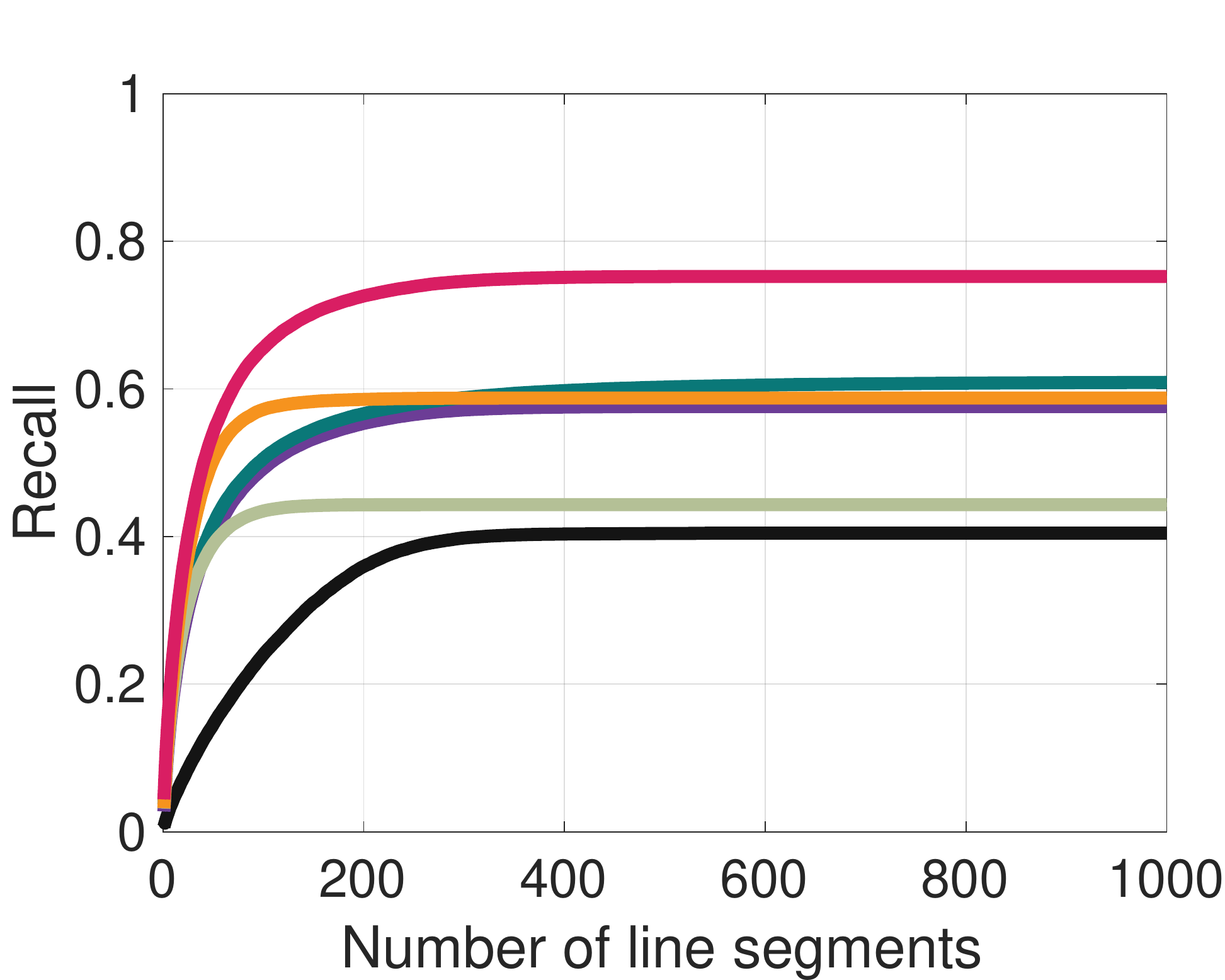}}
	\subfloat[]{\includegraphics[width=0.32\textwidth]{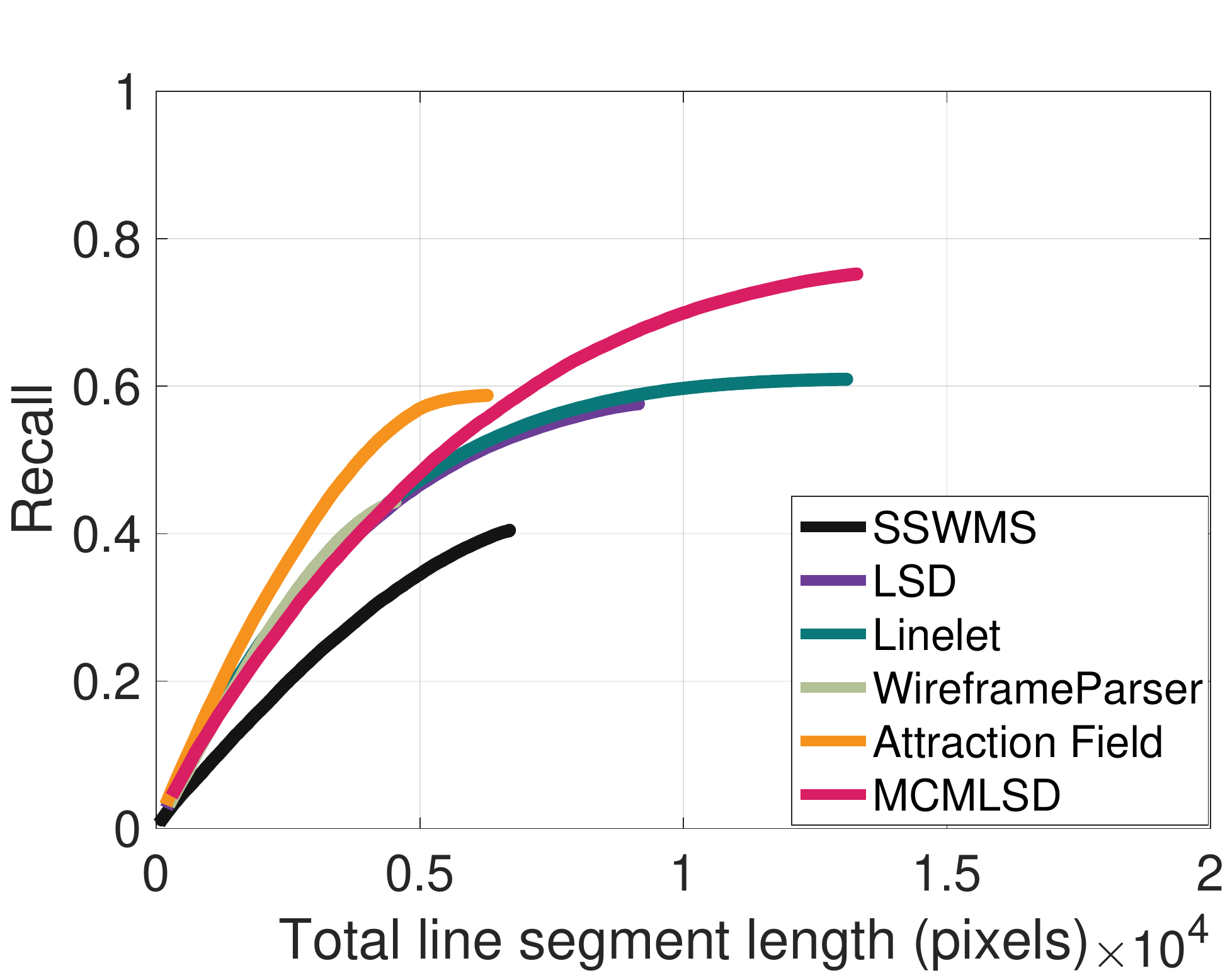}}
	\subfloat[]{\includegraphics[width=0.32\textwidth]{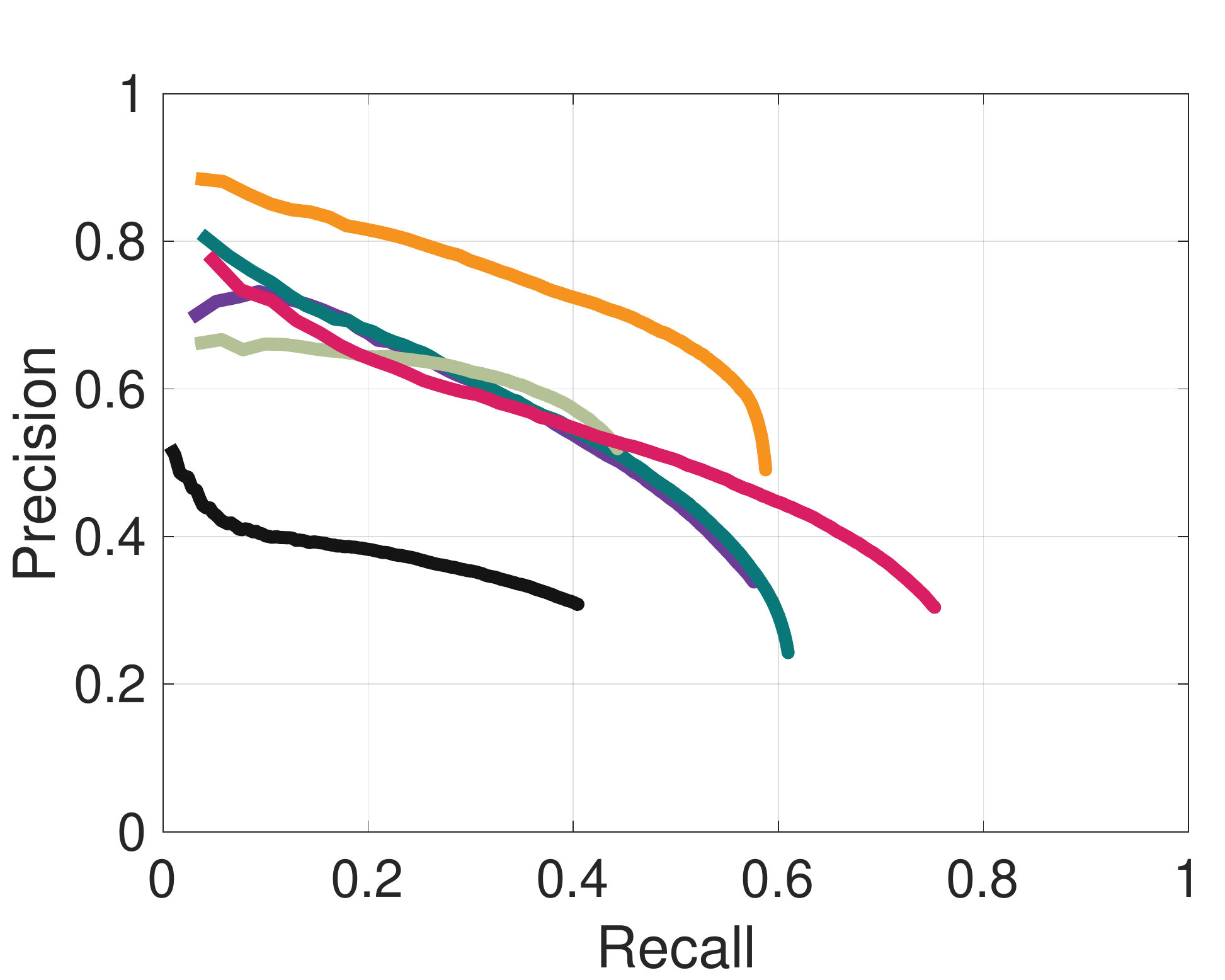}}
	\caption{Performance of the proposed MCMLSD methods compared with the state of the art on the Wireframe dataset\cite{wireframe_cvpr18}.  (a) Recall as a function of number of segments returned.  (b) Recall as a function of the total length of segments returned.  (c) Precision-Recall.}
	\label{fig:evalwireframe}
\end{figure*}

Fig. \ref{fig:evalwireframe} shows the same evaluation on the Wireframe test set.  The maximum recall achieved by MCMLSD is 0.75, almost as high as for YorkUrbanDB, even  without  fine-tuning of parameters or distributions, indicating good generalization ability.  The performance advantage is smaller than for YorkUrbanDB, but MCMLSD is still 26\% better than its closest competitors. In terms of maximum recall, the Attraction Field algorithm is now competitive with the LSD and Linelet algorithms.  MCMLSD still leads the pack when the number of line segments is constrained (Fig. \ref{fig:evalwireframe}(a)).  As for the YorkUrbanDB dataset, when total line segment length is constrained, MCMLSD dominates in the high-recall regime.  However, for the Wireframe dataset, the Attraction Field algorithm now dominates in the low-recall regime.  Fig. \ref{fig:evalwireframe}(c) tells a similar story for precision-recall:  MCMLSD dominates in the high-recall regime, but the Attraction Field method is superior in the low-recall (high-precision) regime.  The improved performance of the Attraction Field method for the Wireframe dataset relative to the YorkUrbanDB dataset is not surprising, as it was trained on the Wireframe training partition.

\subsubsection{Ranking Revisited}
What accounts for the superiority of MCMLSD in the high-recall regime, and the superiority of the Attraction Field algorithm (and, for YorkUrbanDB, the LSD and Linelet algorithms) in the low-recall regime?
One possible factor is  the quality of the line segments they return.  But another possible factor is the ranking employed by each method.  To dissociate these two factors, we employed an oracle to rank the segments returned by each algorithm for the YorkUrbanDB dataset according to ground truth precision.  Specifically, after 1:1 association with ground truth segments, the algorithm segments are ranked in terms of the proportion of their points that have a 1:1 ground truth match.  Ties are resolved by length, with longer segments ranked first.  

\begin{figure*}[!htbp]
	\centering
	\subfloat[]{\includegraphics[width=0.32\textwidth]{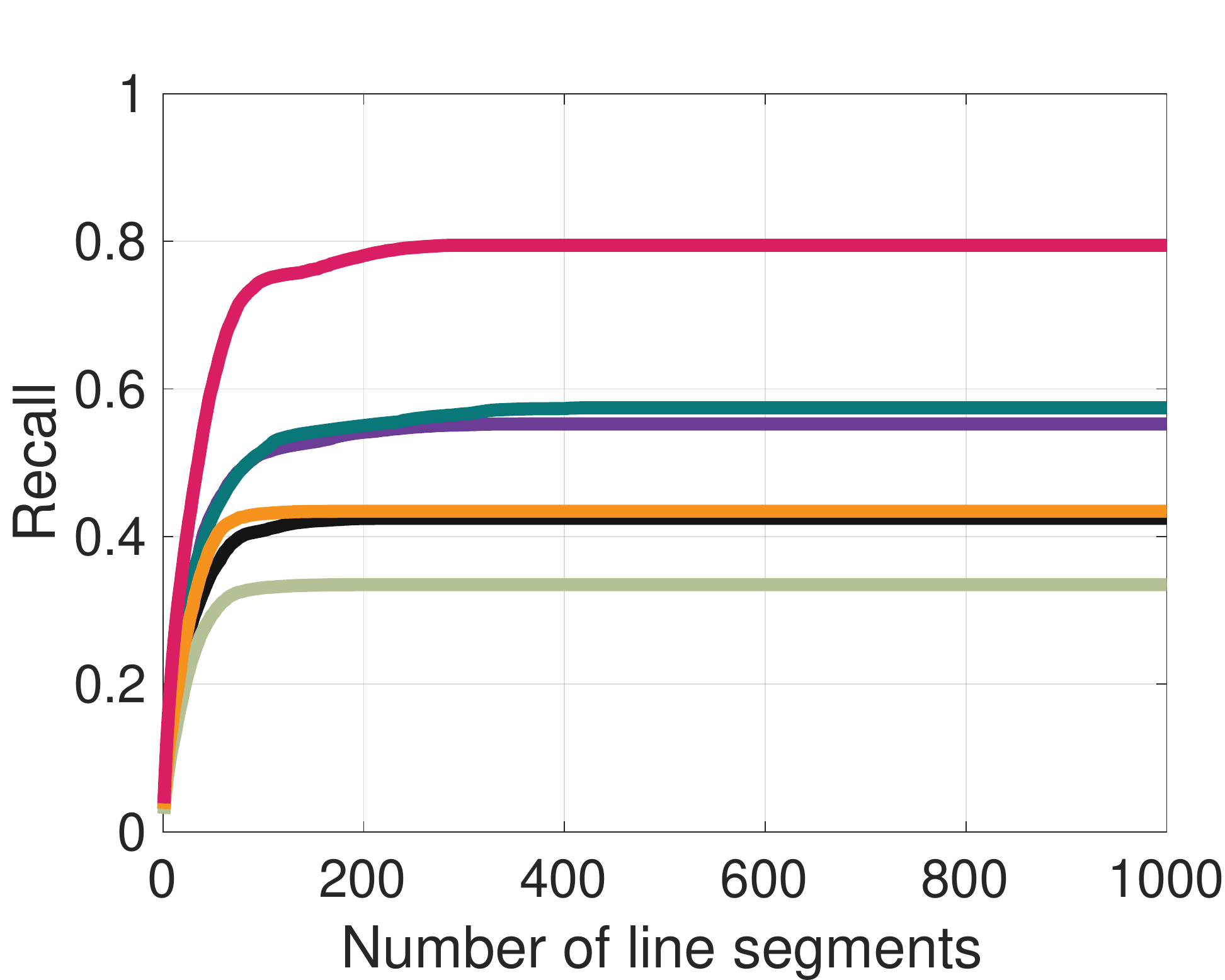}}
	\subfloat[]{\includegraphics[width=0.32\textwidth]{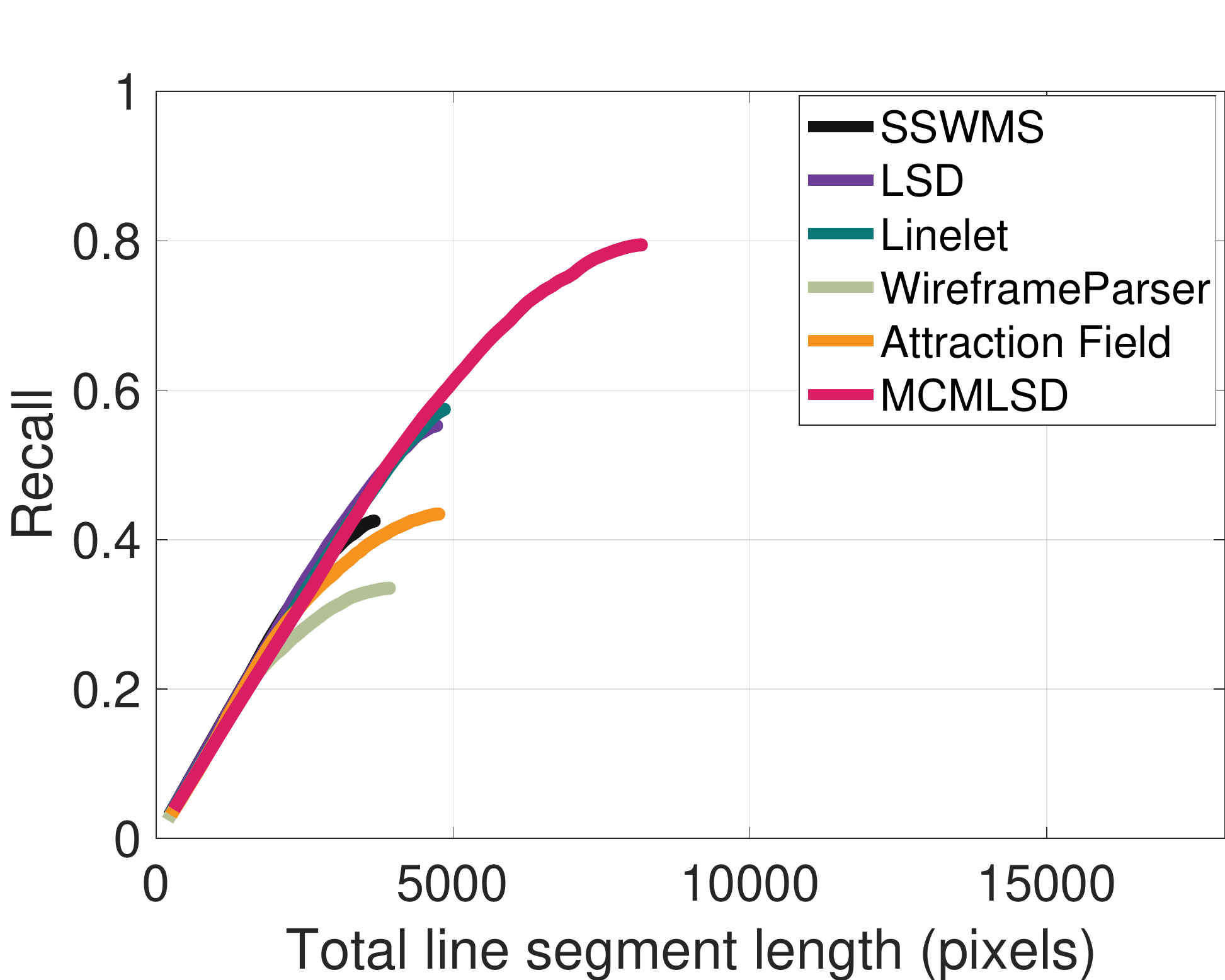}}
	\subfloat[]{\includegraphics[width=0.32\textwidth]{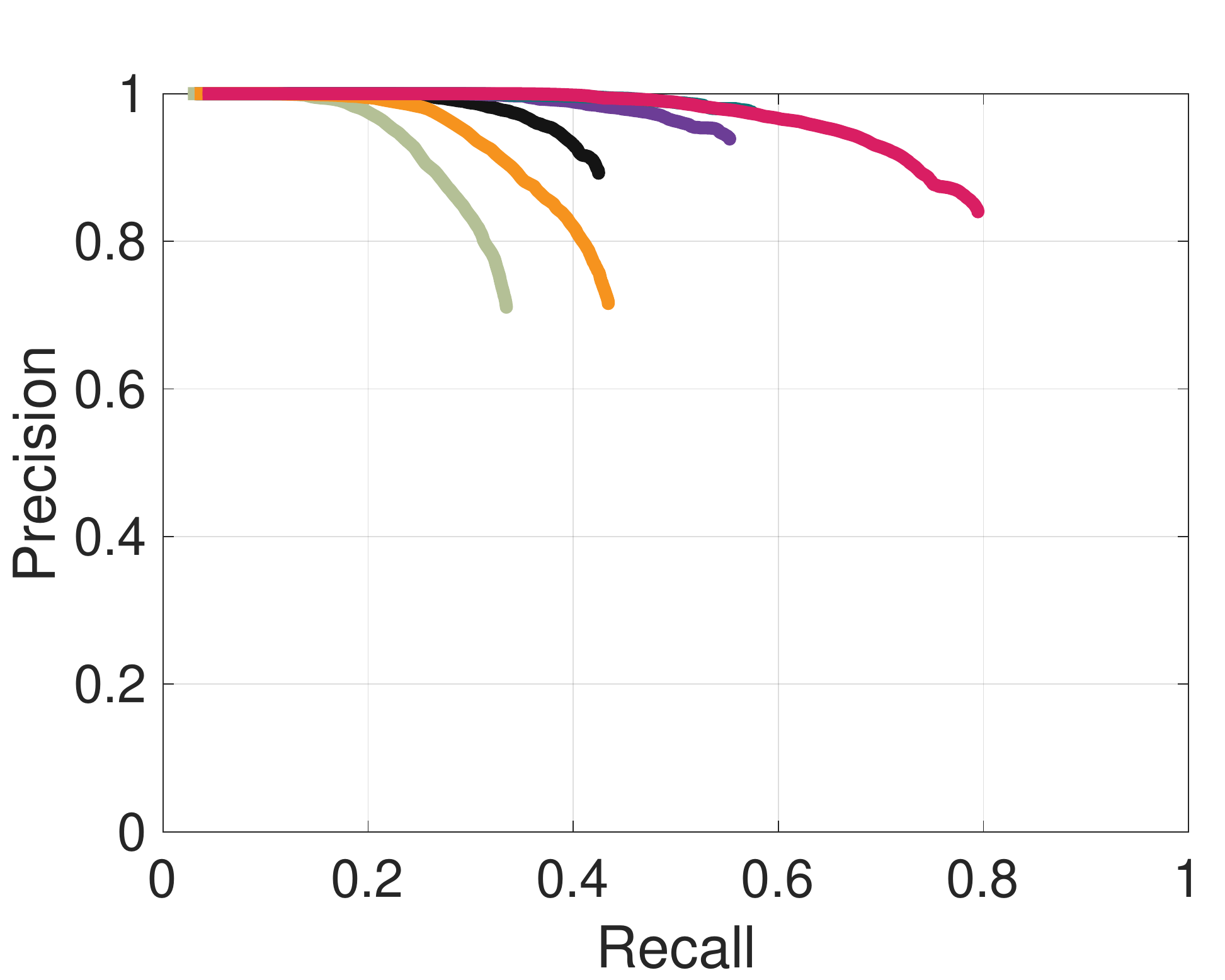}}
	\caption{Performance on the YorkUrbanDB dataset when an oracle is used to rank the segments by their precision.   (a) Recall as a function of number of segments returned.  (b) Recall as a function of the total length of segments returned.  (c) Precision-Recall.}
	\label{fig:oracle}
\end{figure*}

The results are illuminating (Fig. \ref{fig:oracle}).  While MCMLSD necessarily still achieves highest recall, and still dominates when the number of line segments is constrained, the precision advantage of the Attraction Field method in the low-recall regime has evaporated.  This indicates that the advantage of the Attraction Field algorithm in the low-recall regime derives not from superior line segments but from superior ranking.  This in turn suggests that the performance of other methods such as MCMLSD in the low-recall regime might be improved by adopting a revised ranking strategy.  

One limitation of the ranking strategy adopted in our original CVPR paper~\cite{almazen2017dynamic} is that it considers only the location and orientation of edges detected by the Elder \& Zucker multiscale edge detector~\cite{elder1998local}, which employs a signal detection approach based only on the local luminance contrast.  This ignores local colour and texture cues that can signal the relative importance of these edges. 

To incorporate this appearance information, we employ the structured forests edge detector of Doll\'ar and Lawrence~\cite{dollar2014fast} (code obtained from \url{https://github.com/pdollar/edges}), which was trained on the BSDS 500 dataset to use the local pattern of colours and textures to identify edges delineating ``distinguished things", as judged by human observers~\cite{Martin:04}.  Our hypothesis is  that the output of the structured forests edge detector will thus carry appearance cues  complementary to our  probabilistic ranking measure, which is based solely on the geometry of the edges.

To test this hypothesis,  we construct a logistic regressor that takes both of these cues as input to predict the precision of each segment,  train the regressor on the YorkUrbanDB training set, and then use it to rank segments in the test set.  Since we are interested in improving the precision of the detector, we employ a modified version of Ranking Method 4 (Section \ref{sec:ranking}), using the mean of the marginal probabilities along the segment instead of the sum.  To form the appearance cue we average the scalar responses of the structured forests edge detector  at the locations of the Elder \& Zucker edges within a 2-pixel distance of the line segment. 

\begin{figure*}[!htbp]
	\centering
	\subfloat[]{\includegraphics[width=0.32\textwidth]{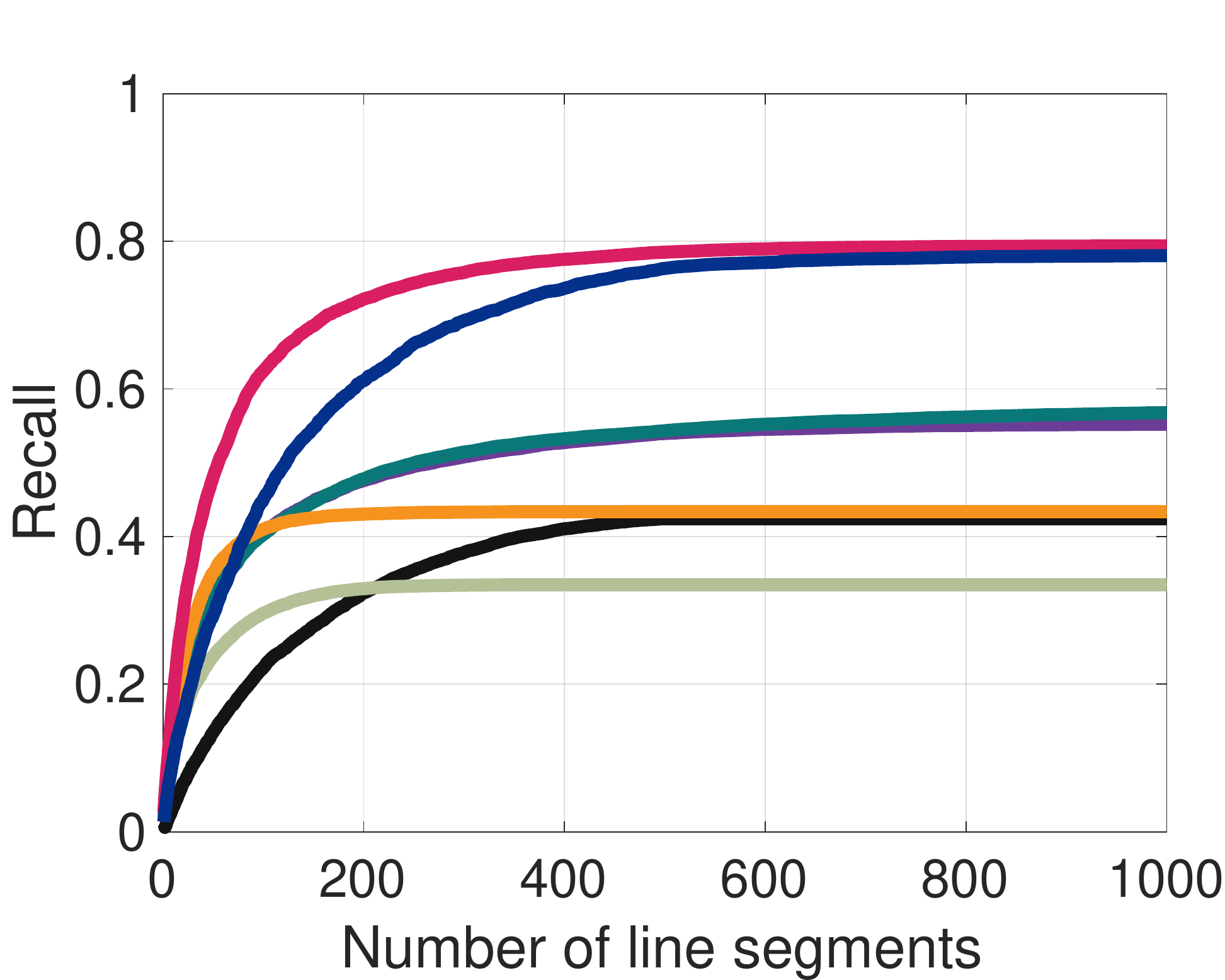}}
	\subfloat[]{\includegraphics[width=0.32\textwidth]{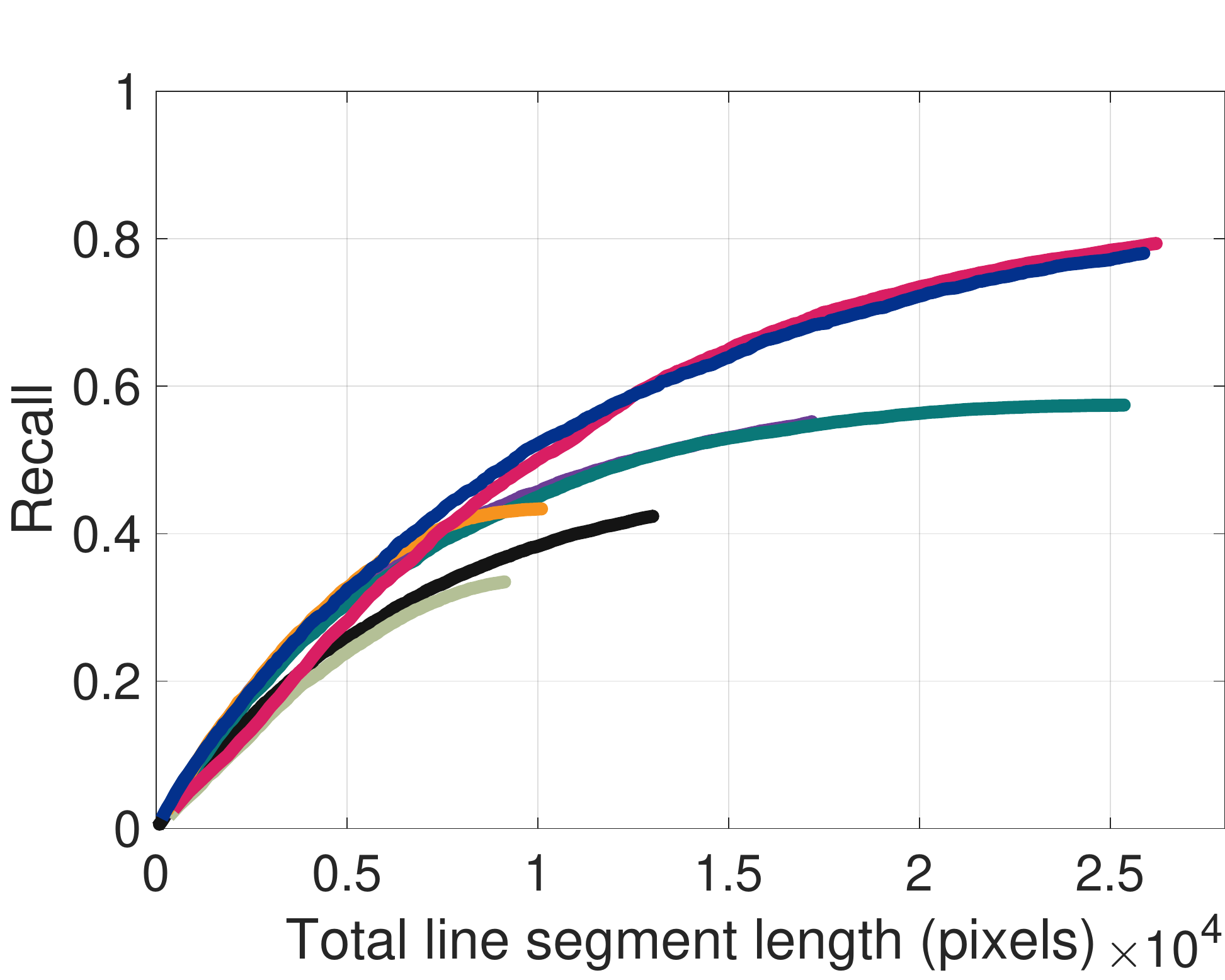}}
	\subfloat[]{\includegraphics[width=0.32\textwidth]{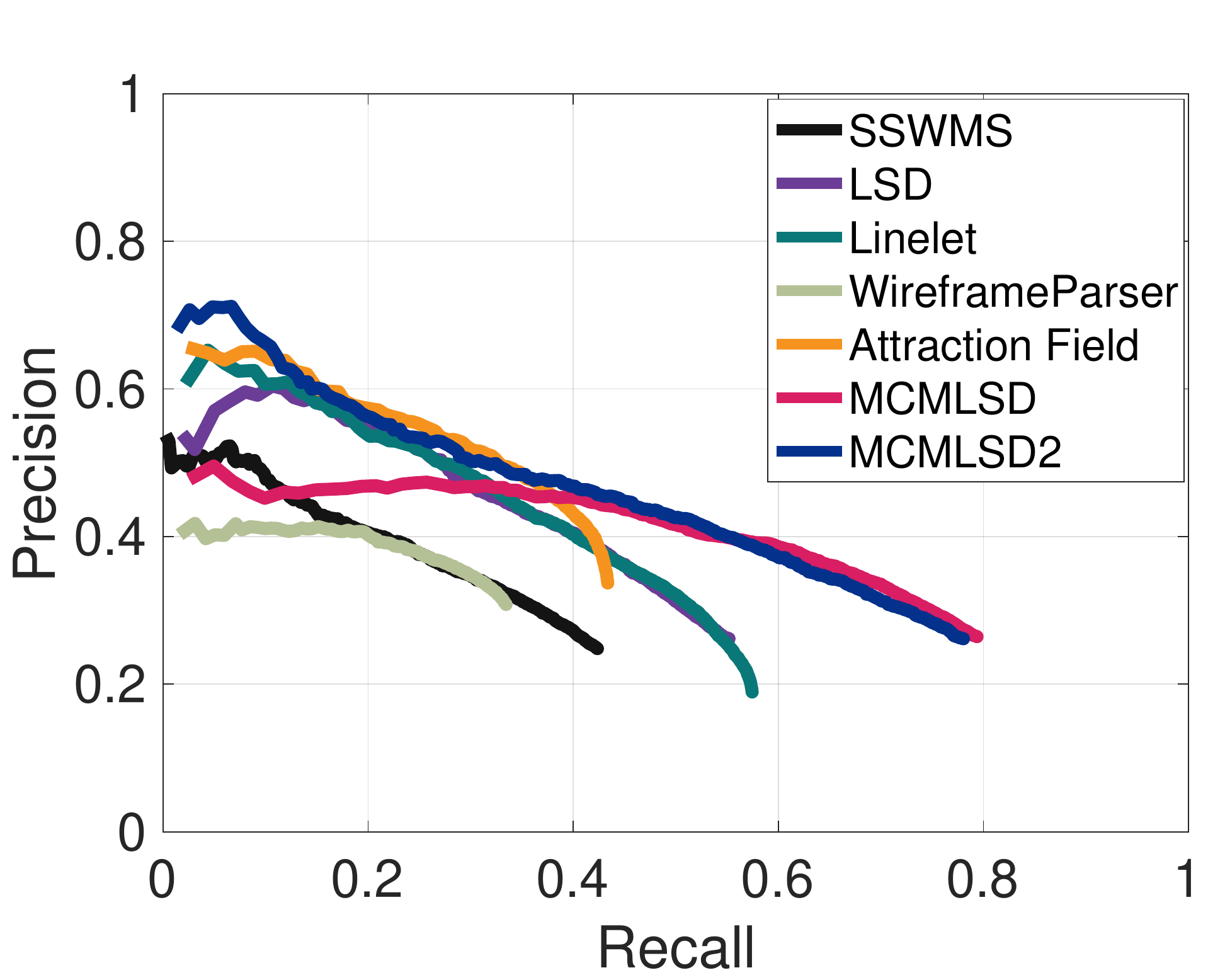}}
	\caption{Results including MCMLSD2, which uses the structured forests edge detector~\cite{dollar2014fast} to incorporate local appearance cues when ranking segments.  (a) Recall as a function of number of segments returned.  (b) Recall as a function of the total length of segments returned.  (c) Precision-Recall.}
	\label{fig:rerank}
\end{figure*}

Fig. \ref{fig:rerank} shows results of the MCMLSD algorithm using this revised ranking strategy (dubbed MCMLSD2), alongside the original MCMLSD algorithm and the five competitors.  
We see that with this revised ranking strategy, MCMLSD2 loses some recall performance when the number of line segments is constrained (Fig. \ref{fig:rerank}(a)), but is still vastly superior to the other methods.  At the same time, the precision of MCMLSD2 matches that of the Attraction Field, LSD and Linelet algorithms in the low-recall regime, and is far superior in the high-recall regime (Fig. \ref{fig:rerank}(c)).

\subsection{Reconciling with Recent Evaluations}
The results above may at first seem puzzling, since they seem to contradict the evaluations reported in recent papers that claim superiority of the deep Wireframe Parser and Attraction Field methods~\cite{wireframe_cvpr18,xue2018learning}.  This contradiction is due to differences in how algorithms were evaluated.  We laid out our evaluation methods in Section \ref{sec:eval} of this paper and in our original CVPR  paper~\cite{almazen2017dynamic}.  There are two key deviations between our evaluation approach and the approach employed in the Wireframe and Attraction Field
papers that account for this contradiction.

\subsubsection{Distance Threshold}
\begin{figure*}[!htbp]
	\centering
		\subfloat[]{\includegraphics[width=0.45\textwidth]{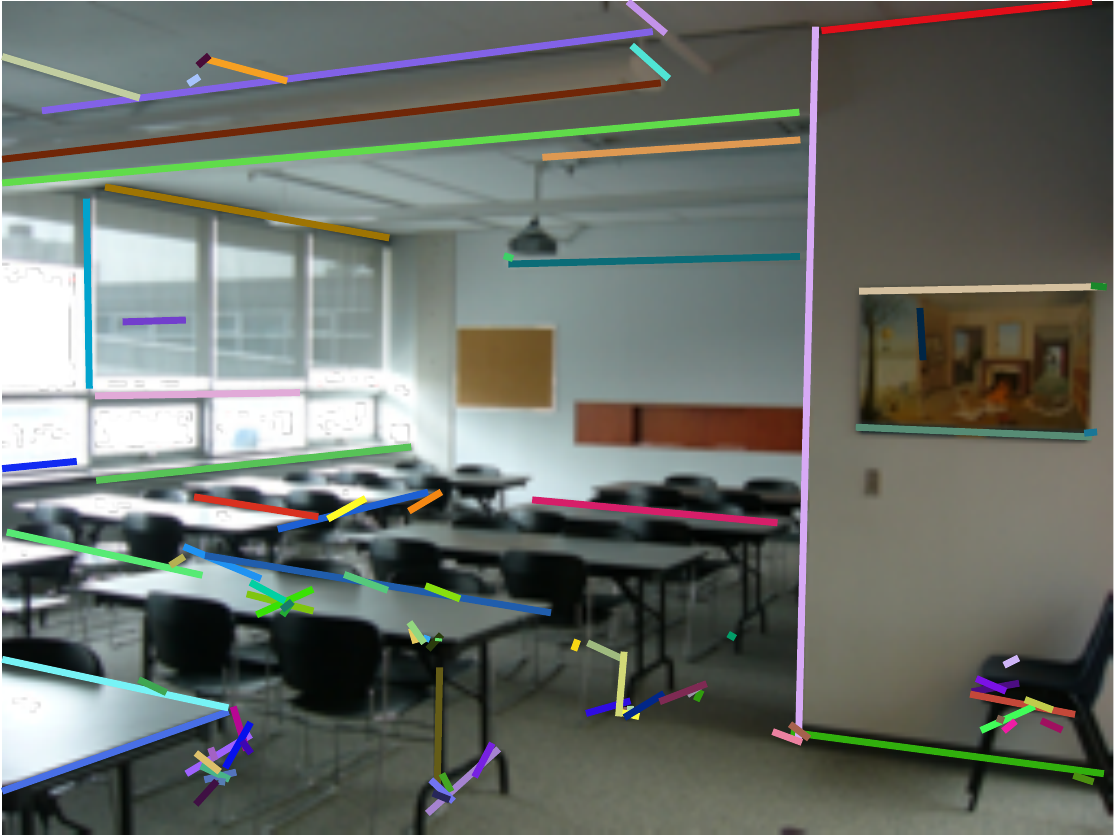}}\hspace{0.05\textwidth}
		\subfloat[]{\includegraphics[width=0.45\textwidth]{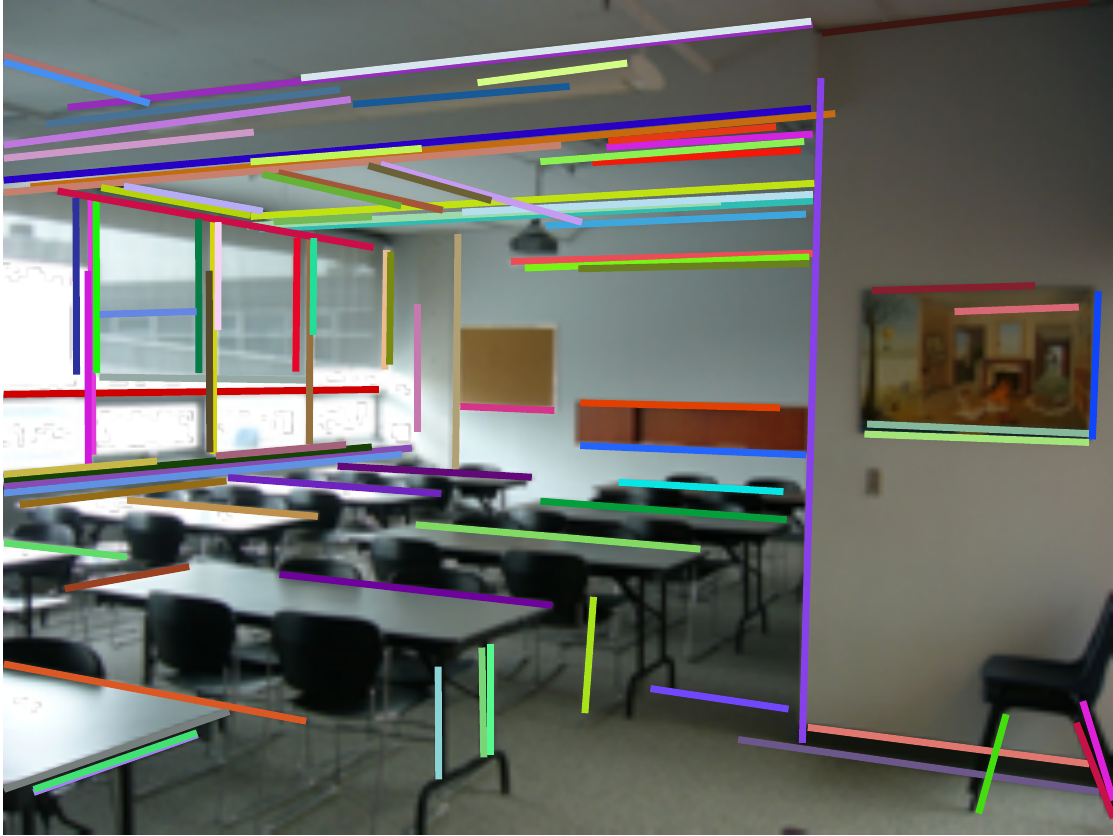}}\\
	\caption{Crop of example YorkUrbanDB test result for the (a) Attraction Field and (b) MCMLSD algorithms.  Observe the  alignment errors of some of the segments returned by  the Attraction Field algorithm.}
	\label{fig:crop}
\end{figure*}

In our evaluations, we employ a distance threshold of $2\sqrt{2}$ pixels, to associate any pair of lines that could potentially appear in the image with less than a one-pixel intervening gap.  This seems like a reasonable threshold, able to account for small localization errors in edge detection due to pixel discretization.  However, in the deep network papers, a threshold of 1\% of diagonal image size was employed, which for the YorkUrbanDB results in a threshold of 8 pixels, 2.8 times the threshold we employed.  This looser threshold is convenient for the deep networks, which by necessity use sub-sampled images and struggle to localize segments with precision.   Fig. \ref{fig:crop} shows an example.

\begin{figure*}[htbp]
	\centering
	\subfloat[]{\includegraphics[width=0.32\textwidth]{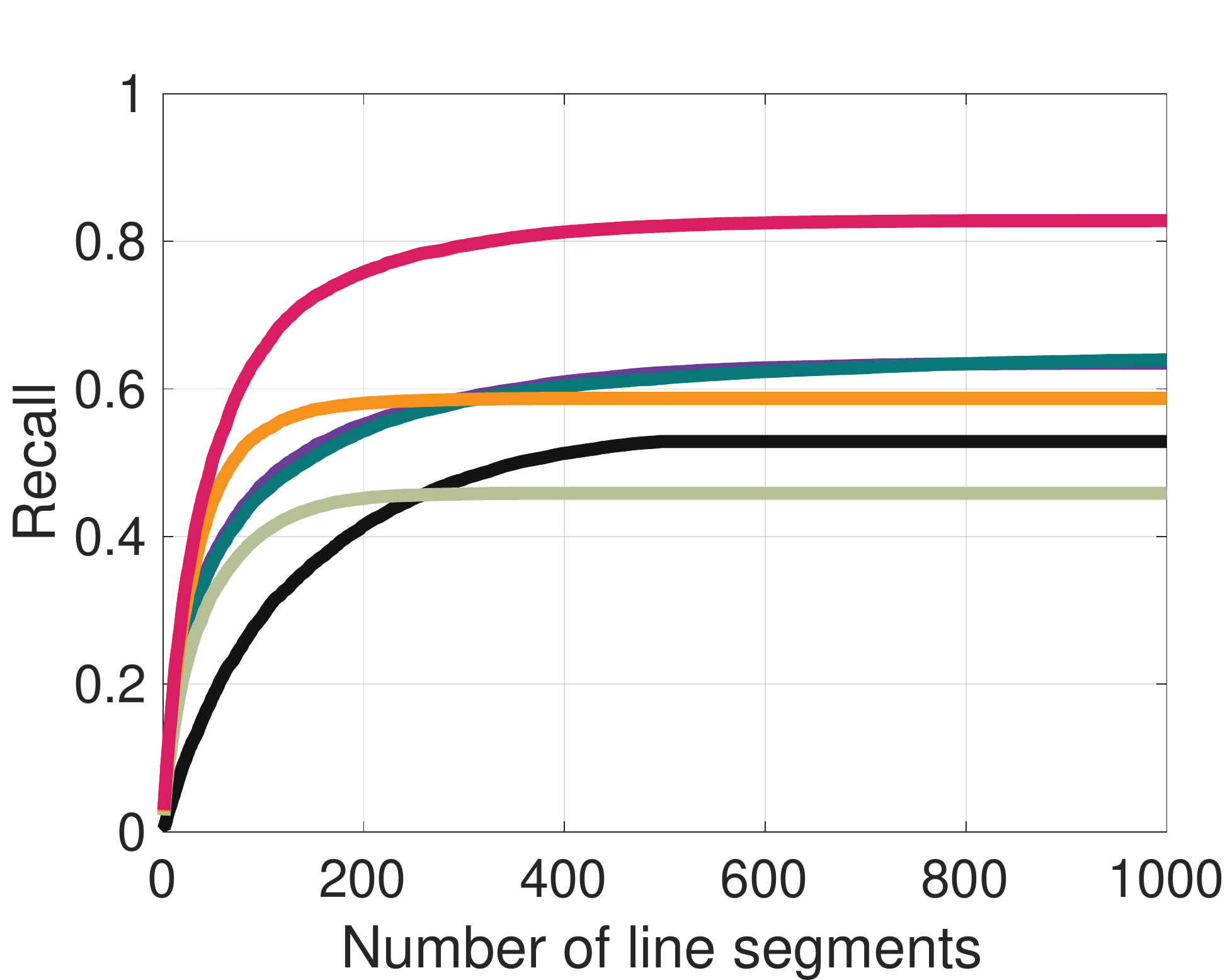}}
	\subfloat[]{\includegraphics[width=0.32\textwidth]{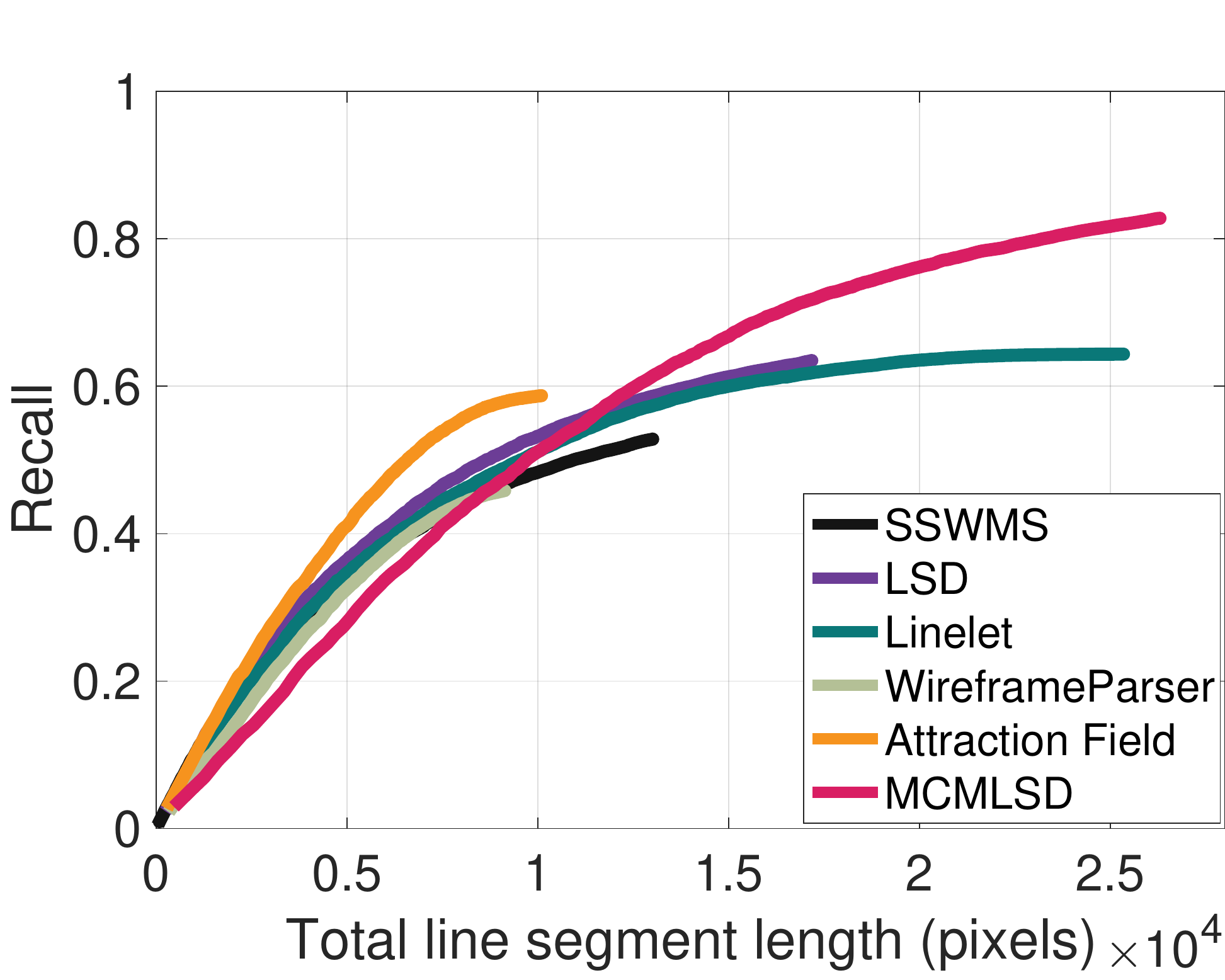}}
	\subfloat[]{\includegraphics[width=0.32\textwidth]{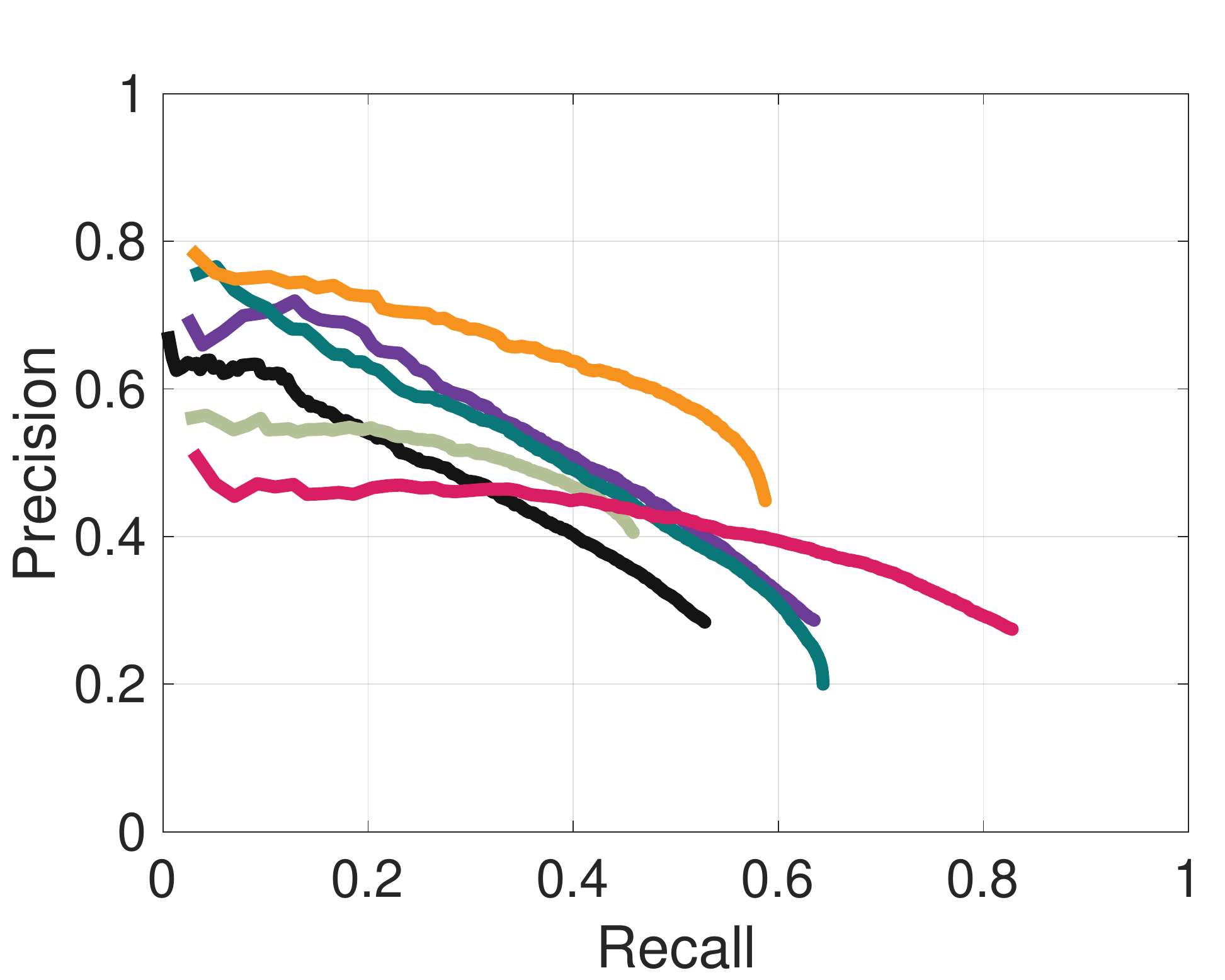}}
	\caption{Evaluation on the YorkUrbanDB dataset {\bf using the looser distance threshold of 8 pixels} employed in the Wireframe~\cite{wireframe_cvpr18} and Attraction Field~\cite{xue2018learning} papers.  (a) Recall as a function of number of segments returned.  (b) Recall as a function of the total length of segments returned.  (c) Precision-Recall.}
	\label{fig:8pixel}
\end{figure*}

To assess the importance of this threshold, we re-evaluated all algorithms using the looser threshold of 8 pixels.  Fig. \ref{fig:8pixel} shows the results for the YorkUrbanDB dataset.  We see that as we loosen the threshold, performance rises for all algorithms, but the performance of the deep algorithms (Attraction Field and Wireframe Parser) rises disproportionately, confirming the poorer localization performance of these methods.

\subsubsection{Enforcing 1:1 Matches}
While the looser distance threshold clearly helps the deep algorithms, Fig. \ref{fig:8pixel} makes clear that this alone cannot fully account for the claim that deep networks uniformly perform better than MCMLSD and earlier algorithms such as LSD and Linelet.  Here we address a more serious issue that gets to the heart of what we mean by line segment detection. 

In both the Wireframe and Attraction Field papers, a very simple method is employed to match algorithm segments and ground truth:  Points on detected segments that lie within 8 pixels of a ground truth segment are identified as hits.  Normalizing by the total length of the detected segments and the ground truth segments forms the precision and recall measures, respectively.

In Section \ref{sec:eval} of this paper and in our original CVPR  paper~\cite{almazen2017dynamic}, we were careful to articulate the problems with this simplistic approach.
First, when matching points on detected segments with points on ground truth segments, it is important that these matches be 1:1.  In other words, the same ground truth point should not be used to generate hits for multiple points on detected segments.  Similarly, the same point on a detected segment should not be matched to multiple ground truth points.   Importantly, this constraint penallizes algorithms that generate multiple detections for a single ground truth segment, or that confuse two neighbouring ground truth segments as a single segment.  Note that not enforcing this constraint leads to pathological results.  For example, an algorithm that generates dense, 16-pixel wide regions of filled pixels centred on the ground truth segments will be evaluated to have perfect precision and perfect recall.

However, enforcing 1:1 matches between points is not enough.  The problem of line segment detection is not to detect isolated edges but to recover the continuous line segments present in an image.  The output line segments can be coded in various ways, e.g., by the 2D locations of their endpoints.  Critically, the output is more than an edge map:  each line segment is a higher-level organization of edge points into a more global representation.

This means that to fairly evaluate a line segment detector, the 1:1 constraint must be applied at the segment level, not at the pixel level.  As articulated in Section \ref{sec:eval} of this paper and in our original CVPR  paper~\cite{almazen2017dynamic}, this is critical in order to penalize under- and over-segmentation.  Again, not imposing this constraint will lead to pathological results.  For example, an algorithm that returns a scatter of tiny line segments that are all only one pixel long but lie within the distance threshold of ground truth and account for all ground truth points will generate perfect precision and recall scores.

\begin{figure*}[htbp]
	\centering
	\subfloat[]{\includegraphics[width=0.32\textwidth]{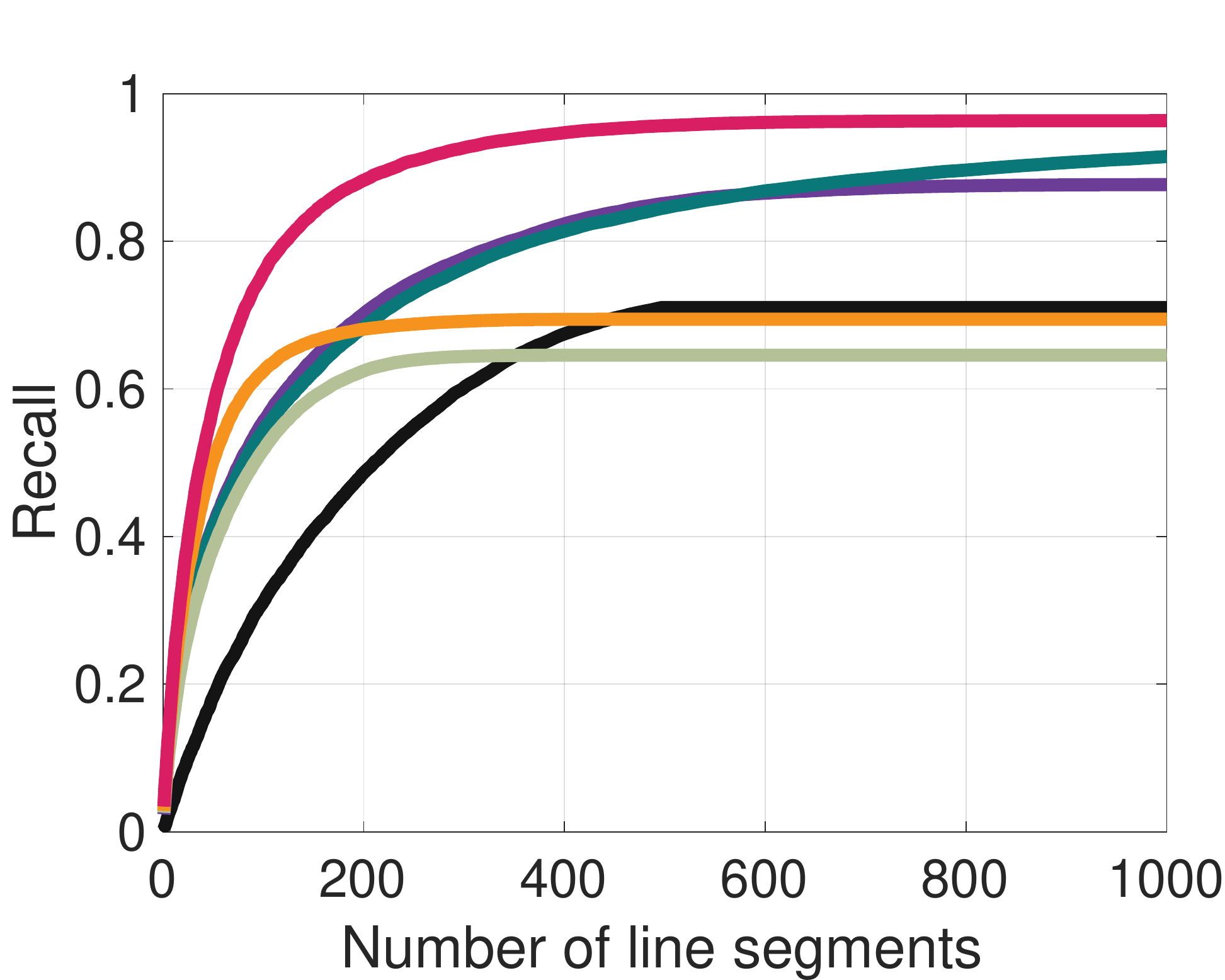}}
	\subfloat[]{\includegraphics[width=0.32\textwidth]{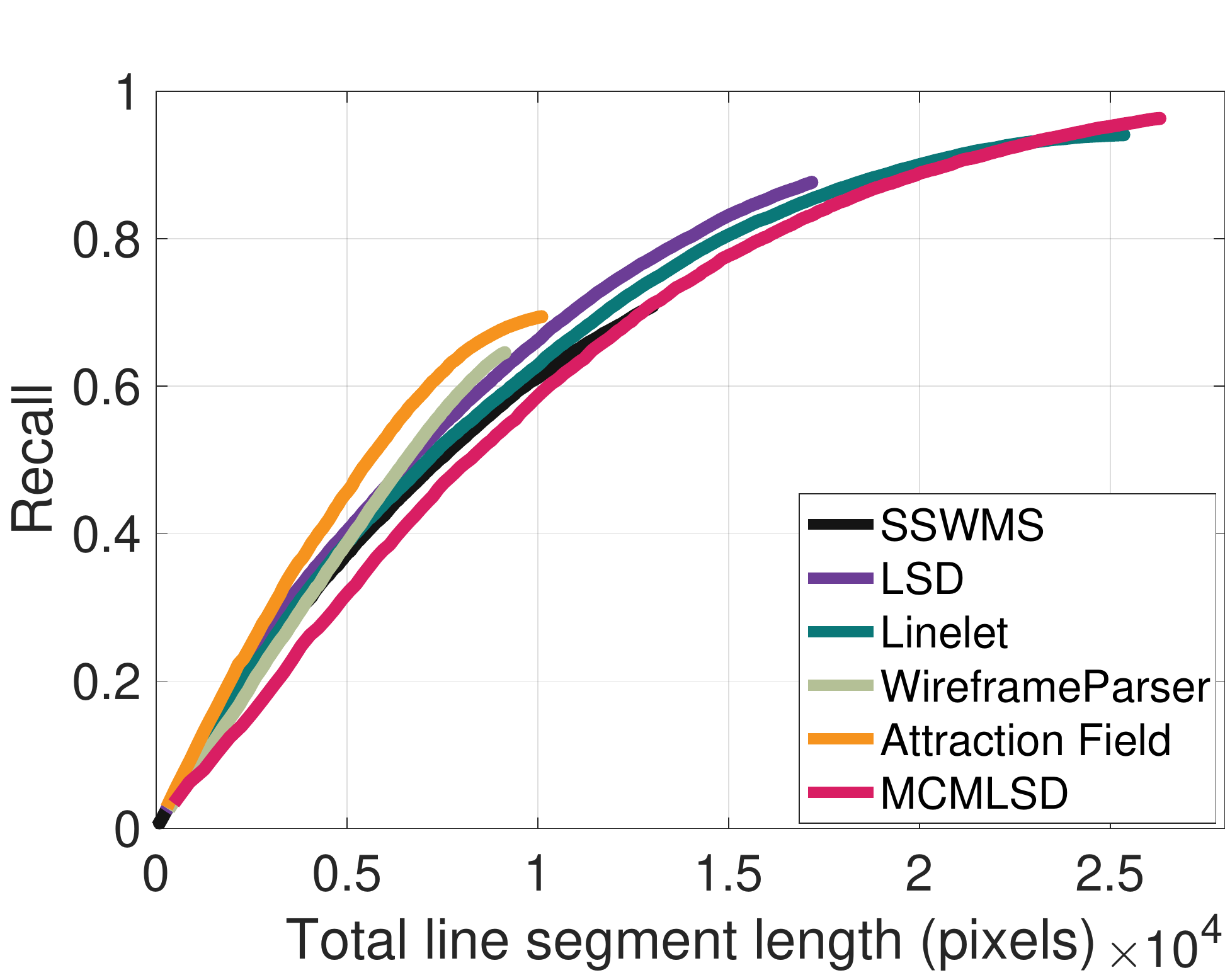}}
	\subfloat[]{\includegraphics[width=0.32\textwidth]{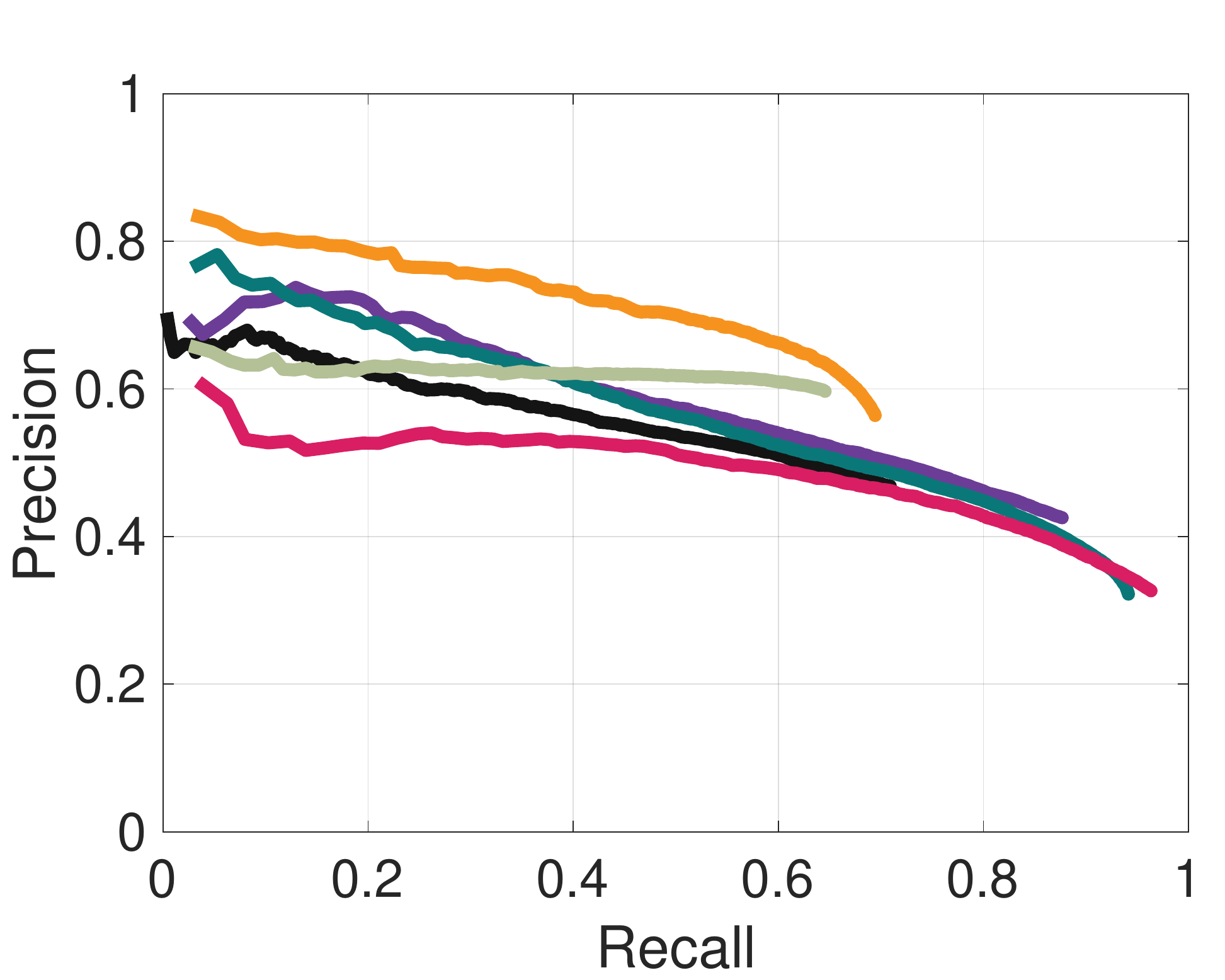}}
	\caption{Performance of MCMLSD compared with the state of the art on the YorkUrban Dataset {\bf using pixel level evaluation}.  (a) Recall as a function of number of segments returned.  (b) Recall as a function of the total length of segments returned.  (c) Precision-Recall. }
	\label{fig:evalEdgeYork}
\end{figure*}

\begin{figure*}[htbp]
	\centering
	\subfloat[]{\includegraphics[width=0.32\textwidth]{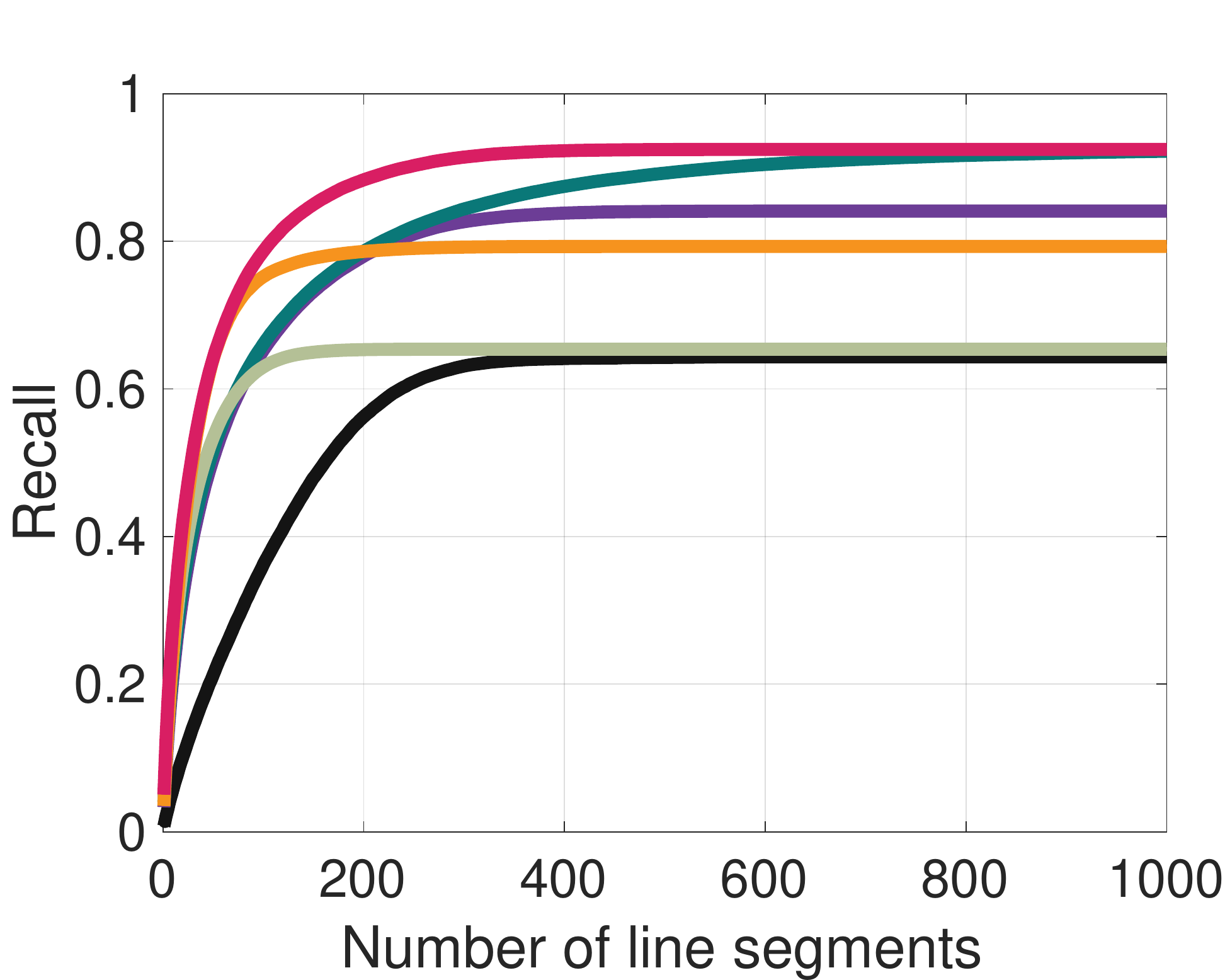}}
	\subfloat[]{\includegraphics[width=0.32\textwidth]{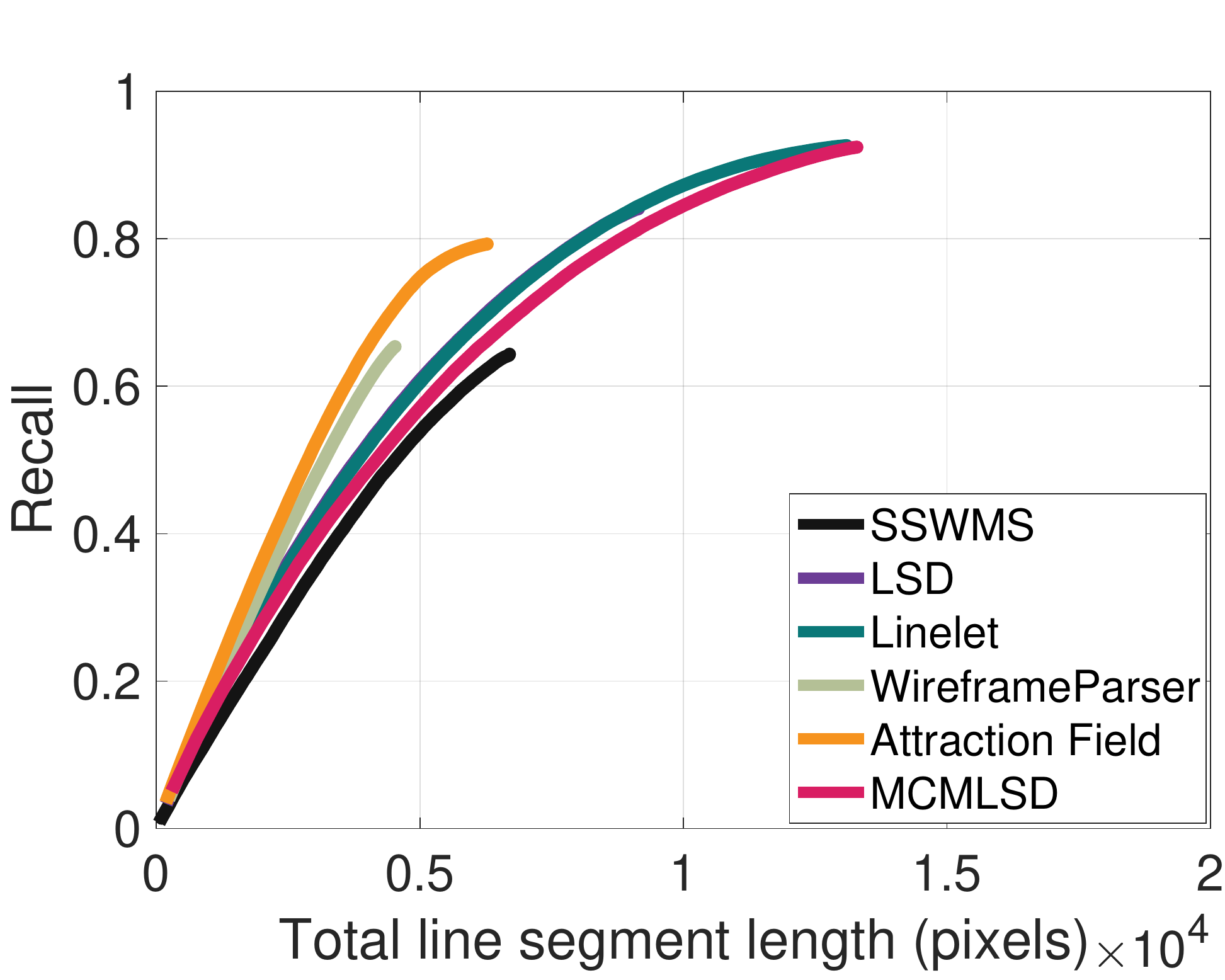}}
	\subfloat[]{\includegraphics[width=0.32\textwidth]{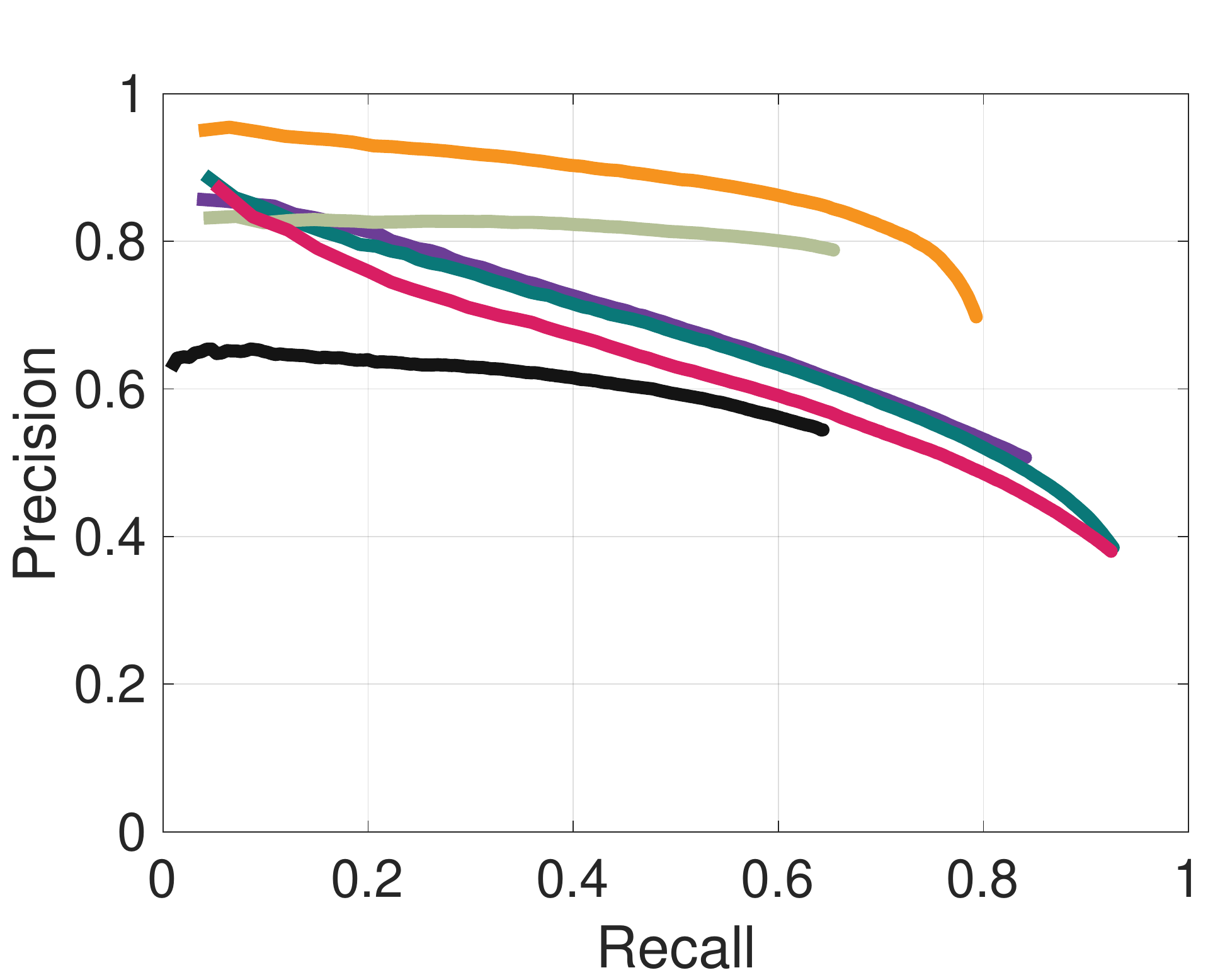}}
	\caption{Performance of MCMLSD methods compared with the state of the art on the Wireframe dataset {\bf using pixel level evaluation}.  (a) Recall as a function of number of segments returned.  (b) Recall as a function of the total length of segments returned.  (c) Precision-Recall. }
	\label{fig:evalEdgeWireframe}
\end{figure*}

To assess the importance of this segment-level 1:1 matching constraint, we re-evaluated all algorithms {\em without} this constraint, i.e., using the simple matching method employed in the Wireframe Parser~\cite{wireframe_cvpr18} and Attraction Field~\cite{xue2018learning} papers, and also using the looser distance threshold employed in these papers.  As shown in Figs. \ref{fig:evalEdgeYork} and \ref{fig:evalEdgeWireframe}, despite this relaxation in the evaluation criteria, MCMLSD still outperforms the deep learning algorithms in terms of maximum recall and recall as a function of the number of line segments returned.  However, the authors of these deep learning papers did not report these measures of performance but only the precision-recall curves shown in Figs. \ref{fig:evalEdgeYork}(c) and \ref{fig:evalEdgeWireframe}(c).  Here we see that
removing the 1:1 matching constraint particularly advantages the deep Wireframe and Attraction Field algorithms, leading to clear superiority of the Attraction Field method in the low-recall regime, although MCMLSD and Linelet methods still achieve much higher recall.  But again, we reminder the reader of the limitations of precision measures for these incomplete datasets (Section \ref{sec:eval}).

\subsection{Summary of Quantitative Results}
To summarize, the relative performance of line segment detection algorithms very much depends on how performance is measured.
Recent deep learning papers have loosened distance thresholds and not enforced 1:1 matching constraints, and under these conditions
they achieve higher precision, although still inferior recall.  It is possible that there are some applications for which this measure of performance
is appropriate.  For example, one may attempt to use only a pixel-level Hausdorff distance to register two images or to register an image to a CAD model.
However, for most downstream applications, e.g.,  single-view 3D reconstruction~\cite{ramalingam2013lifting,Qian:18}, an {\em organization} of points
into accurate line segments is desirable, and to evaluate this one must impose 1:1 matching constraints.

\begin{figure*}[!htbp]
	\centering
	\subfloat[]{\includegraphics[width=0.32\textwidth]{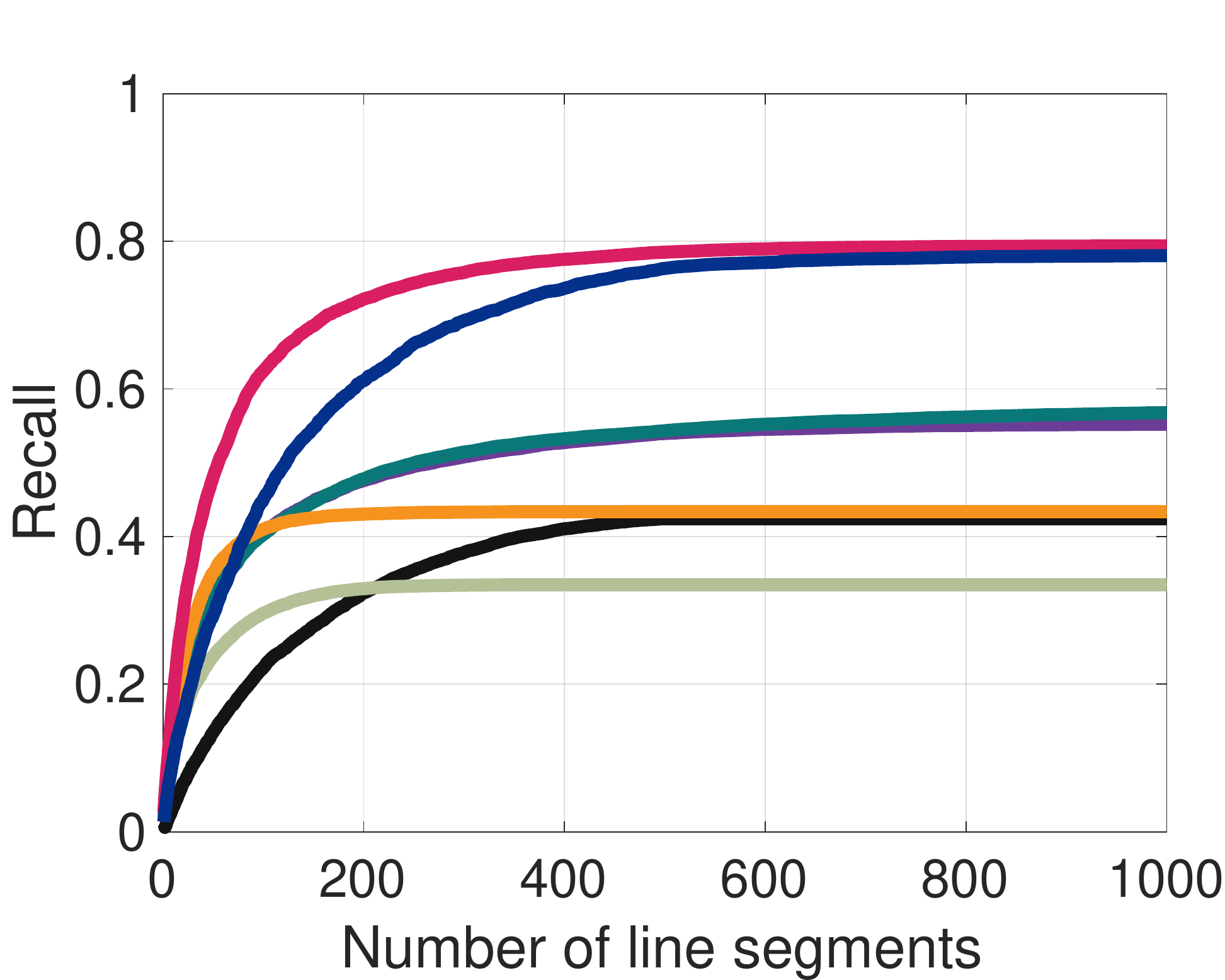}}
	\subfloat[]{\includegraphics[width=0.32\textwidth]{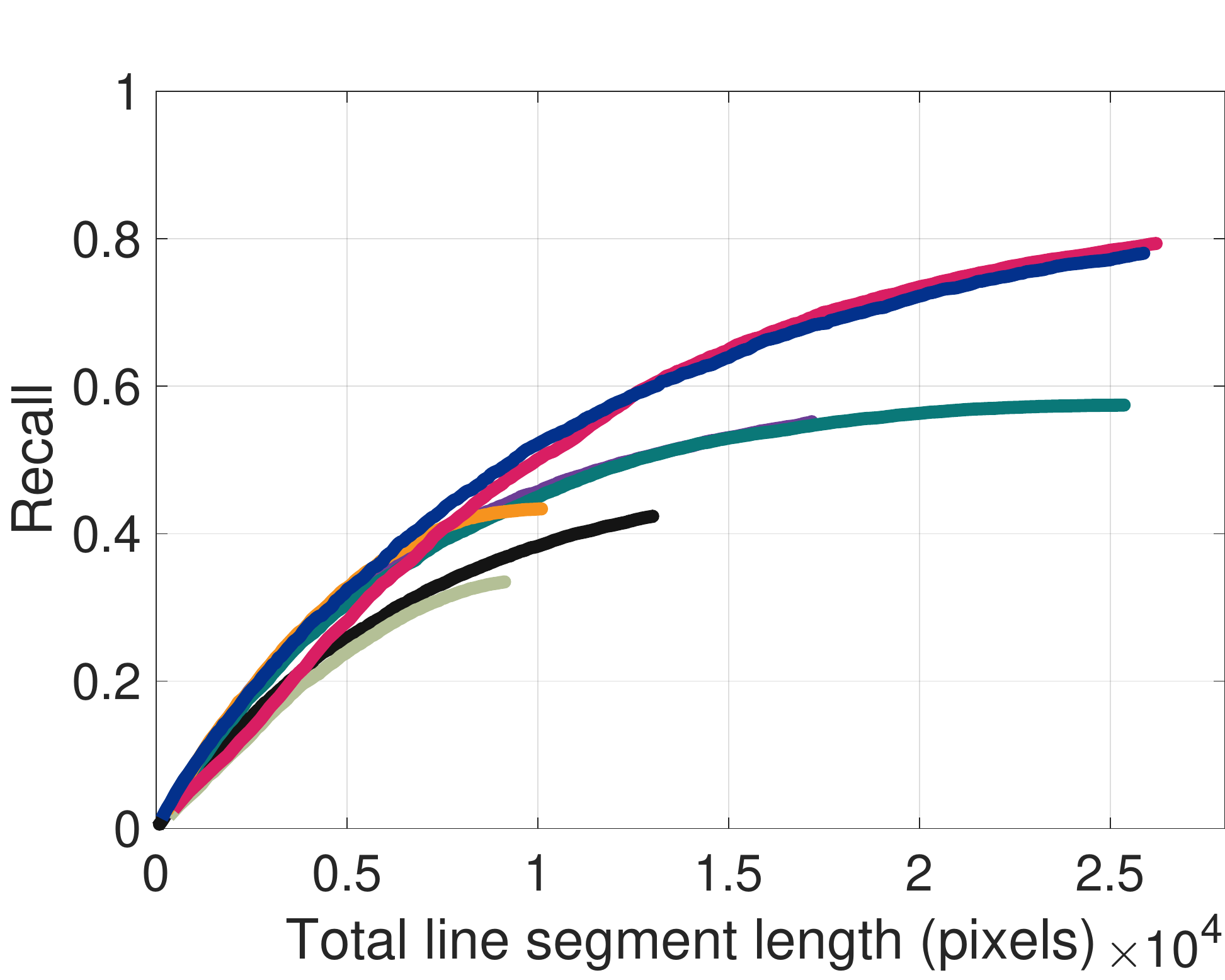}}
	\subfloat[]{\includegraphics[width=0.32\textwidth]{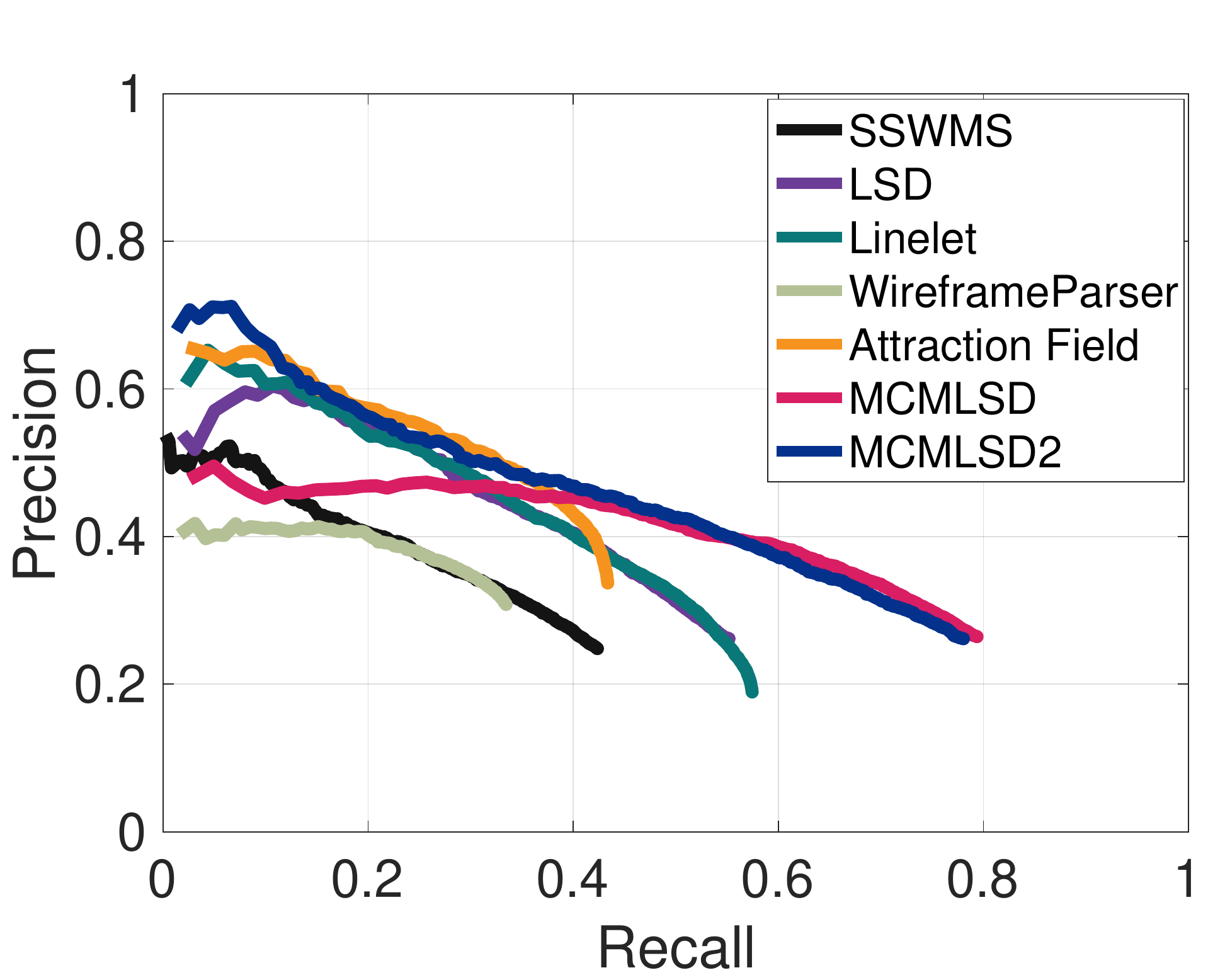}}
	\caption{Summary of results. (a) Recall as a function of number of segments returned.  (b) Recall as a function of the total length of segments returned.  (c) Precision-Recall.}
	\label{fig:rerank2}
\end{figure*}

Fig. \ref{fig:rerank2} (copied from Fig. \ref{fig:rerank} for convenience) summarizes performance relevant to these requirements, specifically for a distance
threshold of $2\sqrt{2}$ to ensure accuracy, and a 1:1 matching constraint imposed at the segment level.  Here we see that by any of the three measures
of performance, one of the two versions of MCMLSD is recommended.  If the number of output lines is to be restricted and recall is the priority, the original
MCMLSD algorithm vastly outperforms other methods.  However, if precision-recall performance is the priority, then MCMLSD2 is recommended, as it matches
or surpasses the performance of all other methods in the low-recall regime while vastly outperforming in the high-recall regime.

\section{Image Resolution}
One limitation of current deep learning methods is that the computational load for learning and inference may become untenable for higher image resolutions.
In contrast, the MCMLSD algorithm adapts well to different image resolutions without fine-tuning as long as the transition probabilities are scaled appropriately (Section \ref{sec:segDetection}).  For example, doubling the resolution requires that the transition probabilities from OFF to ON and from ON to OFF be halved.

\begin{figure*}[!htbp]
	\centering
	\subfloat[$640{\times}480$ pixels]{\includegraphics[width=0.45\textwidth]{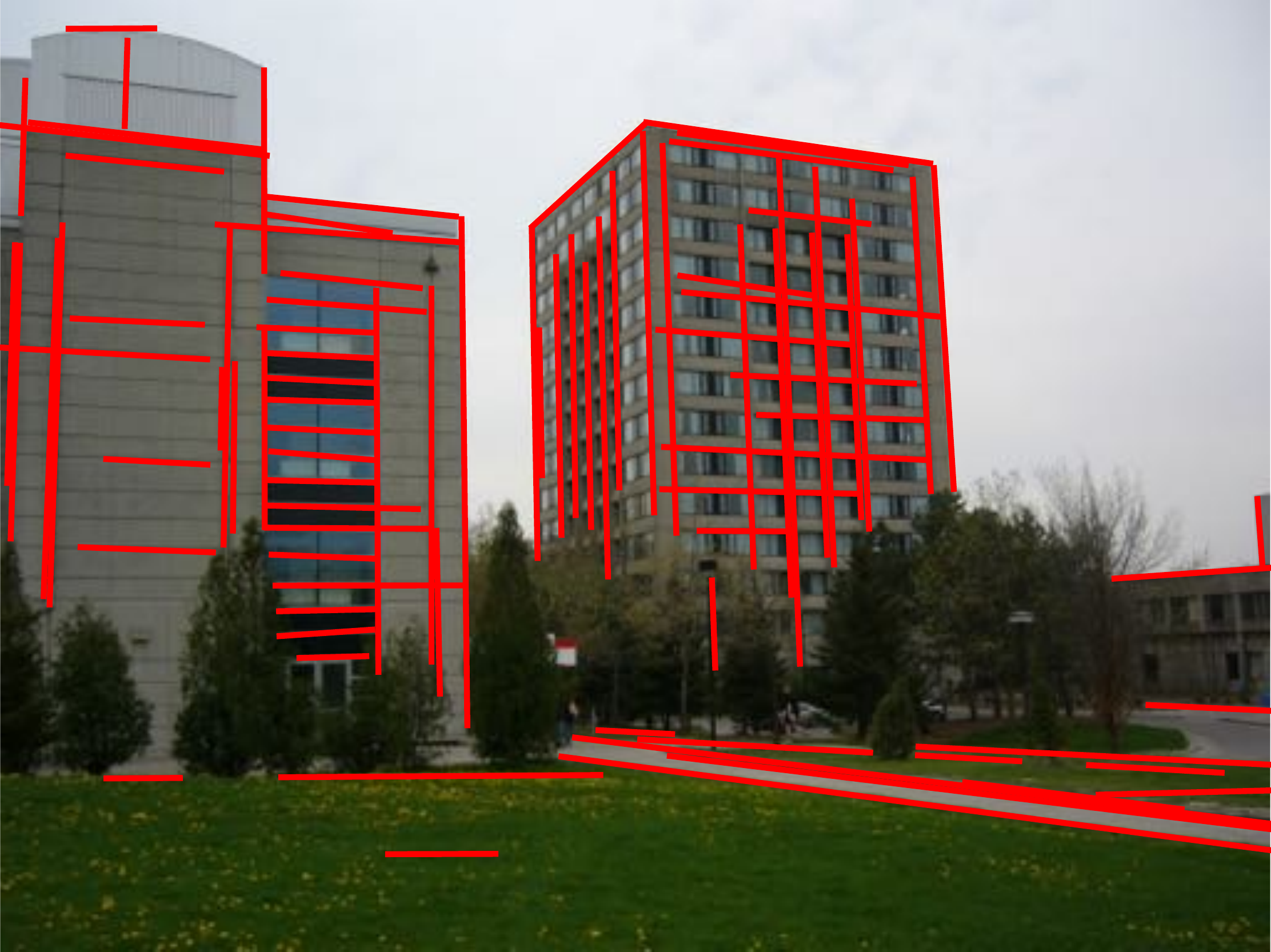}}\hspace{0.05\textwidth}
	\subfloat[$1280{\times}960$ pixels]{\includegraphics[width=0.45\textwidth]{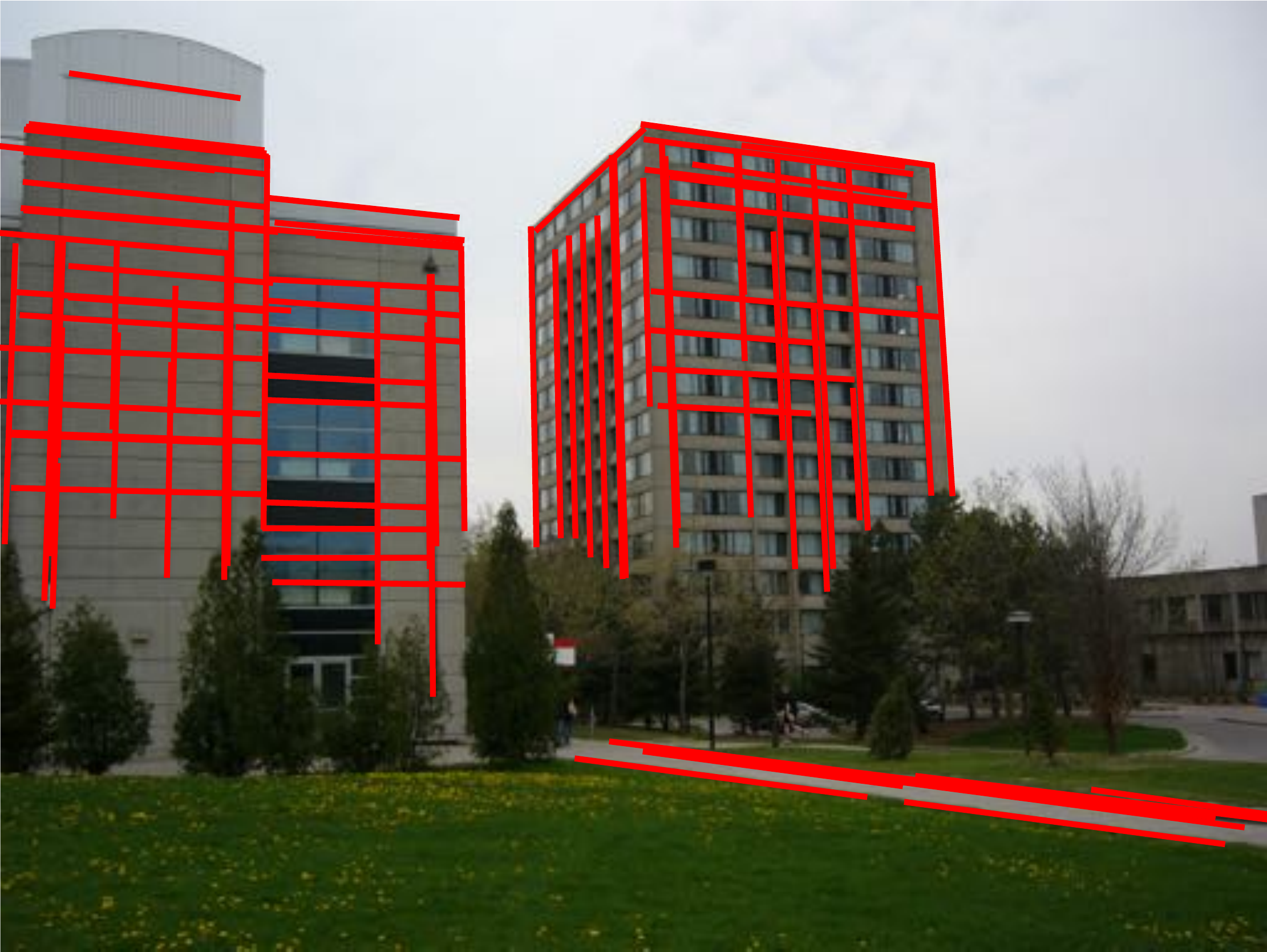}}
	\caption{Top 90 segments for MCMLSD on an example image at low and high resolutions.}
	\label{fig:high_res} 
\end{figure*}

Fig.~\ref{fig:high_res} shows the top 90 segments returned for an example image from the York UrbanDB dataset at normal ($640{\times}480$ pixel) and high ($1280{\times}960$ pixel) resolutions.  Note that the algorithm is able to take advantage of the higher resolution to deliver more complete and accurate segments.

\section{Run Time}
The dynamic programming solution for line segment detection runs in $O\left(N\right) = O\left(\sqrt{n}\right)$ time, where $N$ is the number of point samples on the line and $n$ is the number of pixels in the image.  Given a set of $m$ detected lines, the total time complexity of line segment extraction is $O\left(m\sqrt{n}\right)$.

Table~\ref{tab:runtime} shows the average run time for the six algorithms tested here on the $640{\times}480$ pixel images of the YorkUrbanDB training dataset. The SSWMS, LSD, Linelet and MCMLSD algorithm were tested using a 3.4 GHz Intel Core i7 with 8GB RAM. The deep network Wireframe and Attraction Field algorithms were tested using an NVIDIA Titan X GPU with Xeon E5-2620 2.10GHz CPU. 

The MATLAB implementation of our MCMLSD algorithm has an average run time of 2.81 sec per image.    Aside from the Linelet algorithm, which is very slow, the other algorithms are optimized and implemented in C++, returning results within a few hundred milliseconds.  

About 63\% of our run time is taken by the probabilistic Hough method for line extraction~\cite{tal2013accurate}, which we believe could be sped up considerably with more efficient coding practices and implementation in C or C++.  There are also many opportunities for mapping to parallel hardware, as edge detection is dominated by convolutions and in the dynamic programming line segment detection stage, lines separated by more than 4 pixels are processed independently.
\begin{table}[htbp]
	\caption{Average number of segments returned and run time per image for the six systems evaluated.}
	\centering
	\begin{tabular}{|c|c|c|}
		\hline
		Algorithm & \# Segments & Run Time (sec)\\
		\hline
		SSWMS & 391 & 0.30\\
		\hline
		LSD & 537 & 0.27\\
		\hline
		Linelet & 967 & 34.5\\
		\hline
		Wireframe & 228 & 0.446\\ 
		\hline
		Attraction Field & 303 & 0.152\\
		\hline
		MCMLSD & 488 & 2.81\\
		\hline
	\end{tabular}
	\label{tab:runtime}
\end{table}

\section{Conclusions}
We have developed and evaluated a novel method for line segment detection called MCMLSD that combines the  advantages of global probabilistic Hough methods for line detection with spatial analysis in the image domain to identify segments.    The key insight is that limiting segment search to Hough-detected lines leads naturally to a Markov chain formulation that allows maximum probability solutions to be computed exactly in linear time.  Our method also has the advantage that it can detect multiple segments lying on the same line, a common scenario for images of the built environment.  This formulation leads to a natural probabilistic measure for ranking segments based upon the sum over point marginals, which maximizes the expected number of correctly labelled points on detected lines. 

A second contribution is our new methodology for evaluating line segment detectors on an incomplete labelled dataset.  By constraining matches between ground truth and detector output to be 1:1 at the segment level, we show that under- and over-segmentation are penalized appropriately.  Using this new evaluation methodology we find that MCMLSD outperforms the state-of-the-art by a substantial margin.  The code for MCMLSD and our evaluation method is available at \url{www.elderlab.yorku.ca/resources}.

\section*{Acknowledgements}
This research was supported by an NSERC Discovery grant and by the NSERC CREATE Training Program in Data Analytics \& Visualization.

\ifCLASSOPTIONcaptionsoff
  \newpage
\fi



%
\bibliographystyle{IEEEtran}
\bibliography{IEEEabrv,MCMLSD_journal}

%
\begin{IEEEbiography}
[{\includegraphics[width=1in,height=1.25in,clip,keepaspectratio]{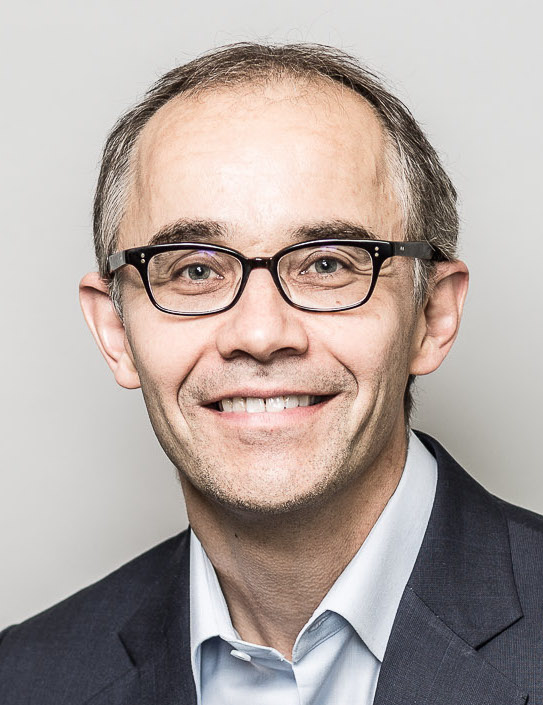}}]
{James H. Elder}
received the BASc degree in
electrical engineering from the University of
British Columbia in 1987 and the PhD degree
in electrical engineering from McGill University in
1995. From 1995 to 1996, he was a senior research associate at the NEC
Research Institute in Princeton, New Jersey. He
joined the faculty of York University, Canada, in
1996, where he is presently Professor and York Research 
Chair in Human and Computer Vision, jointly appointed to the Department
of Electrical Engineering and Computer Science and the Department of Psychology.
His research seeks to improve machine vision systems through a better understanding of visual processing in biological systems, with a current focus on natural scene statistics, perceptual organization, contour processing, shape perception, single-view 3D reconstruction, attentive vision systems and machine vision systems for dynamic 3D urban awareness.  
\end{IEEEbiography}

\begin{IEEEbiography}
[{\includegraphics[width=1in,height=1.25in,clip,keepaspectratio]{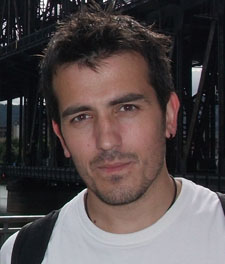}}]
{Emilio~J.~Almaz\`{a}n}
received the BASc degree in computer science and MSc in computer vision from Rey Juan Carlos University in 2005 and 2011 respectively, and the PhD in computer vision from Kingston University in 2014. He was a postdoctoral fellow at the Center for Vision Research, York University in 2015, since then he has been working as a computer vision researcher in the R\&D department at Nielsen. In 2019 he was appointed lead data scientist of the same group. His research interests include image categorization, object detection and understanding and interpretation of convolutional neural networks. Recent work has focused on combining natural language processing and vision.
\end{IEEEbiography}

\begin{IEEEbiography}
[{\includegraphics[width=1in,height=1.25in,clip,keepaspectratio]{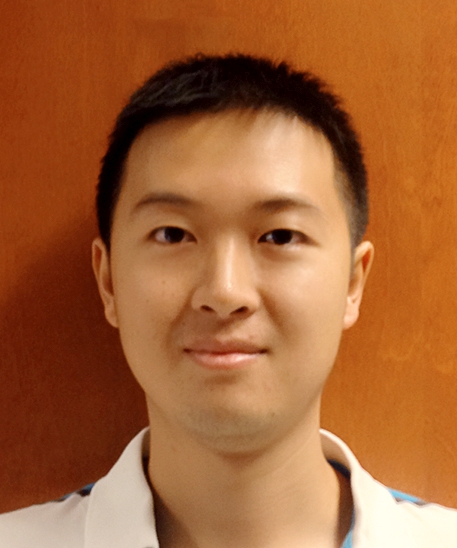}}]
{Yiming Qian}
received the BASc and MASc degree in electrical engineering from Ryerson University in 2012 and 2014. He is currently a PhD student in Computer Science at York University. From 2012 to 2015, he was an R\&D engineer at Siemens Energy Management.  His research interests include single-view 3D reconstruction, image processing, signal processing, manufacturing optimization and data mining.
\end{IEEEbiography}

\begin{IEEEbiography}
[{\includegraphics[width=1in,height=1.25in,clip,keepaspectratio]{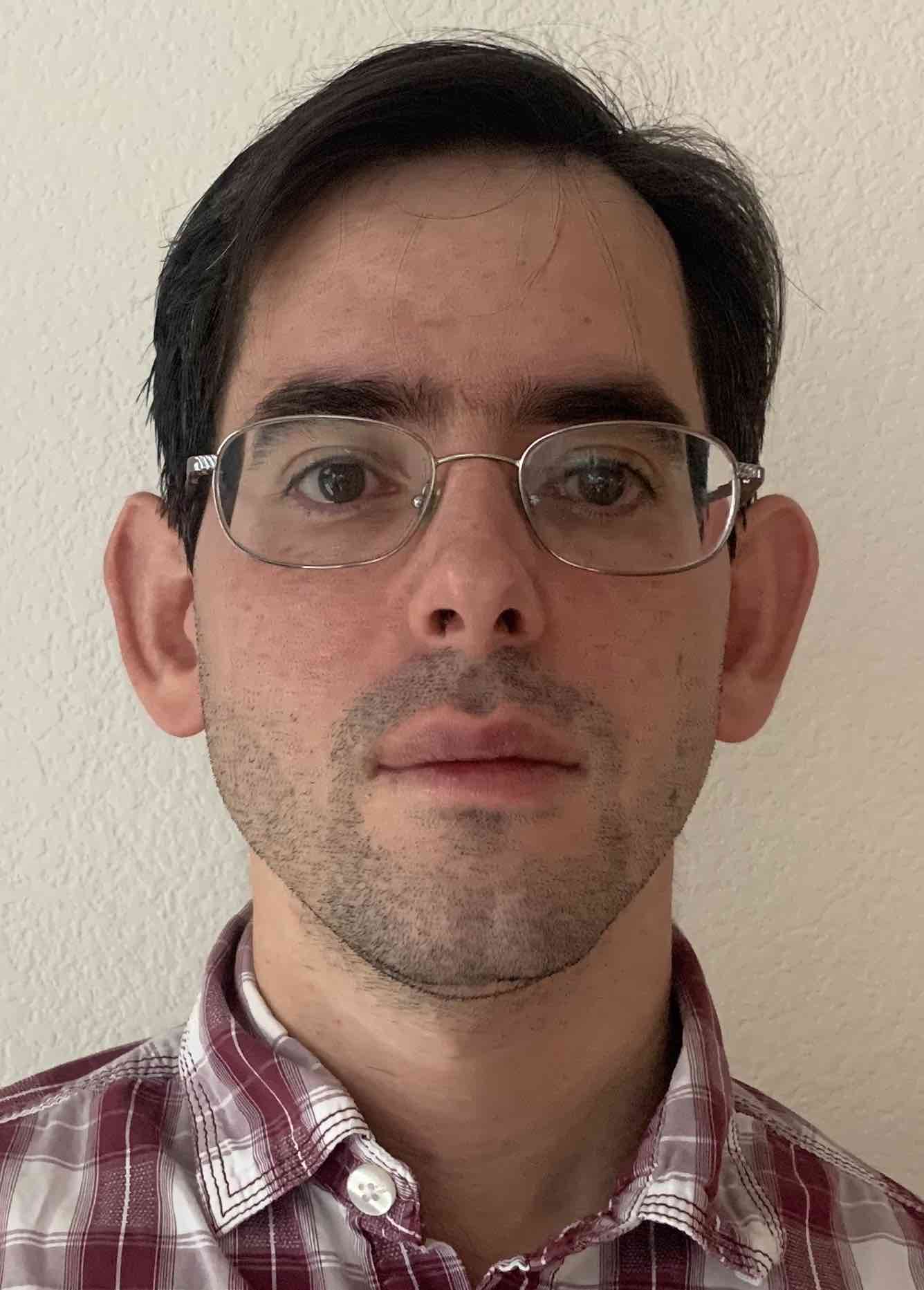}}]
{Ron~Tal}
 received the MASc degree in computer  engineering from York University in 2011.  Following his graduate work, Ron worked at Uber, where he was a founding member of Michelangelo - Uber's comprehensive machine learning service.  Currently he is a senior software engineer in the Machine Learning Platform team at Coinbase Inc. 
\end{IEEEbiography}




\end{document}